\RequirePackage{fix-cm}
\documentclass[twocolumn,natbib]{./svjour3}          
\smartqed  
\usepackage{etex}
\usepackage{graphicx}
\usepackage{amsmath}
\usepackage{url}
\usepackage{footnote}
\usepackage{amssymb}

\usepackage{tikz-qtree}
\usepackage{tikz}
\usetikzlibrary{trees}
\def\eg{\emph{e.g., }}
\def\ie{\emph{i.e., }}

\def\etc{\emph{etc}}

\newcommand{\myparagraph}[1]{\smallskip\noindent\textbf{#1:}}

\usepackage[colorlinks = true,linkcolor = blue,urlcolor  = blue,citecolor = blue,anchorcolor = blue]{hyperref}
\usepackage{color}
\usepackage{amsmath,amsfonts}
\usetikzlibrary{arrows}

\DeclareMathOperator*{\argmax}{\arg\!\max}

\renewcommand{\mid}{\,|\,}

\def\c{\mathit{cat}}
\def\s{\mathit{sc}}

\usepackage{times}
\usepackage{enumerate}
\usepackage{enumitem}

\usepackage{threeparttable}

\definecolor{Lightgray}{RGB}{235,235,235}

\colorlet{blcolor}{gray!80}

\usepackage{booktabs}
\usepackage{colortbl}
\usepackage{xcolor}

\usepackage{url}
\usepackage{wrapfig}
\usepackage{floatflt}
\usepackage{textpos}
\usepackage{chngcntr}

\usepackage{algorithmic}

\usepackage[lined,ruled,commentsnumbered]{algorithm2e}

\colorlet{tableheadcolor}{gray!25} 
\newcommand{\headcol}{\rowcolor{tableheadcolor}} %
\colorlet{tablerowcolor}{gray!10} 
\newcommand{\rowcol}{\rowcolor{tablerowcolor}} %
\newcommand{\topline}{\arrayrulecolor{black}\specialrule{0.1em}{\abovetopsep}{0pt}%
            \arrayrulecolor{tableheadcolor}\specialrule{\belowrulesep}{0pt}{0pt}%
            \arrayrulecolor{black}}
\newcommand{\midline}{\arrayrulecolor{tableheadcolor}\specialrule{\aboverulesep}{0pt}{0pt}%
            \arrayrulecolor{black}\specialrule{\lightrulewidth}{0pt}{0pt}%
            \arrayrulecolor{white}\specialrule{\belowrulesep}{0pt}{0pt}%
            \arrayrulecolor{black}}

\newcommand{\bottomlinec}{\arrayrulecolor{tablerowcolor}\specialrule{\aboverulesep}{0pt}{0pt}%
            \arrayrulecolor{black}\specialrule{\heavyrulewidth}{0pt}{\belowbottomsep}}%

\begin{document}

\title{Information Pursuit: A Bayesian Framework for Sequential Scene Parsing}

\author{Ehsan Jahangiri \and
        Erdem Y\"{o}r\"{u}k \and	
        Ren\'{e} Vidal \and
        Laurent Younes \and
	    Donald Geman
}

\institute{Ehsan Jahangiri \\
            \email{ejahang1@jhu.edu}
           \\ \\
            Erdem Y\"{o}r\"{u}k \\
            \email{eyoruk1@jhu.edu}
           \\ \\
           Ren$\acute{\text{e}}$ Vidal \\
            \email{rvidal@cis.jhu.edu}
           \\ \\
           Laurent Younes \\
            \email{laurent.younes@jhu.edu}
           \\ \\
           Donald Geman \\
            \email{geman@jhu.edu}
            \\ \\
            Center for Imaging Science, Johns Hopkins University, Baltimore, MD, USA.
}

\date{Received: date / Accepted: date}

\maketitle

\begin{abstract}
Despite enormous progress in object detection and classification, the
problem of incorporating expected contextual relationships among
object instances into modern recognition systems remains a key
challenge. In this work we propose \emph{information pursuit}, a
Bayesian framework for scene parsing that combines prior models for
the geometry of the scene and the spatial arrangement of objects
instances with a data model for the output of high-level image
classifiers trained to answer specific questions about the scene. In
the proposed framework, the scene interpretation is progressively
refined as evidence accumulates from the answers to a sequence of
questions. At each step, we choose the question to maximize the mutual
information between the new answer and the full interpretation given
the current evidence obtained from previous inquiries. We also propose
a method for learning the parameters of the model from synthesized,
annotated scenes obtained by top-down sampling from an easy-to-learn
generative scene model. Finally, we introduce a database of annotated
indoor scenes of dining room tables, which we use to evaluate the
proposed approach.

\end{abstract}

\section{Introduction}
\label{intro}

The past few years have seen dramatic improvements in the performance
of object recognition systems, especially in 2D object detection and
classification. Much of this progress has been driven by the use of
deep learning techniques, which allow for end-to-end learning of
multiple layers of low-, mid- and high-level image features, which are
used to predict, e.g., the object's class, its 2D location, or its 3D
pose, provided that sufficiently many annotations for the desired
output are provided for training the corresponding deep net.

On the other hand, automatic semantic parsing of natural scenes that
typically exhibit contextual relationships among multiple object
instances remains a core challenge in computational vision. As an
example, consider the dining room table scene shown in Figure
\ref{fig::Patch_Hierarchy}, where it is fairly common for collections
of objects to appear in a specific arrangement on the table. For
instance, a plate setting often involves a plate with a knife, a fork
and a spoon to the left or right of the plate, and a glass in front of
the plate. Also, the knife, fork and spoon often appear parallel to
each other rather than in a random configuration. These complex
spatial relationships among object poses are often not captured by
existing deep networks, which tend to detect each object instance
independently. We argue that modeling such contextual
relationships is essential for highly accurate semantic parsing
because detecting objects in the context of other objects can
potentially provide more coherent interpretations (e.g., by
avoiding object detections that are inconsistent with each other).

\myparagraph{Proposed Bayesian Framework} We propose to leverage recent advances in object classification, especially deep learning of low-, mid- and high-level features, to build high-level generative models that reason about objects in the scene rather than features in the image. Specifically, we assume we have at our disposal a battery of classifiers trained to answer specific questions about the scene (e.g., is there a plate in this image patch?) and propose a model for the output of these high-level classifiers.

The proposed model is Bayesian, but can be
seen as a hybrid of learning-based and model-based approaches.  By the
former, we refer to parsing an image by scanning it with a battery of
trained classifiers (e.g., SVMs or deep neural nets).  By the latter,
we refer to identifying likely states under the posterior distribution
in a Bayesian framework which combines a prior model over
interpretations and a data model based (usually) on low-level image
features.  In a nutshell, we maintain the battery of classifiers {\it
and} the Bayesian framework by replacing the low-level features with
high-level classifiers.  This is enabled by defining the latent
variables in one-to-one correspondence with the classifiers. In
particular, there are no low-level or mid-level features in the model;
all variables, hidden and measured, have semantic content.  We refer
to the set which indexes the latent variables and corresponding
classifiers as ``queries'' and to the latent variables as
``annobits''. For example, some annobits might be lists of binary indicators of
the presence or absence of visible instances from a subset of object
categories in a specific image patch, and the corresponding
classifiers might be CNNs which output a vector of weights for each of
these categories.  Annobits can be seen as a perfect (noiseless)
classifier and, vice-versa, the classifier can be seen as an imperfect
(noisy) annobit. The data model is the conditional distribution of the
family of classifiers given the family of annobits.

The prior model encodes our expectations about how scenes are
structured, for example encoding preferred spatial arrangements among
objects composing a dining room table setting.  Hence the posterior
distribution serves to modulate or ``contextualize'' the raw
classifier output. We propose two prior models. The first one combines a prior model of the 3D scene and camera geometry, whose parameters can be encoded by a homography, and a Markov random field (MRF) model of the 2D spatial arrangement of object instances given the homography. The model is motivated by our particular application to parsing dining room table scenes, where most objects lie on the table plane. This model is easy to sample from its posterior, but it is hard to learn tabula-rasa due to lack of modularity and therefore the need for a great many training samples. The second model is based on an attributed graph where each node corresponds to an object instance that is attributed with a category label and a pose in the 3D world coordinate system. The attributed graph is built on top of a random skeleton that encodes spatial relationships among different object instances. This model is easy to learn and sample, but sampling from its posterior is much harder. We get the best of both worlds by using the second model to synthesize a large number of annotated scenes, which are then used to learn the parameters of the first model.

\myparagraph{Proposed Scene Parsing Strategy}
Depending on the scene, running a relatively small
subset of all the classifiers might already provide a substantial
amount of information about the scene, perhaps even a sufficient
amount for a given purpose. Therefore, we propose to
annotate the data sequentially,
identifying and applying the most informative classifier (in an
information-theoretic sense) at each step given the accumulated
evidence from those previously applied.

The selection of queries is task-dependent, but some general
principles can be articulated.  We want to structure them to allow the
parsing procedure to move freely among different levels of semantic
and geometric resolution, for example to switch from analyzing the
scene as a whole, to local scrutiny for fine discrimination, and
perhaps back again depending on current input and changes in target
probabilities as evidence is acquired.  Processing may be terminated
at any point, ideally as soon as the posterior distribution is peaked
around a coherent scene description, which may occur after only a
small fraction of the classifiers have been executed.

The Bayesian framework provides a principled way for deciding what
evidence to acquire at each step and for coherently integrating the
evidence by updating likelihoods.  At each step, we select the
classifier (equivalently, the query) which achieves the maximum value
of the conditional mutual information between the global scene
interpretation and any classifier given the existing evidence (i.e.,
output of the classifiers already implemented).  Consequently, the
order of execution is determined online during scene parsing by
solving the corresponding optimization problem at each step. The proposed Information Pursuit (IP) strategy then
alternates between selecting the next classifier, applying it to the
image data, and updating the posterior distribution on interpretations
given the currently collected evidence.

\myparagraph{Application to 2D Object Detection and 3D Pose Estimation in the JHU Table-Setting Dataset}
We will use the proposed IP strategy to detect instances from multiple object categories in an image and estimate their 3D poses. More precisely, consider a 3D scene and a semantic description consisting of a variable-length list of the identities and 3D poses of visible instances from a pre-determined family of object categories. We want to recover this list by applying high-level classifiers to an observed image of the scene acquired from an unknown viewpoint. As a proof of concept, we will focus on indoor scenes of dinning room tables, where the specific categories are plate, glass, utensil and bottle. Such scenes are challenging due to severe occlusion, complex photometry and intra-class variability. In order to train models and classifiers we have collected and manually labeled $3000$ images of table settings from the web. We will use this dataset for learning our model, training and testing the classifiers, and evaluating system's performance.  We will show that we can make accurate decisions about existing object instances by processing only a small fraction of patches from a given test image.  We will also demonstrate that coarse-to-fine search naturally emerges from IP.

\myparagraph{Paper Contributions} In summary, the core contribution of our work is a Bayesian framework for semantic scene parsing that combines (1) a data model on the output of high-level classifiers as opposed to low-level image features, (2) prior models on the scene that captures rich contextual relationships among instances of multiple object categories, (3) a progressive scene annotation strategy driven by stepwise uncertainty reduction, and (4) a dataset of table settings.

\myparagraph{Paper Outline}
The remainder of the paper is organized as follows.
In section~\ref{sec:related} we summarize some related work.
In section~\ref{sec:approach} we define
the main system variables and formulate information pursuit in
mathematical terms.  In section~\ref{sec:foundation} we introduce the annobits and the annocell hierarchy.  In section~\ref{sec:prior} we
introduce our prior model on 3D scenes, which includes a prior model on interpretation units and a prior model on scene geometry and camera parameters. In section \ref{sec:scene_generation} we introduce a novel scene generation model for synthesizing 3D scenes, which is used to learn the parameters of the prior model. The algorithm for sampling from the posterior distribution, a crucial
step, is spelled out in section~\ref{Conditional_Sampling} and the particular classifiers (CNNs) and data model (Dirichlet distributions) we use in our
experiments are described in section~\ref{sec:CNNs}. In section \ref{sec:dataset} we introduce the ``JHU Table-Setting Dataset'', which is composed of about 3000 fully annotated scenes, which we use for training the prior model and the classifiers. In
section~\ref{sec:experiments} we present comprehensive experiments,
including comparisons between IP and using the CNNs alone.  Finally,
there is a concluding discussion in section~\ref{sec:conclusion}.

\section{Related Work}
\label{sec:related}

The IP strategy proposed in this work is partially motivated by the
``divide-and-conquer'' search strategy employed by humans in playing
parlor and board games such as ``Twenty Questions,'' where the
classifiers would represent noisy answers, as well as by the capacity
of the human visual system to select potential targets in a scene and
ignore other items through acts of selective
attention~\citep{Serences2006,Reynolds99}.  An online algorithm
implementing the IP strategy was first introduced
by~\citet{GemanJedynak96} under the name ``active testing'' and
designed specifically for road tracking in satellite images. Since then,
variations on active testing have appeared in~\citep{SznitmanAT2010} for face detection and localization, in~\citep{branson2014ignorant}
for fine-grained classification, and in~\citep{Sznitman2013} for
instrument tracking during retinal microsurgery.
However, it has not yet been applied to problems of the
complexity of 3D scene interpretation.

CNNs, and more generally deep learning with
feature hierarchies, are everywhere. Current CNNs are designed
based on the same principles introduced years ago in~\citep{Homma88,Lecun98}.
In the past decade, more efficient ways to train neural
networks with more layers~\citep{Hinton2006,Bengio2007,Ranzato2007} together with far larger annotated training sets (e.g.,
large public image repositories such as ImageNet~\citep{imagenet_cvpr09}) and efficient implementations on
high-performance computing systems, such as GPUs and large-scale
distributed clusters~\citep{Dean2012,Ciresan2011} resulted in the success of deep learning and more specifically CNNs. This has resulted
in impressive performance of CNNs on a number of benchmarks and
competitions including the \emph{ImageNet Large Scale Visual
Recognition Challenge} (ILSVRC)~\citep{ILSVRC15}.  To achieve better
performance, the network size has grown constantly in the past few
years by taking advantage of the newer and more powerful computational
resources.

State-of-the-art object detection systems (\eg RCNN~\cite{GirshickMalik2016} and faster RCNN~\cite{ren_fasterrcnn}) initially generate some proposal boxes which are likely to contain object instances; these boxes are then processed by the CNN for classification, and then regressed to obtain better bounding boxes for positive detections. In RCNN~\cite{GirshickMalik2016}, the proposals are generated using the ``selective search'' algorithm~\cite{Uijlings13}. The selective search algorithm generates candidates by various ways of grouping the output of an initial image segmentation.  The faster region-based CNN (faster RCNN) of \cite{ren_fasterrcnn} does not use the selective search algorithm to generate the candidate boxes; their network generates the proposals internally in the forward path. These approaches do not use contextual relations to improve disambiguation and prevent inconsistent interpretations, allow for progressive annotation, or accommodate 3D representations.  There is no image segmentation in our approach.

There is a considerable amount of work attempting to incorporate
contextual reasoning into object recognition.  Frequently this is
accomplished by labeling pairs of regions obtained from segmentation
or image patches using Conditional Random Fields or Markov Random
Fields \citep{Rabinovich07, MottaghiCVPR14, Sun14,
Desai_IJCV2011}. Compositional vision \citep{Geman2002} embeds context in a broader
sense by considering more general, non-Markovian models related to
context-sensitive grammars.  While most of the work is about
discriminative learning and reasoning in 2D \citep{CTW12,
Sun14,Desai_IJCV2011,Felzenszwalb2010,Porway2a,Hoai14,Rabinovich07}, several attempts have been made recently at designing models that
reason about surfaces of 3D scenes and the interaction between objects
and their supporting surfaces
\citep{Bao2010,Hoiem2007,Lee10,Silberman12,Saxena2008,Zhu2014}. It
has been shown that reasoning about the underlying 3D layout of the
scene is, as expected, useful in recognizing interactions with other
objects and surfaces \citep{Bao2010, HoiemBook2011}.  However, most of the
current 3D models do not encode contextual relations among objects on
supporting surfaces beyond their coplanarity.

\section{General Framework}
\label{sec:approach}

\subsection{Scenes and Queries}
\label{sec:scenes_queries}
Let $\mathcal Z$ be a limited set of possible interpretations or
descriptions of a physical 3D scene and let $I$ be a 2D image of the
scene.  In this paper, a description $Z \in \mathcal Z$ records the
identities and 3D poses of visible instances from a pre-determined
family of object categories ${\mathcal C}$.  The scene description is unknown, but
the image $I$ is observed and is determined by the scene together with
other, typically unobserved, variables $W$,
including the camera's intrinsic and extrinsic parameters.
We will assume that $Z$, $W$ and $I$ are random
variables defined on a common probability space.

The goal is to reconstruct as much information as possible about $Z$
from the observation $I$ and to generate a corresponding semantic
rendering of the scene by visualizing object instances. In our setting, information about $Z$ is supplied by noisy answers to
a series of image-based queries from a specified set $\mathcal Q$.
We assume the true answer $Y_q$ to a query $q\in\mathcal{Q}$ is determined by $Z$ and
$W$; hence, for each $q\in \mathcal Q$, $Y_q = f_q(Z, W)$ for some
function $f_q$.  The dependency of $Y_q$ on $W$ allows the queries to
depend on locations relative to the observed image.  We regard $Y_q$
as providing a small unit of information about the scene $Z$, and
hence assuming a small set of possible values, even just two, i.e., $Y_q
\in \{0,1\}$ corresponding to the answers ``no'' or ``yes'' to a
binary query.  We will refer to every $Y_q$ as an ``annobit''
whether or not $q$ is a binary query. Also, for each subset of queries $V \subset \mathcal Q$, we will denote the corresponding subset of annobits as $Y_V = (Y_q \mid q \in V)$ and similarly for classifiers $X_V$ (see below).

We will progressively estimate the states of the annobits from a
matched family of image-based predictors.  More specifically, for each
query $q\in \mathcal Q$, there is a corresponding classifier $X_q$, where $X_q = h_q(I)$ for some function $h_q$.  We will assume
that each classifier has the same computational cost; this is
necessary for sequential exploration based on information flow
alone to be meaningful, but can also be seen as a constraint on
the choice of queries $\mathcal Q$. We will further assume that $Y_{\mathcal Q}$ is a sufficient statistic for $X_{\mathcal Q}$ in the sense that
\begin{equation}
P(X_{\mathcal Q}|Z,U) = P(X_{\mathcal Q}|Y_{\mathcal Q}).
\end{equation}
We will use a Bayesian model.  The prior model is composed of
a scene model for $Z$, which
encodes knowledge about spatial arrangements of scene objects, and a
camera model for $W$. Combining the prior model $P(Z)P(W)$
with the data model $P(X_{\mathcal Q}|Y_{\mathcal Q})$ then allows us
to develop inference methods based on (samples from) the posterior
$P(Z,W|X_{\mathcal Q})$. While the specific form of these models naturally depends on the application (see section \ref{sec:prior} for a description of  these models for our applications to tables scenes), the information pursuit strategy is generally applicable to any prior and data models, as explained next.

\subsection{Information Pursuit}
\label{sec:info.pursuit}

Let $(q_1,\dots,q_k)$ be an ordered sequence of the first $k$ distinct queries
and let $(x_1,\ldots,x_k)$ be possible answers from the corresponding classifiers
$(X_{q_1},\ldots,X_{q_k})$.  Consider the event
\begin{equation}
\mathbf E_k = \{X_{q_1}=x_1,\ldots,X_{q_k}=x_k\},
\end{equation}
where, $q_{\ell}$ is the index of the query at step $\ell$ of the process
and $x_{\ell}$ is the observed result of applying classifier $X_{\ell}$ on
$I$. Therefore, $\mathbf E_k$ is the accumulated evidence after $k$ queries.

The IP strategy is defined recursively.  The first query is fixed
by the model:
\begin{equation}
q_1=\argmax_{q \in \mathcal Q} \hspace{1mm} \mathcal{I}(X_{q},Y_{\mathcal Q}),
\end{equation}
where $\mathcal I$ is the mutual information, which is determined by the joint distribution of $X_{q}$ and $Y_{\mathcal Q}$.
Thereafter, for $k>1$,
\begin{align}
\label{EPdecision2}
q_k=\argmax_{q \in \mathcal Q} \hspace{1mm} \mathcal{I}(X_{q},Y_{\mathcal Q}| \mathbf E_{k-1})
\end{align}
which is determined by the {\it conditional} joint distribution of
$X_{q}$ and $Y_{\mathcal Q}$ given the evidence to date,
i.e., given $\mathbf E_{k-1}$. According to~\eqref{EPdecision2} a classifier with maximum expected information gain given the currently collected evidence is greedily selected at each step of IP.

From the definition of the mutual information, we have
\begin{equation}
\mathcal{I}(X_{q},Y_{\mathcal Q}| \mathbf E_{k-1}) =
H(Y_{\mathcal Q}| \mathbf E_{k-1}) - {H}(Y_{\mathcal Q}| X_{q}, \mathbf E_{k-1}),
\end{equation}
where $H$ denotes the Shannon entropy.  Since the first term on the
right-hand side does not depend on $q$, one sees that the next query
is chosen such that adding to the evidence the result of
applying $X_{q}$ to the test image will minimize, on average, the uncertainty
about $Y_{\mathcal Q}$. One point of caution regarding the
notation ${H}(Y_{\mathcal Q}| X_{q}, \mathbf E_{k-1})$: here
$Y_{\mathcal Q}$ and $X_{q}$ are random variables, while $\mathbf
E_{k-1}$ is a fixed event. The notation then refers to the {\it
conditional entropy} of $Y_{\mathcal Q}$ given $X_{q}$ computed under
the {\it conditional probability} $P(\cdot | \mathbf E_{k-1})$, i.e.,
the expectation (with respect to the distribution of $X_{q}$) of the
entropy of $Y_{\mathcal Q}$ under $P(\cdot | X_{q} = x, \mathbf
E_{k-1})$.

Returning to the interpretation of the selection criterion, we can
also write
\begin{equation}
\label{EPdecision4}
\mathcal{I}(X_q, Y_{\mathcal Q}|  \mathbf E_{k-1}) =\\
 H(X_{q}| \mathbf E_{k-1}) - {H}(X_{q}|Y_{\mathcal Q},  \mathbf E_{k-1}).
\end{equation}
This implies that the next question is selected such that:
\begin{enumerate}
\item $H(X_{q}| \mathbf E_{k-1})$ is large, so that its
answer is as unpredictable as possible given the current evidence, and
\item ${H}(X_{q}|Y_{\mathcal Q}, \mathbf E_{k-1})$ is
small, so that $X_q$ is predictable given the ground truth (i.e.,
$X_q$ is a ``good'' classifier).
\end{enumerate}
The two criteria are however
balanced, so that one could accept a relatively poor classifier if it
is (currently) highly unpredictable.

Depending on the structure of the joint distribution of $X$ and $Y$, these conditional entropies may not be easy to compute. A possible simplification is to make the approximation of neglecting the error rates of $X_{q}$ at the selection stage, therefore replacing $X_{q}$ by $Y_{q}$. Such an approximation leads to a simpler definition of $q_k$, namely
\begin{equation}
\label{EPdecision3}
q_k = \argmax_{q \in \mathcal Q\setminus \{q_1, \ldots, q_{k-1}\}} H(Y_{q}| \mathbf E_{k-1}).
\end{equation}
Notice that (in above) the $X$ and $Y$ are not assumed to coincide in the
conditioning event $\mathbf E_{k-1}$ (which
depends on the $X$ variables) so that the accuracy of the classifiers
is still accounted for when evaluating the implications of current
evidence. So here again, one prefers asking questions whose (true)
answers are unpredictable. For example, one would not ask ``Is it an
urban scene?'' after already having got a positive response to ``Is
there a skyscraper?'' nor would one ask if there is an object instance
from category $c$ in patch ``$A$'' if we already know it is highly
likely that there is an object instance from category $c$ in patch
``$B$'', a subset of ``$A$''. Removing previous questions from
the search is important with this approximation, since the mutual
information in \eqref{EPdecision4} vanishes in that case, but not
necessarily the conditional entropy in \eqref{EPdecision3}.

Returning to the general situation, \eqref{EPdecision4} can be simplified
if one makes two independence assumptions:
\begin{enumerate}
\item \label{ass:condindep1}
The classifiers are conditionally
independent given $Y_{\mathcal Q}$;
\item \label{ass:condindep2}
The classifier $X_q$ is conditionally
independent of $Y_{{\mathcal Q} \setminus q}$ given $Y_q$, i.e.,
the distribution of $X_q$ depends on $Y_{\mathcal Q}$ only through
$Y_q$.
\end{enumerate}
Clearly ${H}(X_q|Y_{\mathcal Q},  \mathbf E_{k-1})=0$ if
query $q$ belongs to the history, so assume $q \not\in \{q_1, \ldots, q_{k-1}\}$.
In what follows, let $y=(y_q, q \in \mathcal Q)$, where $y_q$ represents
a possible value of $Y_q$.  Then, under assumptions \ref{ass:condindep1} and \ref{ass:condindep2}, and using
the fact that $\mathbf E_{k-1}$ only depends on
the realizations of $X$, we have:
\begin{equation}
\begin{split}
&  {H}(X_q|Y_{\mathcal Q},  \mathbf E_{k-1})\\
& =  \sum_y {H}(X_q|Y_{\mathcal Q}=y,  \mathbf E_{k-1})P(Y_{\mathcal Q}=y|\mathbf E_{k-1})\\
& =  \sum_y {H}(X_q|Y_{\mathcal Q}=y)P(Y_{\mathcal Q}=y|\mathbf E_{k-1})\\
& =  \sum_y {H}(X_q|Y_q=y_q)P(Y_{\mathcal Q}=y|\mathbf E_{k-1})\\
& =  \sum_{y_q} {H}(X_q|Y_q=y_q)P(Y_q=y_q|\mathbf E_{k-1}).
\end{split}
\end{equation}
This entropy ${H}(X_q|Y_q=y_q)$ can be computed from the data model
and the mixture weights $P(Y_q=y_q|\mathbf E_{k-1})$
can be estimated from Monte Carlo simulations (see section~\ref{Conditional_Sampling}).
Similarly, the first term in \eqref{EPdecision4}, namely $H(X_{q}| \mathbf
E_{k-1})$, can be expressed as the entropy of a mixture:
\begin{align}
\begin{split}
&{H}(X_q|\mathbf E_{k-1})\\
& =  - \sum_x P(X_q=x|\mathbf E_{k-1})\log P(X_q=x|\mathbf E_{k-1})
\end{split}
\end{align}
with
\begin{align}
\begin{split}
&P(X_q=x|\mathbf E_{k-1})\\
& = -\!  \sum_y \! P(X_q=x|Y_{\mathcal Q}=y,\mathbf E_{k-1})P(Y_{\mathcal Q}=y|\mathbf E_{k-1}). \!\!
\end{split}
\end{align}
Arguing as with the second term in \eqref{EPdecision4}, i.e.,
replacing $P(X_q=x|Y_{\mathcal Q}=y,\mathbf E_{k-1})$ by $P(X_q=x|Y_q=y_q)$,
the last expression is the entropy
of the mixture distribution
\begin{equation}
\sum_{y_q} P(X_q=x|Y_q=y_q)P(Y_q=y_q|\mathbf E_{k-1}).
\end{equation}
where $x$ is fixed. Consequently, given an explicit data model, the information pursuit strategy
can be efficiently approximated by sampling from the posterior distribution.

As a final note, we remark that we have used the variables
$Y_{\mathcal Q}$ to represent the unknown scene $Z$. Writing
\begin{equation}
H(Z|\mathbf E_{k-1}) = H(Z|Y_{\mathcal Q}, \mathbf E_{k-1})
+ H(Y_{\mathcal Q}| \mathbf E_{k-1}),
\end{equation}
we see that the residual uncertainty on $Z$ given the current evidence
will only slightly differ from the residual uncertainty of
$Y_{\mathcal Q}$ as soon as the residual uncertainty of $Z$ given
$Y_{\mathcal Q}$ is small,
which is a reasonable assumption when the number of annobits is large enough.

We now pass to a more specific description of the variables $X,Y,Z$
and their distributions.  In particular, the next section provides our
driving principles for the choice of the
annobits. We will then
discuss the related classifiers, followed by the construction of the
prior and data models, their training and the associated sampling
algorithms.

\section{Annobits}
\label{sec:foundation}

\subsection{General Principles}

The choice of the functions $f_q$ that define the
annobits, $Y_q = f_q(Z,W)$, $q\in\mathcal Q$, naturally depends on the specific application.
The annobits we have in mind for scene interpretation, and have used in previous related
work on a visual Turing test \citep{geman2015visual}, fall mainly into three categories:
\begin{itemize}
\item {\bf Scene context annobits:} These indicate full scene labels, such as ``indoor'', ``outdoor'' or ``street"; since our application is focused entirely on
``dinning room table settings'' we do not illustrate these.
\item {\bf Part-of descriptors:}  These indicate whether or not one image
region is a subset of another, e.g., whether an image patch is part of a table.
\item {\bf Existence annobits:} These relate to the presence or absence
of object instances with certain properties or attributes.  The most numerous set
of annobits in our system ask whether or not instances of a given object category
are visible inside a specified region.
\end{itemize}
Functions of these elementary descriptors can also be of interest. For example, we
will rely heavily on annobits providing a list of
all object categories visible in a given image region, as described in section \ref{sec:extended_annobits}.

\subsection{Annocell Hierarchy}
\label{sec:annocell.hierarchy}

Recall from section \ref{sec:scenes_queries} that a scene description $Z$ consists of the object categories and 3D poses of visible instances from a pre-determined family of object categories. Here, motivated by our application to dining room table scenes where objects lie in the table plane, we use a 2D representation of the object pose, which can be put in one-to-one correspondence with its 3D pose via the homography relating the image plane and the table plane (see section \ref{sec:camera_model} for details). More specifically, an object instance is a triple $(C, L, D)$, where $C\in\mathcal C$ denotes the object category in a set of pre-defined categories $\mathcal C$, $L\in\mathcal L$ denotes the locations of the centers of the instances in the image domain $\mathcal L$ and $D>0$ denotes their sizes in the image (e.g., diameter). The apparent 2D pose space is therefore $\mathcal L \times (0, +\infty)$. More refined poses could obviously be considered.

\begin{figure}
\centering
\includegraphics[width=\linewidth]{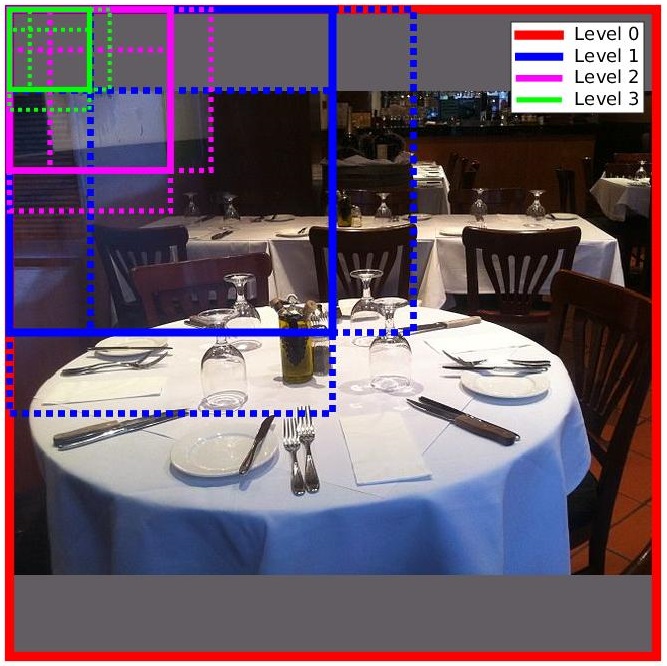}
\caption[Annocell Visualization]{
Some selected cells from different levels of the annocell
hierarchy. Rectangles with dashed lines are the nearest neighbor
patches to the rectangles with solid lines from the same color.}
\label{fig::Patch_Hierarchy}
\end{figure}

To define the queries, we divide the apparent pose space into cells. Specifically, we consider a finite, distinguished subset of sub-windows, $\mathcal A$, and subset of size intervals, $\mathcal M$, and index the queries $q\in\mathcal Q$ by the triplet $q = (C,A,M)$, where $C\in\mathcal C$, $A \in \mathcal A$, and $M \in \mathcal M$. For every category $C \in \mathcal C$, sub-window $A \subset \mathcal A$ and size interval $M \in \mathcal M$, we let $Y_{C,A,M}=1$ if an instance of category $C$ with size in $M$ is
visible in $A$, and $Y_{C,A,M}=0$ otherwise. If $M=(0, +\infty)$, we
simply write $Y_{C,A}$. We refer to $A\in\mathcal A$ as an
``annocell.''  Specifically, assuming $\mathcal L=[0,1]^2$ (by padding and normalizing), $\mathcal A$ consists of square patches of four sizes, $2^{-l}$ for $l \in \{0,1,2,3\}$.  The patches at each ``level'' overlap: for each level, the row and column
shift ratio is $25\%$ \ie $75\%$ overlap between nearest windows. This
leads to 1, 25, 169, and 841 patches for levels 0,1,2, and 3
respectively, for a total of $|\mathcal A|=1036$ patches.
Figure~\ref{fig::Patch_Hierarchy} shows some of these regions selected
from the four levels of the hierarchy.

Using a hierarchical annocell structure has the advantage of allowing
for coarse-to-fine exploration of the pose space. Note also that, by
construction, annocells at low resolution are unions of certain
high-resolution ones. This implies that the value of the annobits at
low resolution can in turn be derived as maximums of high-resolution
annobits.

\subsection{Extended Existence Annobits}
\label{sec:extended_annobits}
Due to the nature of the classifiers we use in our
application, we also introduce annobits that list the
categories that have entirely visible instances in an annocell, i.e.,
the collection
\begin{equation}
Y^{\c}_A = (Y_{C, A}, C\in \mathcal C).
\end{equation}
In addition, we also use category-independent, size-related annobits:
For each annocell $A \!\in\! {\mathcal A}$ and size interval $M \!\in\! {\mathcal M}$,
we define a binary annobit $Y^{\s}_{A,M}$ which indicates whether or not the average
size of the objects present in $A$ belongs~to~$M$.

\subsection{Classifiers for Annobits}

The particular image-based predictors of the annobits we use
in the table-setting application are described in full detail in section \ref{sec:CNNs}. Some examples include:
\begin{itemize}
\item Variables $X^\c_A$, $A\in \mathcal A$, which provide a vector of weights on $\mathcal C$ for predicting $Y^\c_A$.
\item Variables $X^\s_{A}$, $A\in \mathcal A$, which provide a probability vector on $\mathcal M$ for predicting $(Y^{\s}_{A,M}, M \in {\mathcal M})$.
\end{itemize}
Additional variables $X^t_A, A\in\mathcal A'$ (where $\mathcal A'$ is
a subset of $\mathcal A$) will also be introduced. They are designed
to predict information units $Y_A^t=1$ if more than half of $A$ overlaps
the table. Observe that the classifier $X_q$ assigned to $Y_q$ does not necessarily assume the same value as $Y_q$. However, this is not a problem since we are only interested in the conditional distribution of $X$ given $Y$.

\section{Prior Model}
\label{sec:prior}

Following section \ref{sec:approach}, the joint distribution of the
annobits $(Y_q, q\in\mathcal Q)$ is derived from a prior model on the
3D scene description, $Z$, and on camera parameters $W$. We
assume these variables to be independent and model them separately.

\subsection{Scene Model $P(Z \mid S)$}
\label{sec:prior.scene}

Motivated by our application to dining room table scenes, we assume a
fixed dominant plane in the 3D model, and choose a coordinate system $Oxyz$
in $\mathbb R^3$, such that the xy-plane coincides with this dominant plane.
The scene $Z$ is represented as a set of object instances, assumed to
be sitting on a bounded region of the dominant plane, in our case a
centered, rectangular table $S$ characterized by its length and
width. Recall from section \ref{sec:annocell.hierarchy} that
each object instance $i$ is represented by a category  $C_i \in
\mathcal C$, a location $L_i$ and a size $D_i$ in the image.
Here, we assume that objects from a given category have a fixed size,
so that $Z=\{Z_i\}$ with $Z_i = (C_i,L_i)$.  The distribution
of $Z$ will be defined conditional to $S$, since, for example, the
size of $S$ will directly impact the number of objects that it can
support. More generally the table can be replaced by some other
variable $S$ representing more complex properties of the global scene geometry. For convenience we sometimes drop $S$ from our notation. However, most of the model components introduced below depend on $S$, and the proposed model is to be understood conditional to $S$.

We partition the reference plane into small cells ($5\text{cm} \times
5\text{cm}$ in the table-setting case) and use binary variables to indicate
the presence of instances of object categories centered in each
cell. In other words, we discretize the family $(C_i, L_i)$ into a
binary random field that we will still denote by $Z$.  Letting
$\mathcal J$ denote the set of cells, a configuration can therefore be
represented as the binary vector $z = (z_{j,c}, j\in \mathcal J, c\in
\mathcal C)$ where $z_{j,c} = 1$ if and only if an object of category
$c$ is centered in the cell $j$.

\begin{figure}
\centering
\includegraphics[width=\linewidth]{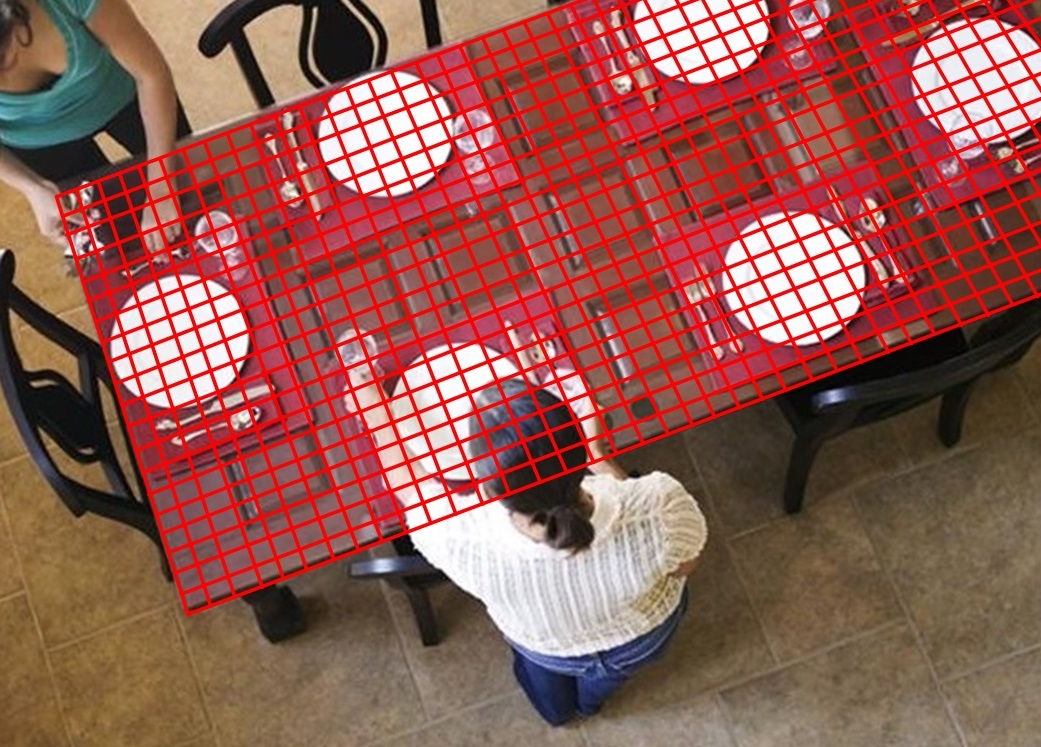}
\caption[Table fitting mesh.]{Table fitting mesh.}
\label{fig:tablemesh}
\end{figure}

The configuration $z$ is obviously a discrete representation of the scene layout restricted to object categories $\mathcal{C}$ and location $L$. Letting $\Omega$ denote the space of all such
configurations, we will use a Gibbs distribution on $\Omega$
associated with a family of feature functions $\boldsymbol \varphi =
(\varphi_i, i=1, \ldots, n)$, with $\varphi_i :
\Omega \mapsto \{0,1\}$, and scalar parameters $\boldsymbol \lambda =
(\lambda_i, i=1, \ldots, n)$.  The Gibbs distribution then has
the following form:
\begin{align}
\label{sect:diss1}
p(z) = \frac{1}{\kappa(\boldsymbol \lambda)}\exp\bigl(\boldsymbol \lambda\cdot\boldsymbol{\varphi}(z)\bigr),
\end{align}
where $\kappa(\boldsymbol \lambda)$ is the normalizing factor (partition function)
ensuring that the probabilities sum up to
one. Figure~\ref{fig:tablemesh} shows a table and its fitted mesh
where each of the cells is a $5\text{cm} \times 5\text{cm}$ square.

\begin{figure}
\centering
\includegraphics[width=0.96\linewidth]{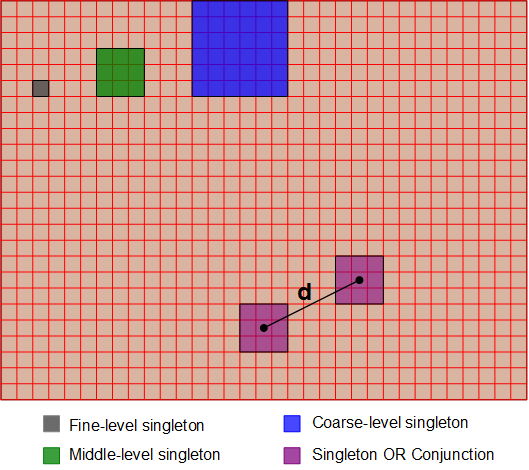}
\caption[Feature functions' domain]{Domain of various types of feature functions.}
\label{fig:features}
\end{figure}

We use the following features:
\begin{itemize}
\item
Existence features, which indicate whether or not an instance from
a given category is centered anywhere in a given set of cells, therefore taking the
form
\begin{equation}
\varphi_{J, c}(z) = \max(z_{j,c}, j\in J)
\end{equation}
with $J\subset \mathcal J$. We consider sets $J$ at three different
granularity levels, illustrated in Figure~\ref{fig:features}. At the
fine level $J= \{j\}$ is a singleton, so that $\varphi_{J,c}(z) =
z_{j,c}$. We also consider middle-level sets (3$\times$3 array of fine
cells) and coarse-level sets (6$\times$6 array of fine cells) that cover the reference
plane without intersection.
\item Conjunction features,  which are products of two middle-level existence features (of the same or different categories), and therefore signal their co-occurrence:
\begin{equation}
\varphi_{J_1,c_1, J_2, c_2}(z) = \varphi_{J_1,c_1}(z)\varphi_{J_2,c_2}(z).
\end{equation}
To limit model complexity, only pairs $J_1, J_2$ whose centers are less than a
threshold away are considered where the threshold can depend on the pair
$c_1,c_2$.
\end{itemize}

Invariance and symmetry assumptions about the 3D scene are then
encoded as equality constraints among the model parameters thereby
reducing model complexity.  Grouping binary features $\varphi_i$ with
identical parameters $\lambda_i$ is then equivalent to considering a new set of features that count the number of layout configurations satisfying some conditions on the locations and categories. For table settings, it is natural to assume invariance by
rotation around the center of the table.  Hence we assume that
existence features whose domain $J$ is of the same size and located at
the same distance from the closest table edge all have the same
weights ($\lambda$'s), and hence the probability only depends on the
number of such instances.

We group conjunction feature functions based on the distance of the first patch to the edge of the table, and the relative position of the second patch (left, right, front, or back) with respect to the first patch.

\begin{remark}:
The model can be generalized to include pose attributes other
than location, e.g., orientation, size and height. If $\Theta$
denotes the space of poses, then
one can extend the state space for $z_{j,c}$ to
$\{0,1\}\times \Theta$, interpreting $z_{j,c} = (1, \theta)$ as the
presence of an object with category $c$ and pose $\theta$ in cell $j$,
and $z_{j,c} = (0, \theta)$ as the absence of any object with category
$c$, $\theta$ being irrelevant. Features can then be extended to this
state space to provide a joint distribution that includes pose. The
simplest approach would be to only extend univariate features,
so that object poses and other attributes are conditionally
independent given their categories and locations (and the geometry
variable $S$, since the model is always assumed conditional to it).
Other attributes (color, style, etc.) can be incorporated
in a similar way.
\end{remark}

\subsection{Camera Model $P(W)$}
\label{sec:camera_model}

The second component of the prior model determines the probability
distribution of the extrinsic and intrinsic camera parameters, such as its pose and focal length, respectively. The definition of these parameters is fairly standard in computer vision (see e.g., \cite{MASKS03}), but the definition of generative models for these parameters is not. In what follows we summarize the typical definitions, and leave the details of the generative model to the Appendix.

Remember that we assumed a fixed coordinate system in 3D in
which the $xy$-plane coincides with the dominant ``horizontal''
plane. Consider also a second camera coordinate system
$O'x'y'z'$, such that $x'y'$-plane is equal to the image plane. The extrinsic camera parameters are defined by the pose $(R,T)$ of the camera coordinate system $O'x'y'z'$ relative to the fixed coordinate system $Oxyz$, where $R$ is the camera rotation, which maps the unit axis vectors of $Oxyz$ to the unit axis vectors of $O'x'y'z'$, and  $T = OO'$ is the translation vector. We parametrize the rotation $R$ by three angles
$\boldsymbol\psi = (\psi_x, \psi_y, \psi_z)$ representing, respectively,
counter-clockwise rotations of the camera's coordinate system about
the x-axis, y-axis, and z-axis of the world coordinate system (see
equation \eqref{eq:angles} for conversion of unit vectors to angles). Observe that one can
express the coordinates $m = (x,y,z)^\top$ of a 3D point in the world
coordinate system as functions of its coordinates in the camera coordinate system $m' = (x',y',z')^\top$  in the form $m = Rm'+ T$. Since in our case 3D points lie in a plane $N^\top m' = d$, where $N$ is the normal to the plane (\ie table) measured in the camera coordinate system and $d$ is the distance from the plane to the camera center, we further have $m = H m'$, where $H = (R + T N^\top/d)$ is the homography between the camera plane and the world plane.

The intrinsic camera parameters are defined by the coordinates of the focal point, $(x_0, y_0,-f)$, where $f>0$ is the focal length and $(x_0, y_0)$ is the intersection of the principal axis of the camera with the image plane, as well as the pixel sizes in directions $x'$ and $y'$, denoted by $\gamma_x$ and $\gamma_y$.

The complete set of camera parameters is therefore 11-dimensional and
given by $W = (f, \gamma_x, \gamma_y, x_0, y_0, \boldsymbol \psi,
T)$. Our generative model for $W$ assumes that:
\begin{itemize}
\item Intrinsic camera parameters are independent from extrinsic camera parameters.

\item Pixels are square, i.e., $\gamma_x = \gamma_y$, but intrinsic parameters are otherwise independent. The focal length $f$ is uniformly distributed between 10 and 40 millimeters, $x_0$ (resp. $y_0$) is uniformly distributed between $W_p/4$ and $3W_p/4$ (resp. $H_p/4$ and $3H_p/4$), where $W_p$ and $H_p$ are the width and height of the image in pixels, and $\gamma_x=\gamma_y$ is uniformly distributed between $1/W_p$ and $1.2/W_p$.

\item The vertical component of $T$ is independent of the other two and the
distribution of the horizontal components is rotation invariant. Specifically, letting $T = (T_x,T_y, T_z)$, we assume that $(T_z-0.3)/2.7$ follows a Beta distribution so that $T_z\in [0.3, 3]$ (expressed in meters). Then, letting $r = \sqrt{T_x^2+T_y^2}$ denote the distance between the horizontal projection of $T$ on the table plane and the center of the table, we assume that $r/4$ follows a Beta distribution. We assume independence of $r$ and $t_z$ and invariance by rotation around the vertical axis, which specifies the distribution of $T$.

\item The distribution of the rotation angles $\mathbf \psi$ is defined conditionally to $T$. Specifically, we assume that the camera roughly points towards the center of the scene and the horizontal direction in the image plane is also horizontal in the 3D coordinate system. Additional details of the model for $p(\psi | T)$ are provided in the Appendix.
\end{itemize}

\subsection{Scene Geometry Model $P(S)$ and Global Model}

We assume that the scene geometry $S$ takes value in a finite set of
``template geometries'' that coarsely cover all possible
situations. Note that these templates are defined up
to translation, since we can always assume that the 3D reference frame
is placed in a given position relative to the geometry. For table
settings, where the geometry represents the table itself, our templates were simply square tables with size distributed according to a shifted and scaled Beta distribution ranging from 0.5 to 3 meters. This rough approximation was sufficient for our purposes, even though tables in real scenes are obviously much more variable in shape and size.\\

Finally, the joint prior distribution $p(z,s,w) = P(Z=z,S=s,W=w)$ of all the variables is defined by:
\begin{align}
\label{ImageJointDist}
p(z,s,w) =  p(z |s) ~ p(s) ~ p(w).
\end{align}

\subsection{Learning the Prior Model}
\label{ModelLearning}

The models for $P(S)$ and $P(W)$ are simple enough that we specified their model parameters manually, as described before. Therefore, the fundamental challenge is to learn the prior model on scene interpretations $P(Z \mid S)$. For this purpose, we assume that a training set of annotated images is available. The annotation for each image consists of a list of
object instances, each one labeled by its category (and possibly other attributes) and apparent 2D pose represented by an ellipse in the image plane. We also assume that sufficient information is provided to propagate the image annotation to a scene annotation in 3D coordinates; this will allow us to train the scene model independently from the unknown
transformation that maps it to the image. This can be done in several
ways. For example, given four points in the image that are the projections of the corners of a square
in the reference plane, one can reconstruct, up to a scale factor, the
homography mapping this plane to the image. Doing this with a
reasonable accuracy is relatively easy in general for a human
annotator, and allows one to invert the outline of every flat object
on the image that lies on the reference plane to its 3D shape, up to
a scale ambiguity. This ambiguity can be removed by knowing the true
distance between two points in the reference plane, and their
positions in the image. We used this level of annotation and representation for our table settings, based on the fact that all objects of interest were either horizontal (e.g., plates), or had easily identifiable horizontal components (e.g., bottoms of bottles), and we assumed that plates had a standard diameter of 25cm to remove the scale ambiguity.

As can be seen, the level of annotation required to train our prior model is quite high. While we have been able to produce rich annotations for 3,000 images of dining room table settings (see section \ref{sec:dataset}), this is insufficient to train our model. To address this issue, in the next section we propose a 3D scene generation model that can be use to generate a large number of annotations for as many synthetic images as needed. Given the annotations of both synthetic images (section \ref{sec:scene_generation}) as well as real images (section \ref{sec:dataset}), the parameters of our prior model are learned using an accelerated version of the robust stochastic approximation~\citep{Nemirovski09} to match empirical statistics calculated based on top-down samples from the scene generation model (see~\cite{EJ_Dissertation2016} for details).

\section{Scene Generation Model}
\label{sec:scene_generation}
In this section we propose a 3D scene generation model that can be used to generate a large number of annotations to train the prior model described in the section \ref{sec:prior}. The proposed model
mimics a natural sequence of steps in composing a
scene. First, create spontaneous instances by
placing some objects randomly in the scene; the distribution
of locations depends on the scene geometry. Then, allow
each of these instances to trigger the placement of ancillary objects,
whose categories and attributes are sampled conditionally, creating
groups of contextually related objects. This recursive process
terminates when no children are created, or when the number of
iterations reaches an upper-bound.

\subsection{Model Description Using a Generative Attributed Graph}

To formally define this process, we will use the notation $\mathbf n =
(n_c, c\in \mathcal C)$ to represent a family of integer counts $n_c
\in \mathbb N$ indexed by categories, so that $\mathbf n \in \mathbb
N^{|\mathcal C|}$. We will also let $|\mathbf n| = \sum_{c\in \mathcal
C} n_c$.

\begin{figure}[t]
\centering
\begin{tikzpicture}[->,>=stealth',shorten >=1pt,auto,node distance=3cm,
                    thick,main node/.style={circle,fill=gray!50,font=\sffamily\footnotesize}]

        \node[main node] (1) at (0, -0.5) {plate};
        \node[main node] (2) at +(350: 2.5) {bottle};
        \node[main node] (3) at +(300: 2.5) {glass};
        \node[main node] (4) at +(240: 2.5) {utensil};

        \path[every node/.style={font=\sffamily\small}]
             (1) edge (3)
                 edge (4)
             (2) edge (3)
             (4) edge [loop left] (4);
        \end{tikzpicture}
        \caption[Master graph]{An example master graph.}
        \label{fig::mastergraph}
\end{figure}
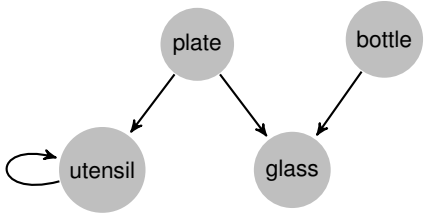

We will assume a probability distribution $p^{(0)}$ on $\mathbb
N^{|\mathcal C|}$, and a family of such distributions $p^{(c)}, c\in
\mathcal C$. These distributions (which are defined conditionally to
$S=s$) are used to decide the number of objects that will be placed in
the scene at each step.  More specifically:
\begin{enumerate}
\item
$p^{(0)}(\cdot \mid  s)$ is the conditional joint distribution of the
number of
object instances from
each category that are placed initially on the
scene.
\item
For each category $c\in \mathcal C$, $p^{(c)}(\cdot \mid s)$ is the joint
distribution of the numbers of new object instances that are triggered by the
addition of an object instance from category $c$.  These distributions can be
thought of as the basis distributions in a multi-type branching
process (see~\cite{Mode71}).
\end{enumerate}
The complexity of the process is controlled by a master graph that
restricts the subset of categories that can be created at each
step. More formally, this directed graph has vertices in $\{0\}\cup
\mathcal C$ and is such that $p^{(v)}$ is supported by categories that
are children of the node $v\in \{0\} \cup \mathcal C$. Adjoining $0$ to
the node labels avoids treating $p^{(0)}$ as a special case in the
derivations below. The master graph we
used on table settings is provided in Figure \ref{fig::mastergraph},
where we regard ``plate'' and ``bottle'' as the children of category
$0$. Note that since we allow spontaneous instances from all categories every category is a child to category 0.

\begin{figure}
\centering
\includegraphics[width=0.95\linewidth]{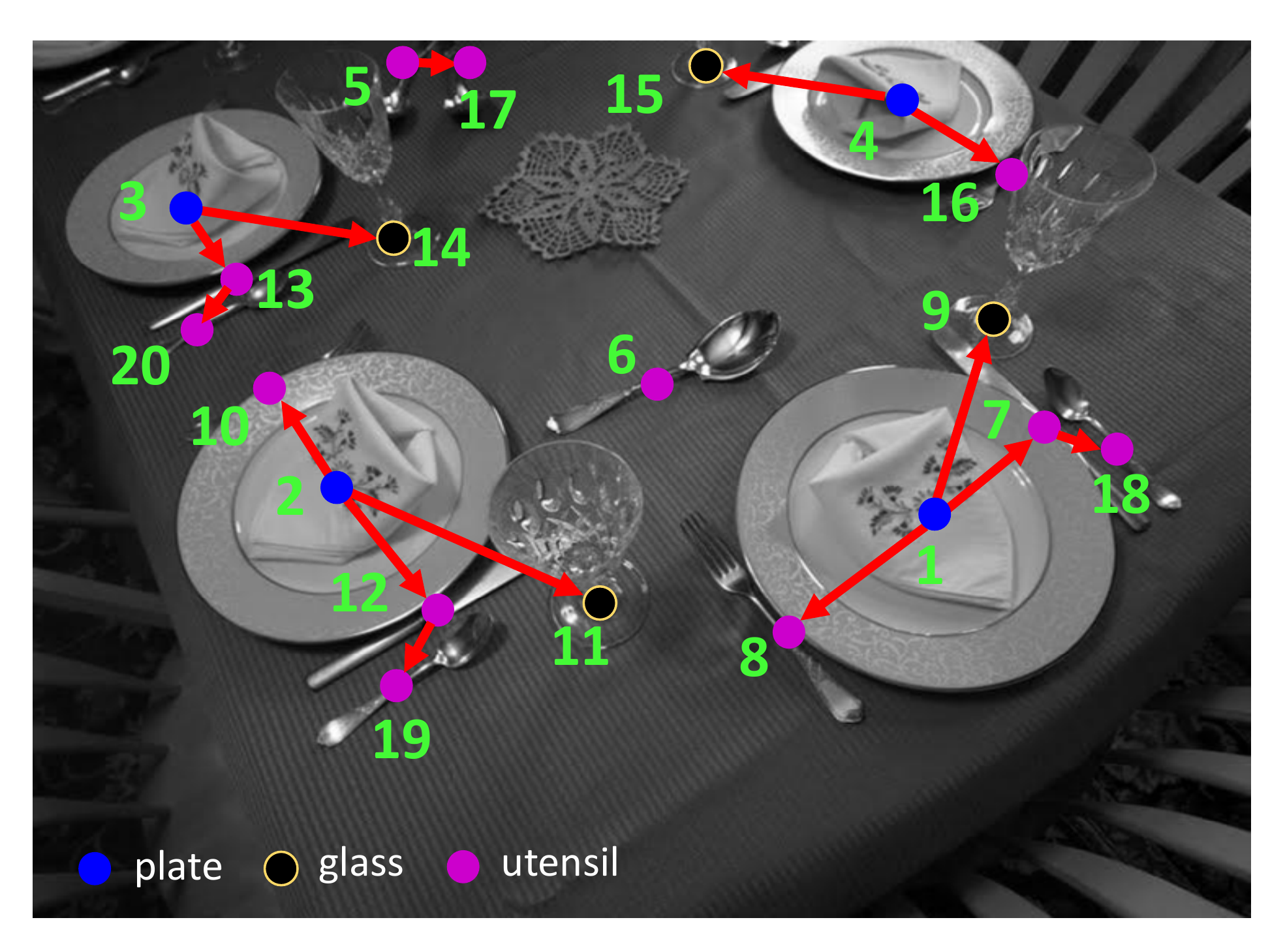}\\
\includegraphics[width=0.85\linewidth]{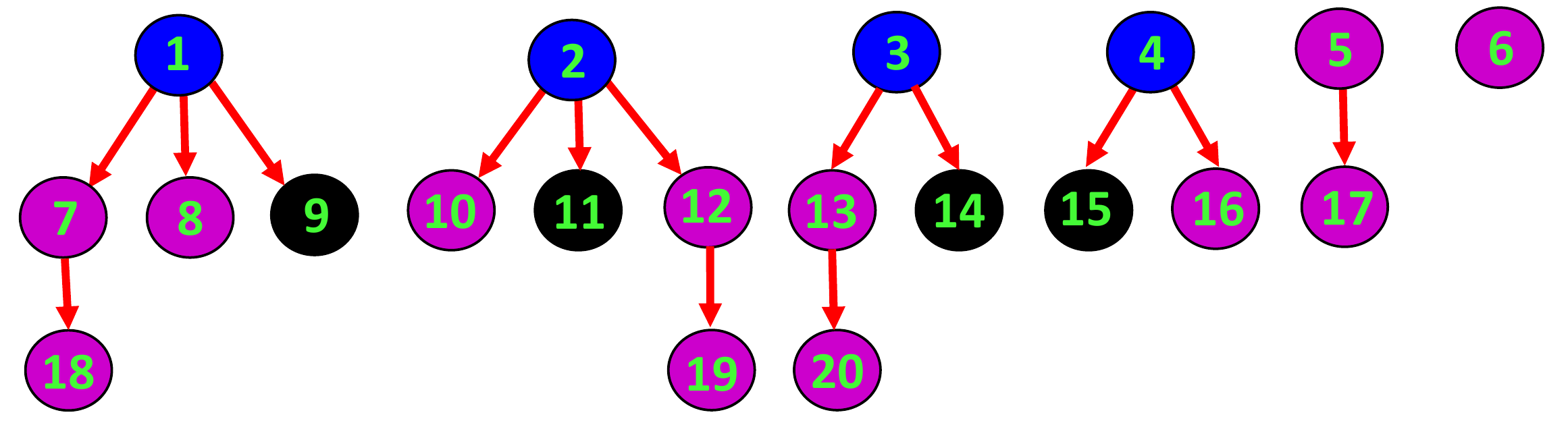}\\
\caption[An example category-labeled attributed graph]{A table-setting scene (top) and its corresponding skeleton
graph (bottom) where the categories (plate, bottle, glass, and utensil)
are color-coded in the graph. Root nodes
$V_0$ initialize the generative process; here there are six.
The terminal nodes for this instance are $V_T =
\{6,8,9,10,11,14,15,16,17,18,19,20\}$. According to the base graph
$n^{(0)}_{\text{plate}}= 4$,$n^{(0)}_{\text{bottle}}= 0$,
$n^{(0)}_{\text{glass}}= 0$ and $n^{(0)}_{\text{utensil}}= 2$.}
\label{fig:BaseGraphPic}
\end{figure}

The output of this branching process can be represented as a directed
tree $G_0 = (V,C,E)$ in which each vertex $v\in V$ is attributed a
category denoted by $C(v)$ and $E$ is a set of edges. The root node of
the tree, hereafter denoted by $0$, essentially represents the empty
scene whose ``category'' is also denoted by 0 (note that $0\not \in \mathcal
C$). All other nodes have categories in $\mathcal C$. Each
non-terminal node $v\in V$ has $|\mathbf N^{(v)}|$ children where
$\mathbf N^{(v)} \sim p^{(c(v))}(\cdot|s)$ so that $N_c^{(v)}$ of
these children have category $c$. We will refer to $G_0$ as a skeleton
tree, which needs to be completed with the object attributes (excluding its category since $G_0$ already includes the category attribute) to obtain
a complete scene description. The probability distribution of $G_0$ is
\begin{equation}
p(G_0 \mid s) = \prod_{v\in V \backslash V_T} p^{(c(v))}(\mathbf n^{(v)} \mid s),
\end{equation}
where $V_T$ is the set of terminal nodes and $\mathbf n^{(v)}$ are the category counts of the children of $v$
(graphs being identified up to category-invariant isomorphisms). An
example of such graph is provided in Figure \ref{fig:BaseGraphPic}.

To complete the description, we need to associate attributes to
objects, the most important of them being their poses in the 3D world,
on which we focus now. In the MRF designed for our experiments, the only relevant information about pose was the location on the table, a 2D parameter. It is however possible to design a top-down generative model that includes richer information, using for example a 3D
ellipsoid. Such representations
involve a small number of parameters denoted generically by
$\theta$: each vertex $v$ in the skeleton graph is attributed by parameters such as its pose denoted by $\theta^{(v)}$. When using ellipsoids, $\theta^{(v)}$
involves eight free parameters (five for the shape of the ellipsoid,
which is a positive definite symmetric matrix, and three for its
center).  Fewer parameters would be needed for flat objects (represented by a 2D ellipse), or vertical ones, or objects with rotational symmetry. In any case, it is obvious that the distribution of an object pose
depends heavily on its category.

In our model, contextual information is important: when placing an
object relative to a parent, the pose also depends on the parent's
pose and category. This is captured by the conditional
distribution $p^{(c)}(\theta \mid c', \theta')$ of the pose
parameters for a category $c$,
relative to a parent with category $c'$ and pose $\theta'$. To
simplify notation, we allow again for $c'=0$ (indicating objects
without parent), in which case $\theta'$ is irrelevant. The complete
attributed graph associated with the scene is now $G = (V,C,\Theta, E)$
(where $\Theta$ is the family of poses) with
distribution
\begin{multline}
\label{eq:graph.model}
p(G=g|s) = \prod_{v\in V \backslash V_T} p^{(c(v))}(\mathbf n^{(v)}|s) \\
\prod_{v\in V\setminus\{0\}} p^{(c(v))}(\theta^{(v)}|c(pa(v)), \theta_{pa(v)}),
\end{multline}
where $pa(v)$ is the parent of $v$. In \eqref{eq:graph.model}, we have mixed discrete probability mass functions for the object counts and
continuous probability density functions for the pose attributes.

If one is only interested in the objects visible in the scene, the
scene description, $Z$, is obtained by discarding the graph structure
from $G$, i.e., only retaining the object categories and
poses. More complex scene descriptors could be interesting as well,
like object relationships or groupings (e.g., whether a family of
plate, utensils, glasses can be considered as belonging to a single
setting), in which case the whole graph structure may also be of
interest; we do not use such ``compositions'' in our
experiments. As a final point, we mention that the samples may require
some pruning at the final stage, since the previous model does not
avoid object collisions or overlaps that one generally wants to avoid. We removed physically impossible samples in which vertical object categories (\ie bottle and glass) were overlapping in the world coordinate system. In general, one can add undirected edges between the children of the same parent to incorporate more context into a single setting. More details on the scene model that we used for table-settings can be found in the Appendix.

\subsection{Algorithm for Learning the Scene Generation Model}
Even though the annotation is assumed to describe the scene in the
world coordinate system, the information it provides on $G$ is still
incomplete, because it does not include the graph structure. To learn
the parameters of the branching process, we used the EM algorithm
\citep{dempster1977EM} or, more precisely, the Monte-Carlo version of
the {\em Stochastic Expectation-Maximization} (SEM)
algorithm \citep{celeux_diebolt85}, usually referred to as MCEM in the
literature \citep{WeiTanner90}. In this framework, the conditional
expectation of the complete log-likelihood, which is maximized at each
step to update the parameters, is approximated by Monte-Carlo
sampling, averaging a sufficient number of realizations of the
conditional distribution of the complete data given the observed one
for the current parameters. Note that the unobserved part of the graph
given $(V, C, \Theta)$ can be represented as a $|V|$-dimensional
vector $\boldsymbol{\zeta}=\bigl(\zeta_1=pa(v_1),
\zeta_2=pa(v_2),...,\zeta_{|V|}=pa(v_{|V|})\bigr)$, with $pa(v) =
\emptyset$ if $v$ is an orphan. These configurations form a subset of
$V\cup\{\emptyset\}$, given the constraints imposed by the master
graph and the fact that $g$ is acyclic. The Gibbs sampling algorithm
iteratively updates each $\zeta_i$ according to its conditional
distribution given the observed variables and the other $\zeta_j$,
$j\neq i$, which can easily be computed using equation
\eqref{eq:graph.model}. Recall that the graph distribution is learned
conditional to a given scene geometry $S=s$.

\subsection{Simulated Table Settings}

\begin{figure}
\begin{center}
\setlength{\fboxrule}{2pt}%
\fbox{\includegraphics[width=0.95\linewidth]{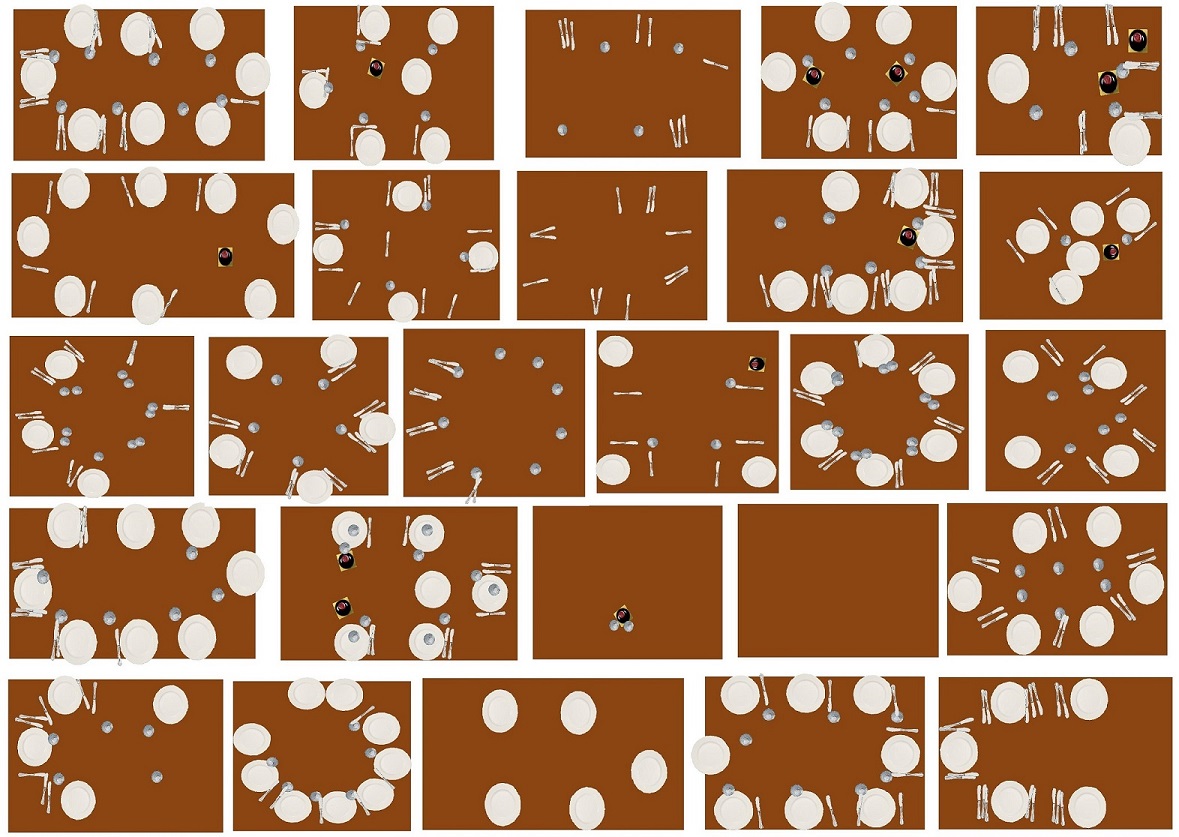}}\\
\vspace{2mm}
\fbox{\includegraphics[width=0.95\linewidth]{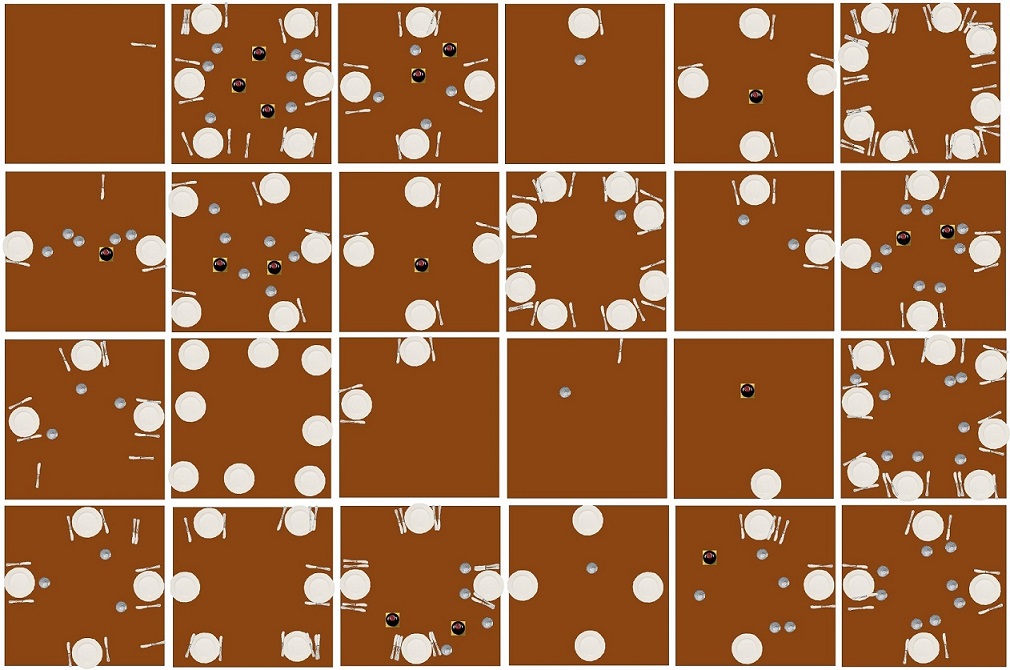}}
\end{center}
\vspace{-3mm}
\caption[Samples from the dataset vs samples from the random graph model]{\footnotesize{Top-view icon
visualization of table-settings considering only plate,
bottle, glass, utensil categories. {\bf Upper Panel}: visualization of
some annotated images in the dataset that roughly match in size to a
$1.5 \times 1.5 \hspace{0.5mm} m^2$ table
{\bf Lower Panel}:
samples from the generative attributed graph model for a square table
of size $1.5 \times 1.5 \hspace{0.5mm} m^2$.}}
\label{fig:samples}
\end{figure}

Figure~\ref{fig:samples} shows top-view visualization of some
annotated images in the dataset that roughly match in size to a $1.5
\times 1.5 \hspace{0.5mm} m^2$ table and some samples drawn from the
generative attributed graph model for a square table of size $1.5
\times 1.5 \hspace{0.5mm} m^2$ learned from matching annotated
images. Visual similarity of the samples taken from the generative
attributed graph model to natural scene samples confirm suitability of
this model for table setting scenes although the proposed model is
quite general and can be used to model different types of scenes.\\

\begin{remark}:
We developed algorithms for unconditional and conditional sampling of
the graph model in the context of IP (the conditional distribution
relative to the current history). The unconditional sampling is
top-down, easy and fast. However, our conditional sampling based on
Metropolis-Hastings (\cite{HASTINGS01041970, Metropolis53}) is
relatively complex and slow to adapt to a new condition \ie long
burn-in period; this is partly due to the innate low acceptance rate
of the Metropolis-Hastings algorithm, normally $< 25\%$
(see~\cite{Roberts2001c} and \cite{EJ_Dissertation2016} for
details). This is why we have not used this model directly in the IP
framework, relying instead on the MRF model described in section
\ref{sec:prior.scene}, in which the feature expectations are learned
on scenes generated by the generative attributed graph model.
\end{remark}

\section{Conditional Sampling}
\label{Conditional_Sampling}

Sampling from the posterior distribution over hidden variables given
evidence is central to our method, being necessary for both IP and
performance evaluation.  Writing $\Xi = (Z, S, W)$ for the unobserved
scene-related variables, the prior distribution $p(\xi)=p(z,s,w)$ was
given in \eqref{ImageJointDist}.  Recall that the annobits $Y_q$
are deterministically related to the scene, with $Y_q = f_q(Z, W)$.
In this discussion, we will work under the simplifying assumption that
the classifiers are conditionally independent given $\Xi$ and that,
for a given $q$, the conditional distribution of $X_q$ given these
variables only depends on $Y_q$. (This assumption can be relaxed to a
large extent without significantly increasing the complexity of the
algorithm. This will be discussed at the end of this section.)  Recall
also (see Section \ref{sec:info.pursuit}) that at step $k$ of IP, in order to compute the
conditional mutual information and determine the next query $q_k$, we
require the mixture weights $P(Y_q=y|\mathbf E_{k-1})$, where $\mathbf
E_{k-1} =\{X_{q_1}=x_1,...,X_{q_{k-1}}=x_{k-1}\}$ is the evidence
after $k-1$ steps. Clearly, then, $P(Y_q=y|\mathbf E_{k-1})$ can be
estimated from samples from $\Xi$ given the history.

The joint distribution of $\Xi$ and all the data $X_{\mathcal Q}$
therefore takes the form
\begin{equation}
\label{eq:complete.prob}
P(x_{\mathcal Q}, \xi) = p(z\mid s) p(w) p(s) \prod_{q\in\mathcal Q} p(x_q\mid f_q(z, w)).
\end{equation}
Since the next query $q_k$ is a deterministic function of $\mathbf
E_{k-1}$, the conditional distribution of $\Xi$ given $\mathbf
E_{k-1}$ is
\begin{equation}
\label{eq:cond.prob}
p(\xi\mid \mathbf E_{k-1}) \propto p(z\mid s) p(w) p(s) \prod_{q=1}^{k-1} p(x_q\mid f_q(z, w)),
\end{equation}
for which we have again used the conditional independence of the $X_q$'s given the scene.\\

\subsection{General Framework}
We use a Metropolis-Hastings sampling strategy to estimate the
conditional distribution of the scene variables given the history. As
a reminder, the algorithm relies on the fact that any transition
probability $\psi(\xi, \xi')$ can be modified by rejection sampling to
be placed in detailed balance with $p(\xi\mid \mathbf E_{k-1})$ by
letting
\begin{multline}
\label{eq:mh}
\psi^*(\xi, \xi') = \\
\begin{cases}
\displaystyle
\psi(\xi, \xi')\max\left(1, \frac{\psi(\xi', \xi)p(\xi'\mid \mathbf E_{k-1})}{\psi(\xi, \xi')p(\xi\mid \mathbf E_{k-1})}\right),\\
\displaystyle
\hskip 0.25\textwidth\text{ if } \xi'\neq \xi\\
\displaystyle
1 - \sum_{\xi'' \neq \xi} \psi^*(\xi, \xi''), \text{ if } \xi' = \xi
\end{cases}
\end{multline}
provided $\psi(\xi, \xi') > 0 \Rightarrow \psi(\xi', \xi) > 0$.  The
Metropolis-Hastings strategy assumes a family of ``elementary moves''
represented by transition probabilities $\{\psi_m(\xi, \xi')\}$.
At each step, say $t$, of the algorithm, a move $m_t$ is
chosen (based on a random or deterministic scheme), and a new
configuration is created with probability $\psi_{m_t}^*(\xi_{t-1},
\cdot)$, where $\xi_{t-1}$ is the current configuration. The set of
elementary moves and the updating scheme must be chosen appropriately
to ensure that the chain is ergodic.

\subsection{Application to the Scene Model}
The feasibility of the method relies on whether the ratio intervening
in \eqref{eq:mh} is tractable. In this equation, all terms can be
relatively easily computed, with the exception of the probabilities
$p(z|s)$ in \eqref{eq:cond.prob} because of the normalizing constant in \eqref{sect:diss1}
which depends on $s$. This constant cancels in the ratio whenever the
values of $s$ in $\xi$ and $\xi'$ coincide, i.e., the elementary move
does not change the scene geometry. Among moves that satisfy this property, moves involving the camera
properties $w$ are generally computationally demanding, because they
modify all the annobits, while elementary changes in $z$ only have a
local impact.

\subsubsection{Changing the Scene Geometry}
To process moves that modify $s$, the normalizing constant in \eqref{sect:diss1}, namely
\begin{equation}
\kappa(\boldsymbol \lambda) = \sum_{z\in \Omega} \exp(\boldsymbol \lambda \cdot \boldsymbol\varphi(z))
\end{equation}
must be computed (where $\boldsymbol\lambda, \Omega$ and
$\boldsymbol\varphi$ all depend on $s$). Whereas an exact computation is
intractable, approximations can be obtained, using, for example, the
formula
\begin{multline}
\log \kappa(\boldsymbol \lambda) = \log \kappa(\boldsymbol \lambda_0) \\
+ \int_0^1 \!\!\!  E\Big((\boldsymbol \lambda - \boldsymbol \lambda_0)\cdot \boldsymbol\varphi\mid \boldsymbol\lambda_0 + t(\boldsymbol \lambda - \boldsymbol \lambda_0)\Big) dt
\end{multline}
in which $\boldsymbol\lambda_0$ is a parameter at which $\kappa$ is
computable (typically making all variables independent) and each
expectation in a numerical approximation of the integral is computed
using Monte-Carlo sampling. This is a costly but can be computed offline for each value of $s$ (which can
be discretized over a finite set).

In our application, however, we have used a simpler approach, relying
on a good estimator of $S$ that is fixed in the rest of the
computation. Letting $\hat S$ be this estimator, we sampled $S$ over a
small neighborhood of $\hat S$, making the additional approximation
that $\kappa$ in constant (as a function of $s$) in this
neighborhood.

\subsubsection{Changing the Camera Properties}
For the camera properties, we use a proposal distribution taking the form
$\psi_W(\xi, \xi') = p(w'|I)$, where the $z$ and $s$ coordinates in $\xi$ and $\xi'$ coincide, and $I$ is the observed image. The dependency on $I$ is implemented through an estimator limiting the camera parameters, which will be described in the next section. The proposal distribution of $S$ can be assumed to be uniform over the finite set of scene geometries which is considered.

\subsubsection{Changing object indicators}
In our implementation, in which $z = \{z_j\}$ is a collection of
binary variables, elementary moves correspond to Gibbs sampling,
taking,
\[
\psi_j(\xi, \xi') = p(z'_j\mid \mathbf{E}_{k-1}, \{z_l \hspace{0.5mm}, l \neq j\})
\]
if $\xi = (z, w, s)$ and $\xi' = (z', w', s')$ are such that $w=w'$,
$s=s'$ and $z_l = z'_l$ for $l\neq j$; and taking $\psi_j(\xi, \xi') = 0$ in
all other cases.

The  overall updating scheme is based on nested loops, where the inner loop updates $z$, the middle one updates $w$ and the outer one $s$. Each loop is run several times before an update is made at a higher level.

\section{Classifiers and Data Model}
\label{sec:CNNs}

We trained three deep CNNs.  The first one, ``CatNet,'' is for
object category classification; the second one,
``ScaleNet,'' is to estimate the size of detected object instances, and
the third, ``SceneNet,'' is to estimate the scene
geometry in a given image. All of these CNNs borrow their network
architecture, up to the last weight layer, \ie layer 15, from the
VGG-16 network \citep{Simonyan14c}. The last fully-connected layer (16-th weight layer)
and the following softmax layer of these three CNNs were modified to
accommodate our design needs. All CNNs rely on ``transfer learning''
by initializing the first 15 weight-layers to the corresponding
weights from the VGG-16 network\footnote{Available at:
\url{http://www.robots.ox.ac.uk/~vgg/research/very_deep/}} trained on
1.2 million images from the ImageNet dataset
(see~\cite{Deng13}). However, since the last layer's architecture for
all three CNNs is different from VGG-16, the corresponding
weights were randomly initialized during training. All CNNs were trained and tested using the Caffe Deep Learning
framework~\citep{jia2014caffe} using an Nvidia Tesla K40 GPU on a
desktop computer with Intel i7-4790K Quad-Core processor (8M Cache and
up to 4.40 GHz clock rate) and 32-GB RAM running Ubuntu 15.04
operating system. The processing time for each patch is about 12
seconds on our end-of-the-line Intel i7-4790K CPU and 0.2 seconds on
the Tesla K40 GPU. Since the input patches are of the same
size, namely $224 \times 224$, and pass through the same network, the
classifiers all have the same computational cost during test time. We
describe the design, training, and performance of these
CNNs in the following subsections.

\subsection{CatNet}

For each object category $c\in \mathcal C$, we want to detect
if there is at least one instance in a given patch
$A$.  This will be done simultaneously for all categories, including
``background.''  Moreover, all patches are resized to $224 \times 224$ and only one
CNN is trained independently of the original size of $A$ in the image.  This suffices
in our framework since patches are restricted to the 4-level annocell hierarchy
and the smallest annocells remain at the scale of objects except in
extreme cases.
CatNet is then a CNN with a softmax output layer, which returns a
vector of scores $X^\c = (X^\c_c, c\in \mathcal C\cup \{0\})$, where
each $X^\c_c$, for $c\in \mathcal C$, reflects a proportional confidence level
about the presence of at least one object from category $c$ in the
patch, while $X^\c_0$ corresponds to an empty patch (or the "No Object"
category). The scores are non-negative and sum to 1, but they should
not be interpreted as probability of existence, since the events
they represent are not incompatible \ie they can co-occur.

The corresponding annobit $Y^\c_A$ is a binary vector $Y^\c_A =
(Y^\c_{c,A}, c\in \mathcal C)$ where $Y^\c_{c,A}=1$ if and only if an
object with category $c$ exists in $A$. The conditional distribution
$P(x^\c|y^\c)$ is taken to be independent of $A$, and modeled as a
Dirichlet distribution separately for each of the $2^{|\mathcal C|}$
possible configurations of $Y^\c$.  We used a fixed-point (without
projection) iterative schemes to perform MLE parameter estimation
(see~\cite{Minka12}).

\begin{figure}[h]
\centering
\setlength{\fboxrule}{2pt}%
\fbox{\includegraphics[width=0.45\linewidth]{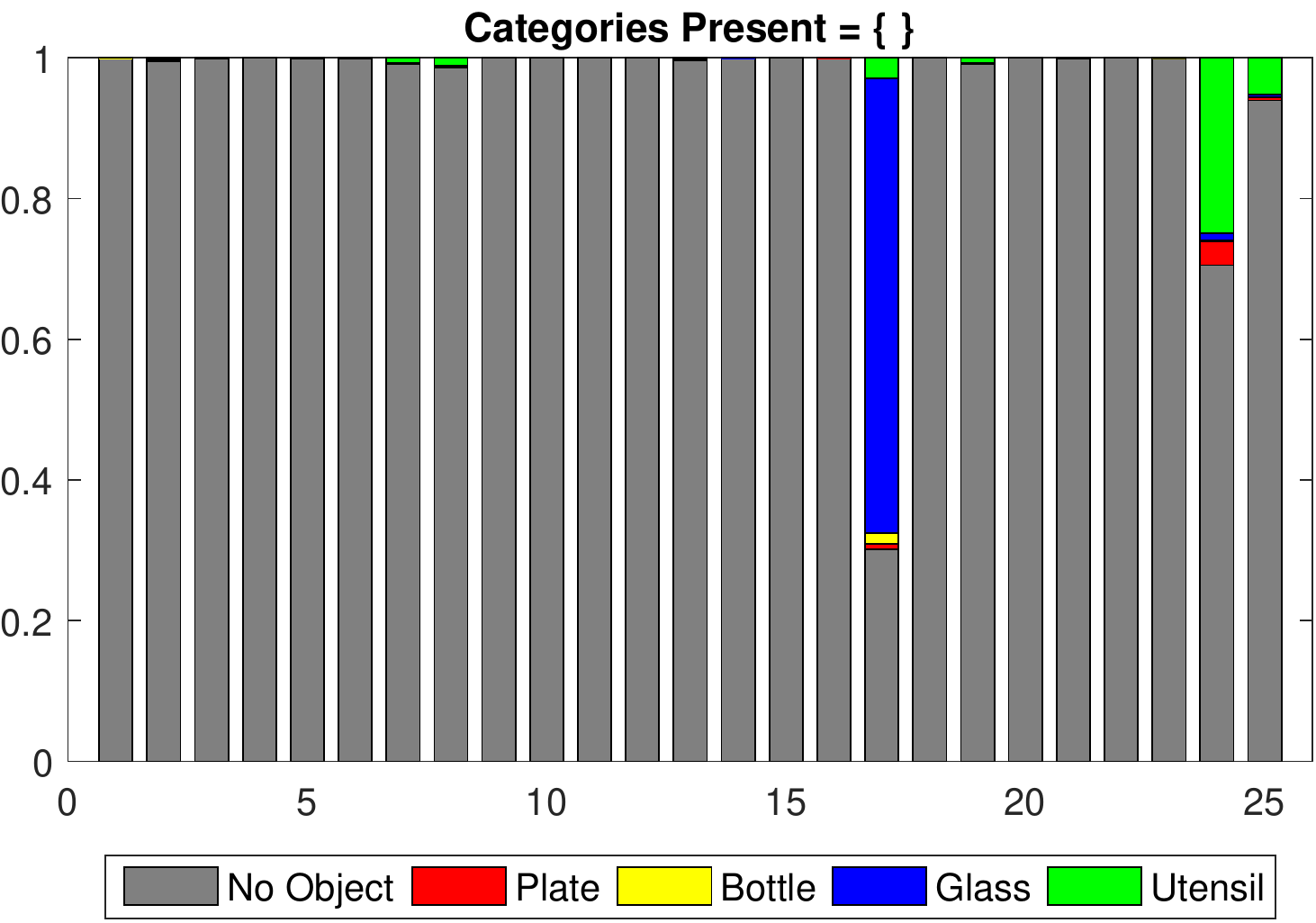} \hspace{2mm}
\includegraphics[width=0.45\linewidth]{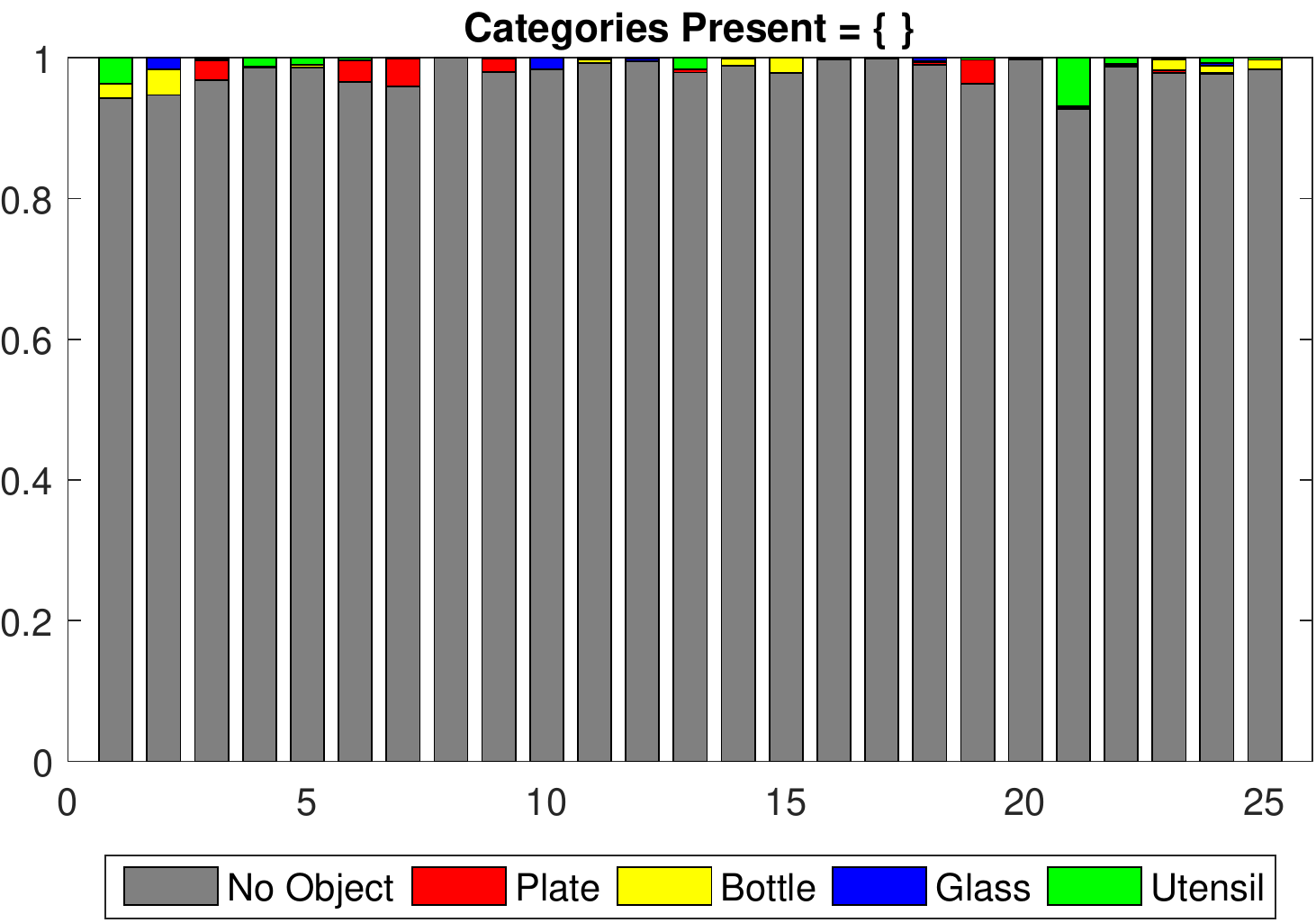}} \\
\vspace{0.5mm}
\fbox{\includegraphics[width=0.45\linewidth]{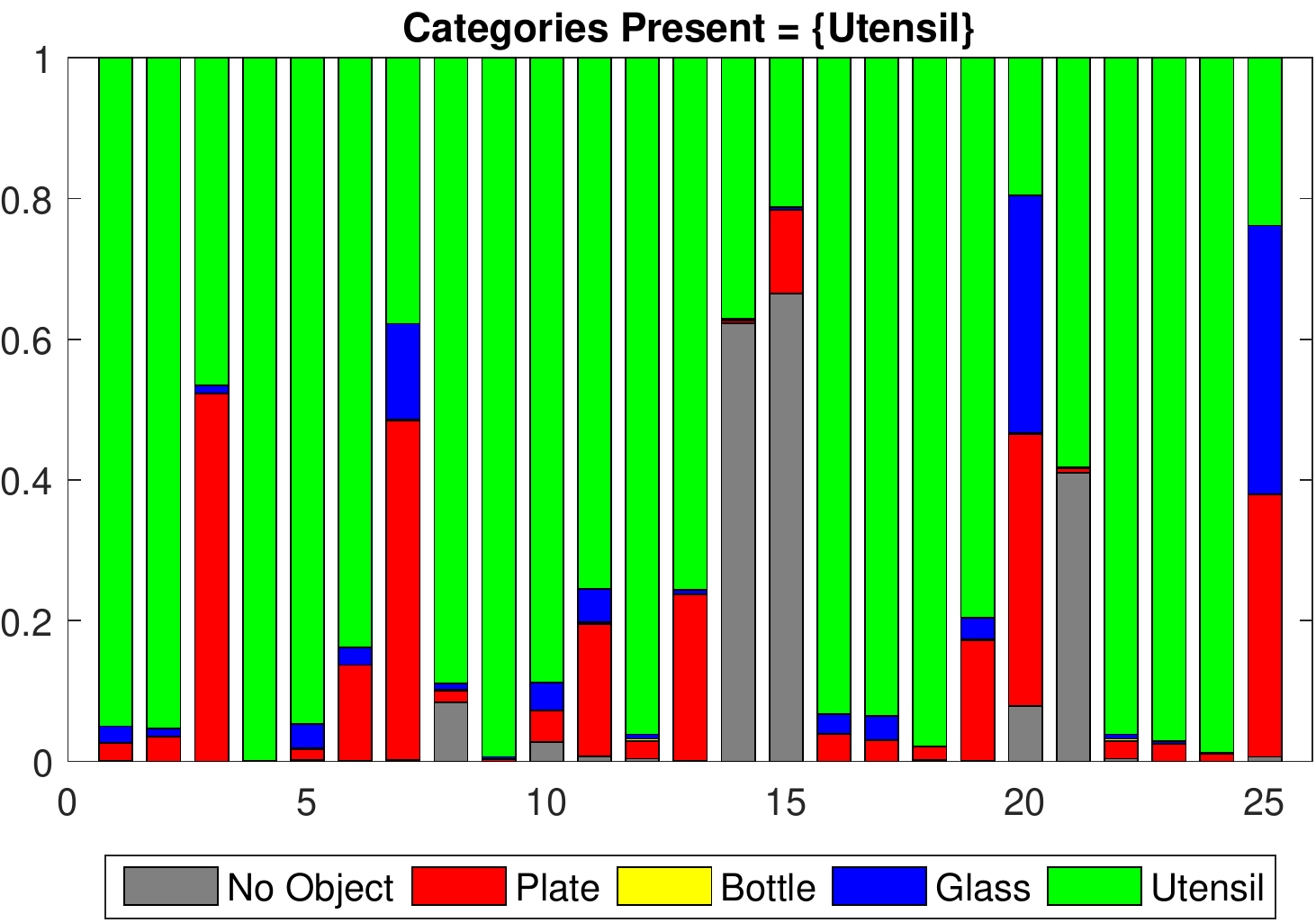} \hspace{2mm}
\includegraphics[width=0.45\linewidth]{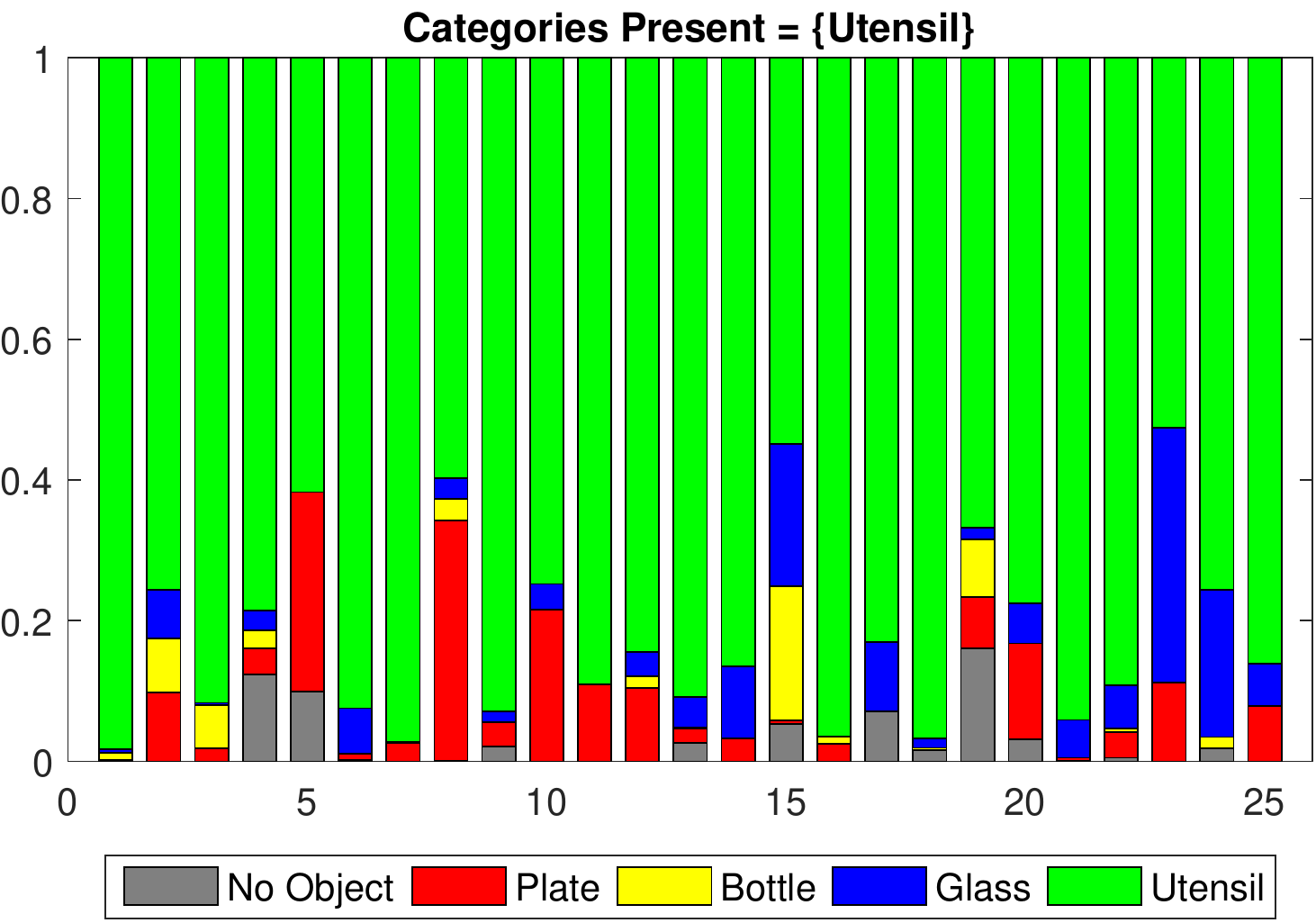}} \\
\vspace{0.5mm}
\fbox{\includegraphics[width=0.45\linewidth]{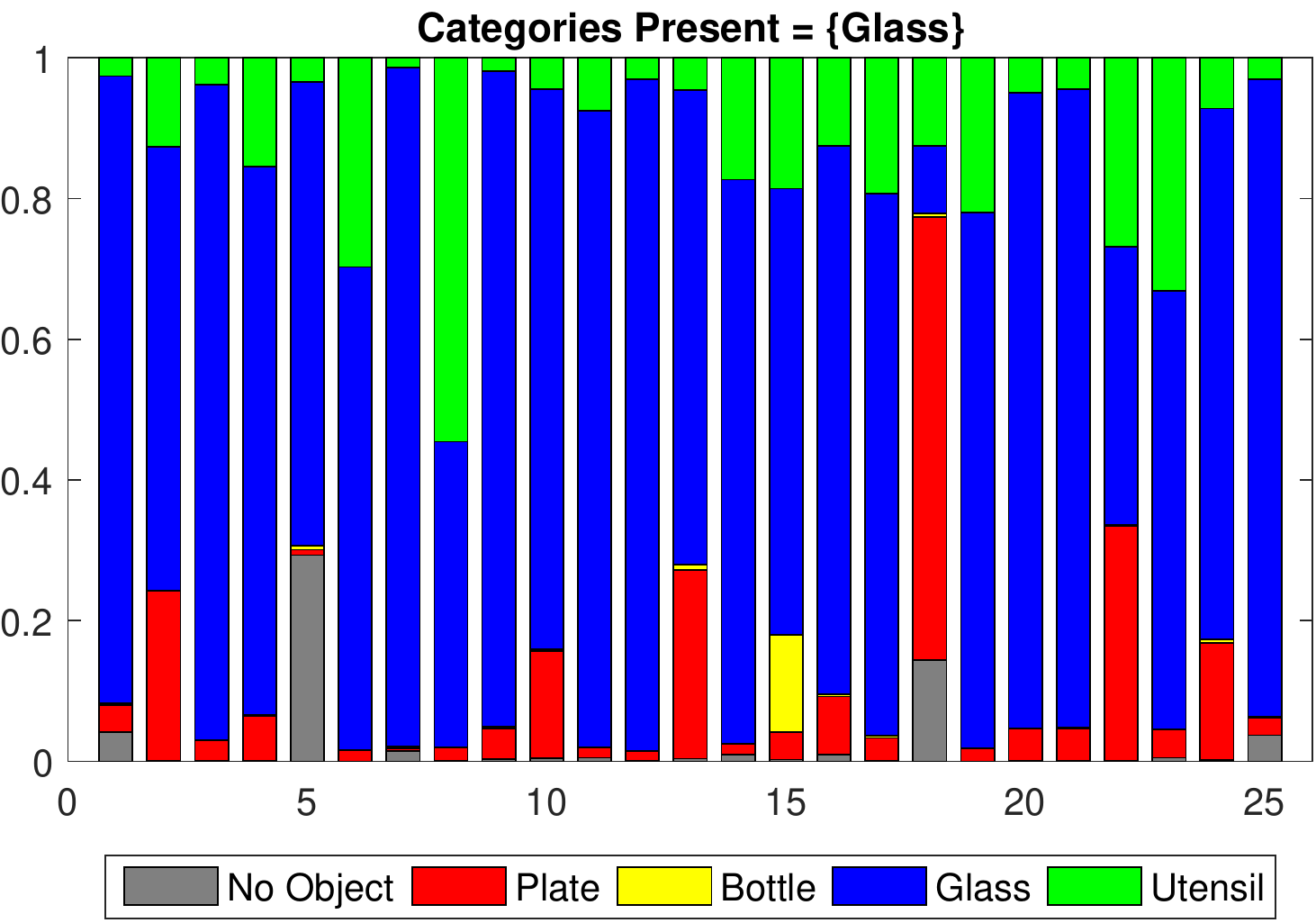} \hspace{2mm}
\includegraphics[width=0.45\linewidth]{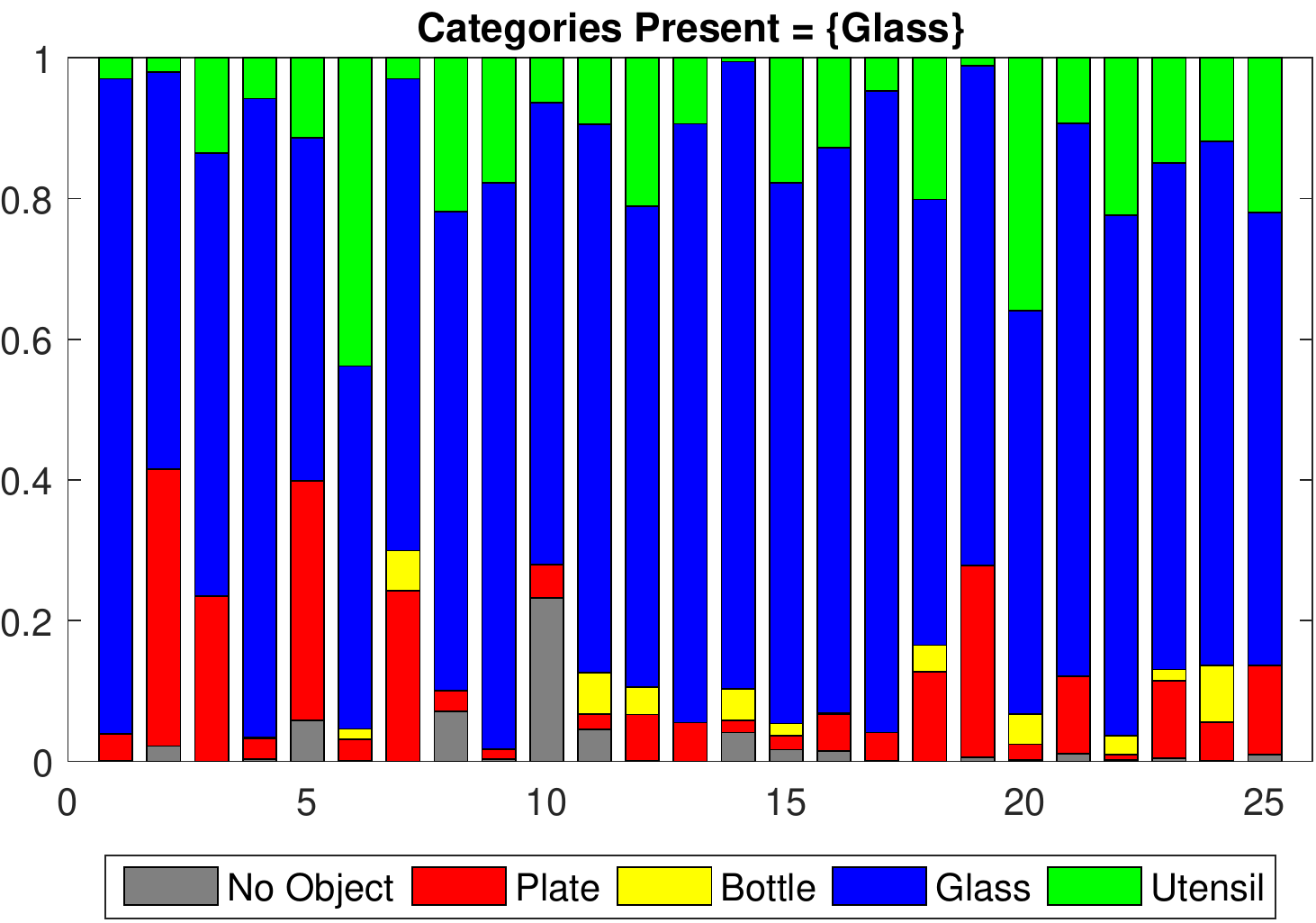}} \\
\vspace{0.5mm}
\fbox{\includegraphics[width=0.45\linewidth]{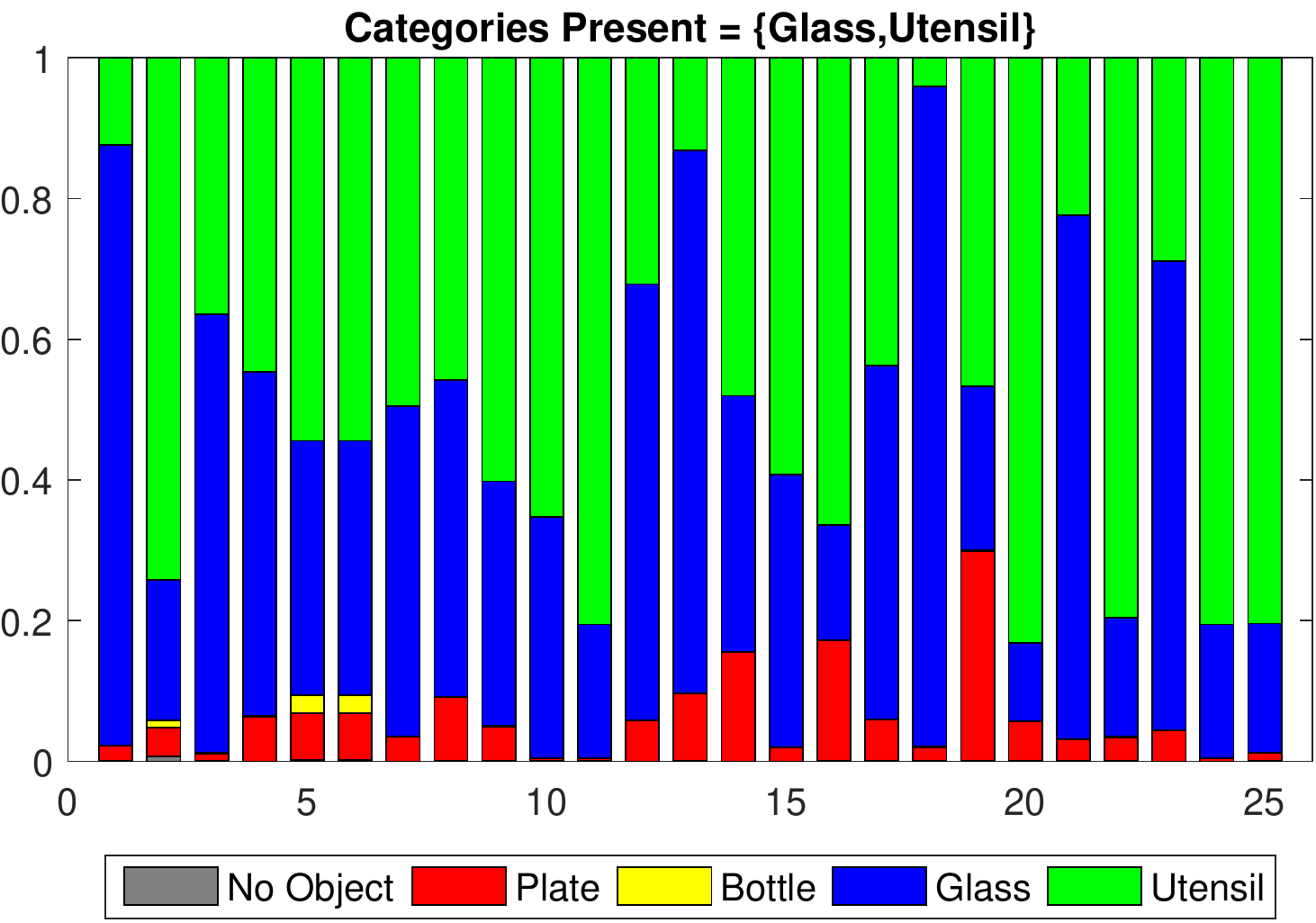} \hspace{2mm}
\includegraphics[width=0.45\linewidth]{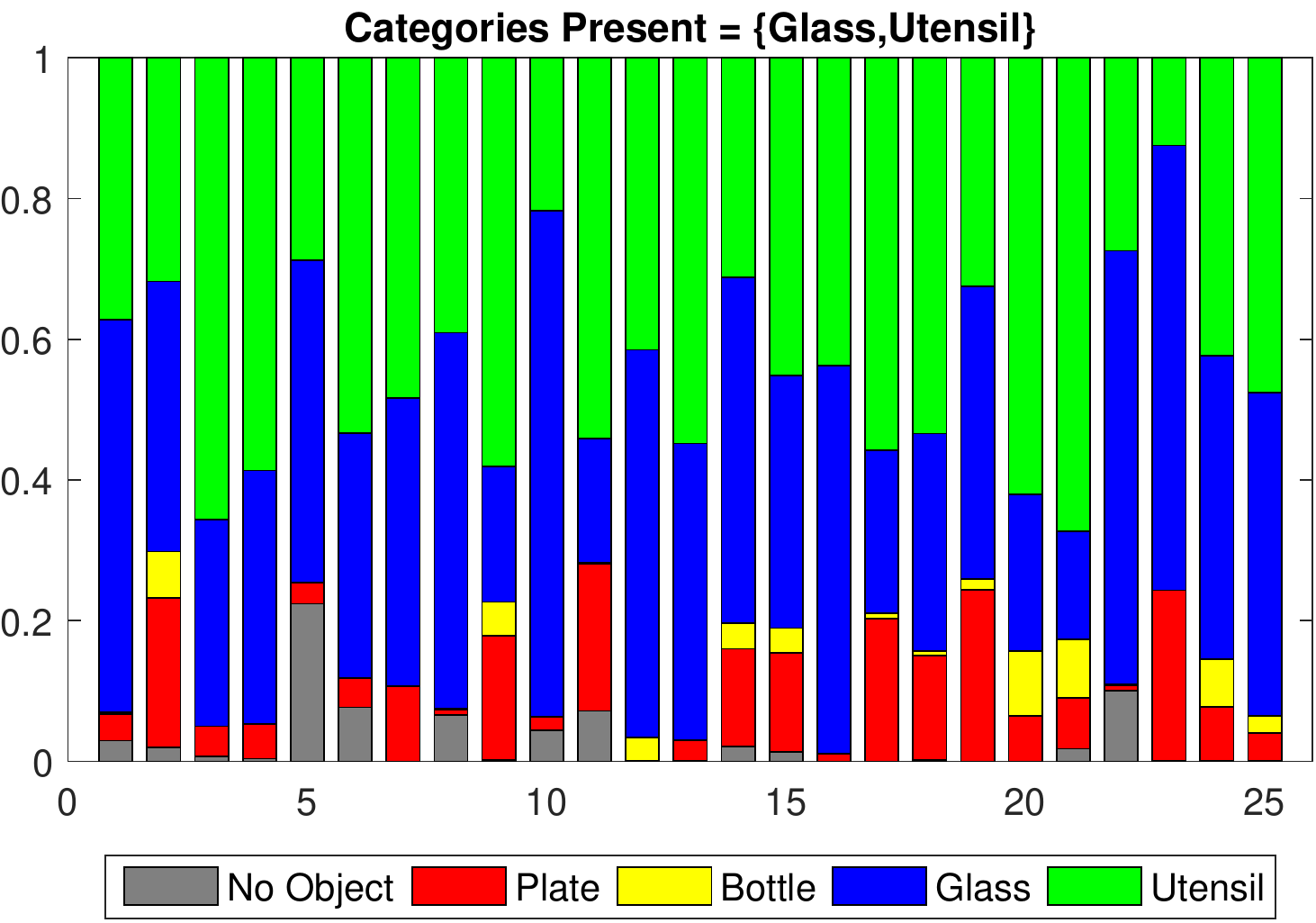}} \\
\caption[Stacked bar CatNet and Dirichlet samples.]{Stacked bar visualization of samples from CatNet output (left) and Dirichlet model (right).}
\label{fig:CatNet_TrainDirchSamples1}
\end{figure}

Figure~\ref{fig:CatNet_TrainDirchSamples1} illustrates some samples from the learned Dirichlet distribution versus some sample CNN outputs for the corresponding annobit $Y^{cat}$ for a few configurations. We have $|\mathcal C|=4$ and therefore
estimated 16 conditional distributions. The figure shows stacked bar
visualization of 25 samples (per configuration) drawn randomly from
data collected by running CatNet on patches (left column) and samples
taken from the Dirichlet model learned from CatNet output data (right
column) where each row corresponds to one of the 16 annobit
configurations. We have shown stacked bars for only four configurations
as example. The length of each colored bar represent the proportion of
each category; therefore, the total length of each stacked bar is
equal to 1. Two interesting observations are: (1) the length of bars
corresponding to the present categories are comparable and usually
considerably larger than the length of absent categories; (2) the
color distribution of CatNet outputs and Dirichlet model samples are
very similar for the same configuration. This supports the argument
for using a Dirichlet distribution in modeling the data
distribution $p(x^\c | y^\c)$. Stacked bars are good means to visually
inspect and compare the true empirical distribution versus the
Dirichlet model.

\subsection{ScaleNet}
\label{ScaleNet}
\begin{figure}[t]
\centering
\includegraphics[width=0.96\linewidth]{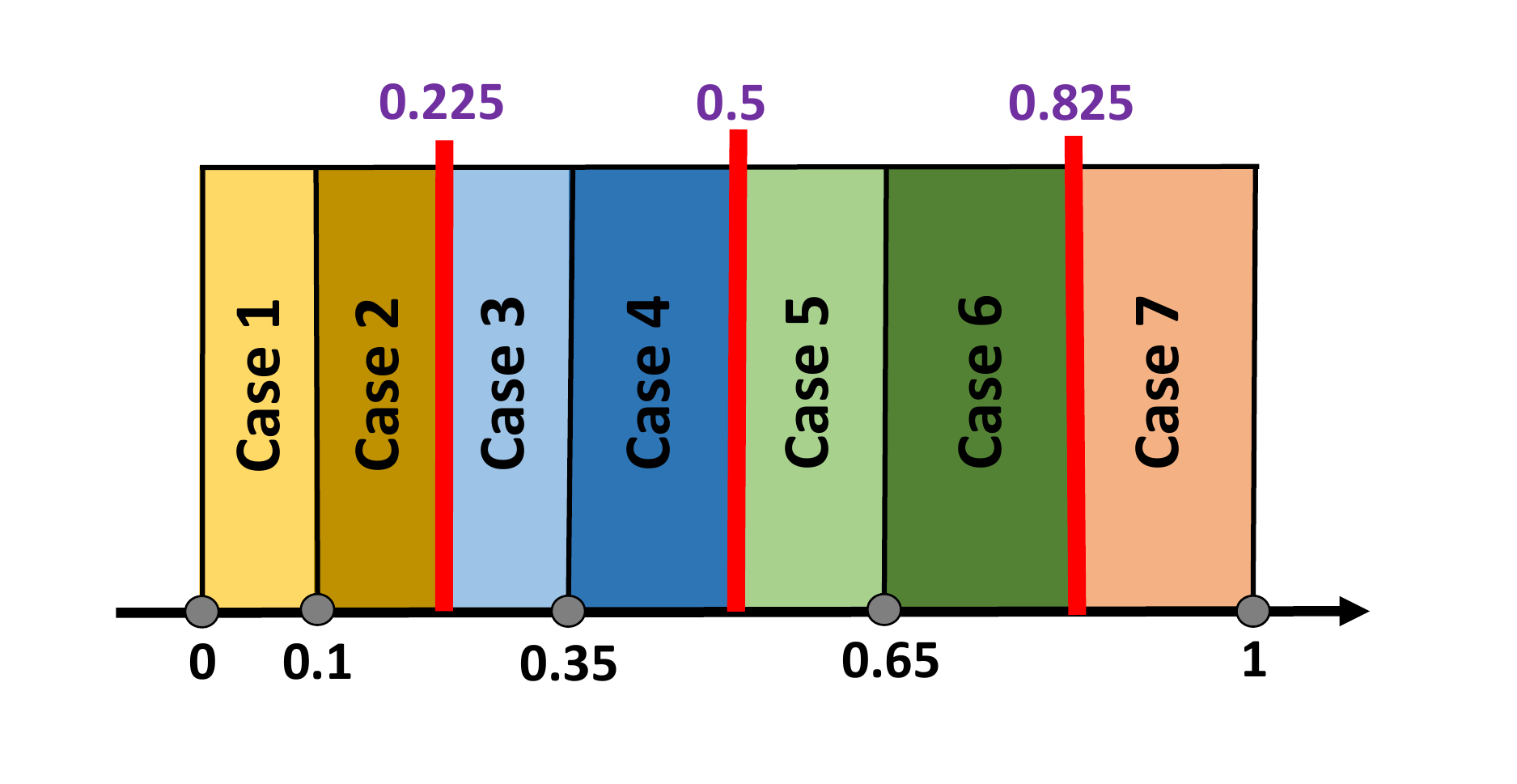}
\vspace{-4mm}
\caption[Scale ratio intervals.]{\small{Scale ratio intervals.}}
\label{fig:ScaleIntervals}
\end{figure}

Define the scale of an object in an image patch as the ratio of its
longest side to the patch size (therefore belonging to $(0,1]$
when completely visible.  The ScaleNet predictor is designed to estimate the average
scale of object instances in a given patch, independent of
their category.

Assume a quantization $(\tau_0 = 0, \tau_1, \ldots, \tau_{d-1}, \tau_d=1)$ of the
unit interval (in our experiments, we used $d=4$ and quantization levels of
$0.1$, $0.35$, $0.65$, and $1$). We modified the VGG-16 network by
assigning $d$ output values to the softmax layer and trained by
assigning to each patch in the training data the index $j\geq 1$ such
that $\tau_j$ is closest to the average scale of the objects it
contains, using only non-empty patches. The output of the CNN is a
vector $X^\s = (X^\s_1, \ldots, X^\s_d)$ of non-negative weights
summing to one. Again, there is only one CNN and patches of different
sizes are aggregated for training.  The associated annobit $Y^\s\in \{1, 2d-1\}$ is the
index of the Vorono\"\i\ cell that contains the average scale,
obtained by adding midpoints $\tau_{j+1/2} = (\tau_j + \tau_{j+1})/2$
to the initial sequence (which separates the unit interval into $2d-1$
regions; see Figure \ref{fig:ScaleIntervals}). The conditional
distribution $P(x^\s|y^\s)$ is then modeled and trained as a Dirichlet
distribution for each value $y^\s = 1, \ldots, 2d-1$.

\begin{figure}[h]
\centering
\setlength{\fboxrule}{2pt}%
\fbox{\includegraphics[width=0.45\linewidth]{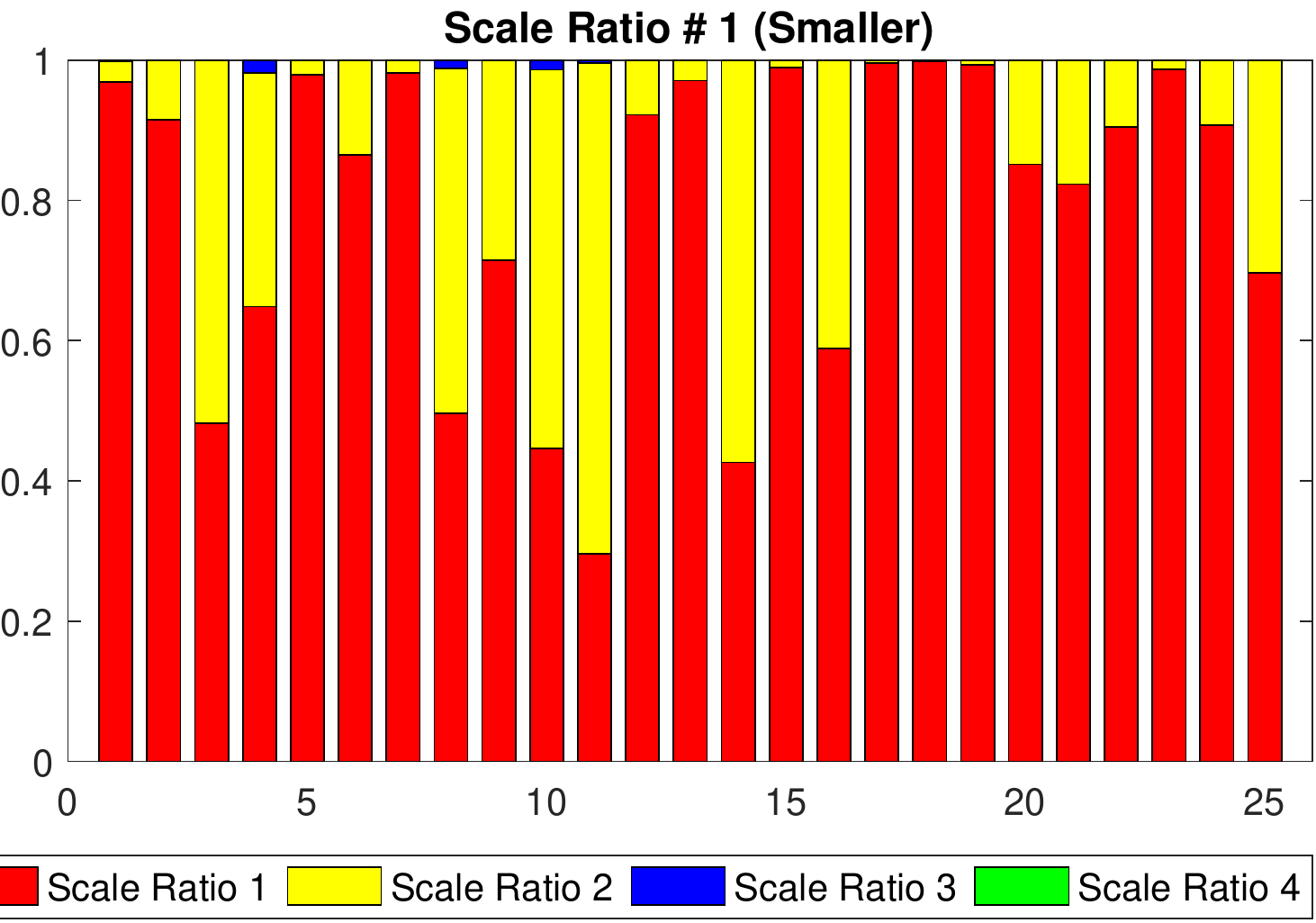} \hspace{2mm}
\includegraphics[width=0.45\linewidth]{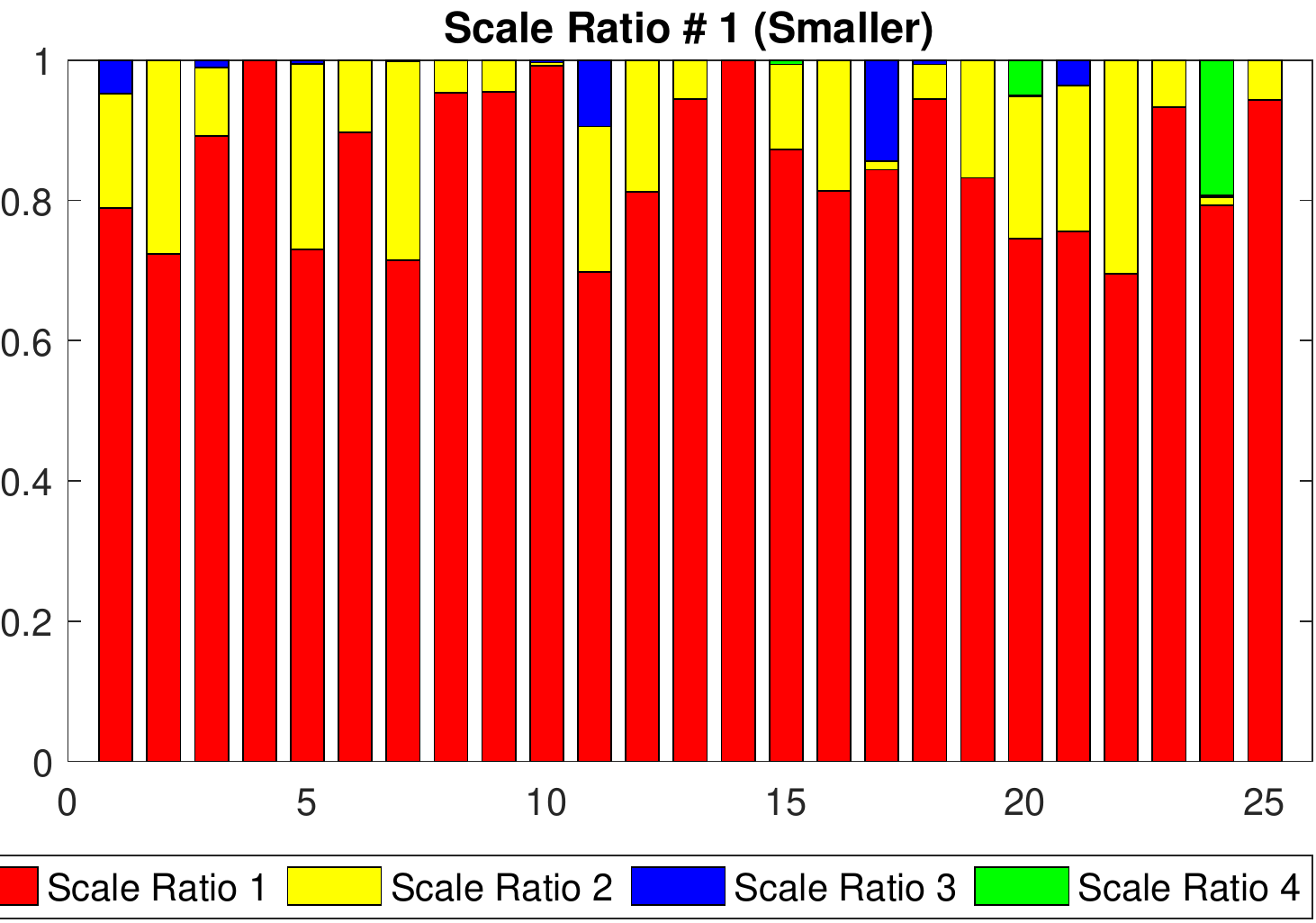}} \\
\vspace{0.5mm}
\fbox{\includegraphics[width=0.45\linewidth]{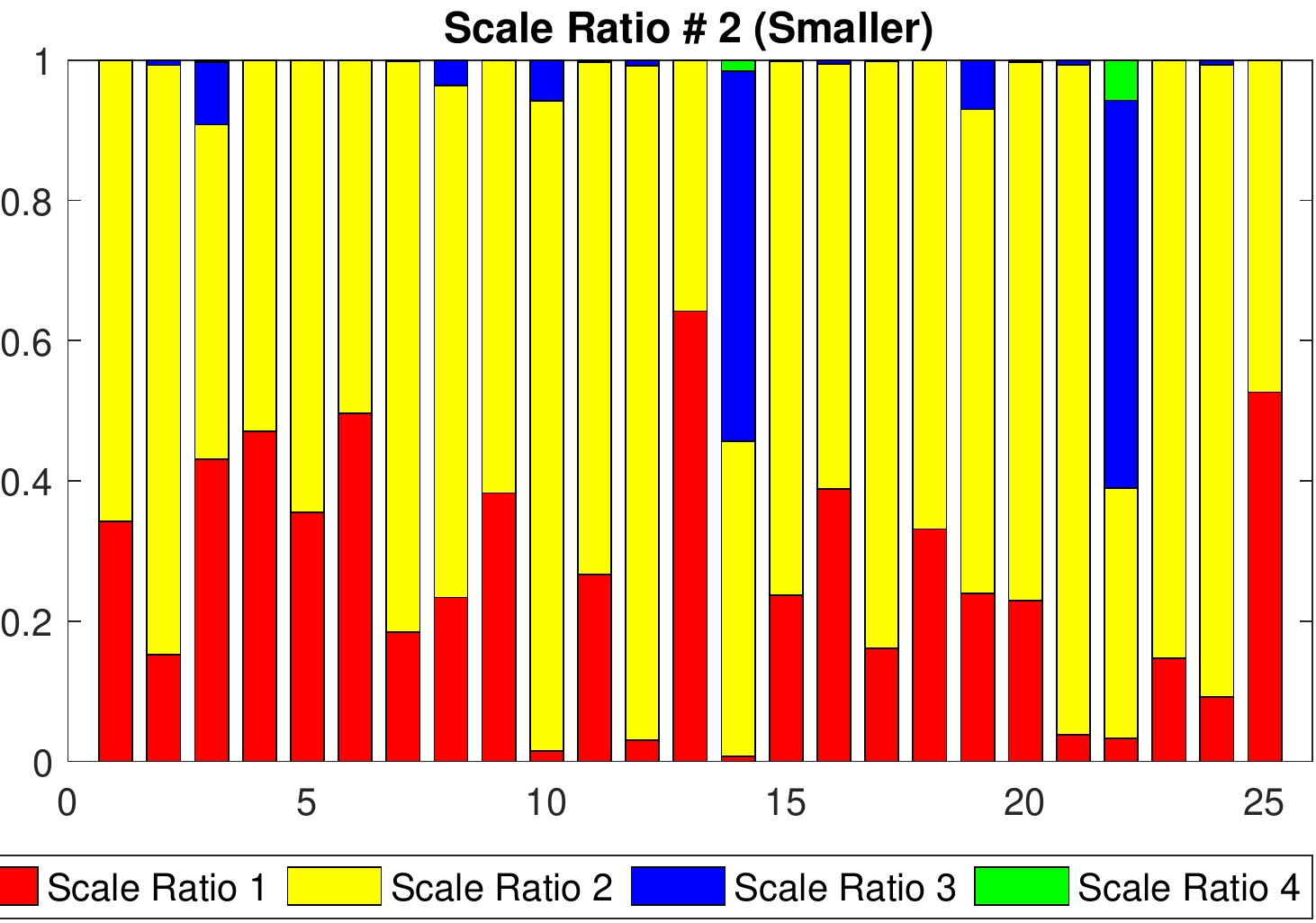} \hspace{2mm}
\includegraphics[width=0.45\linewidth]{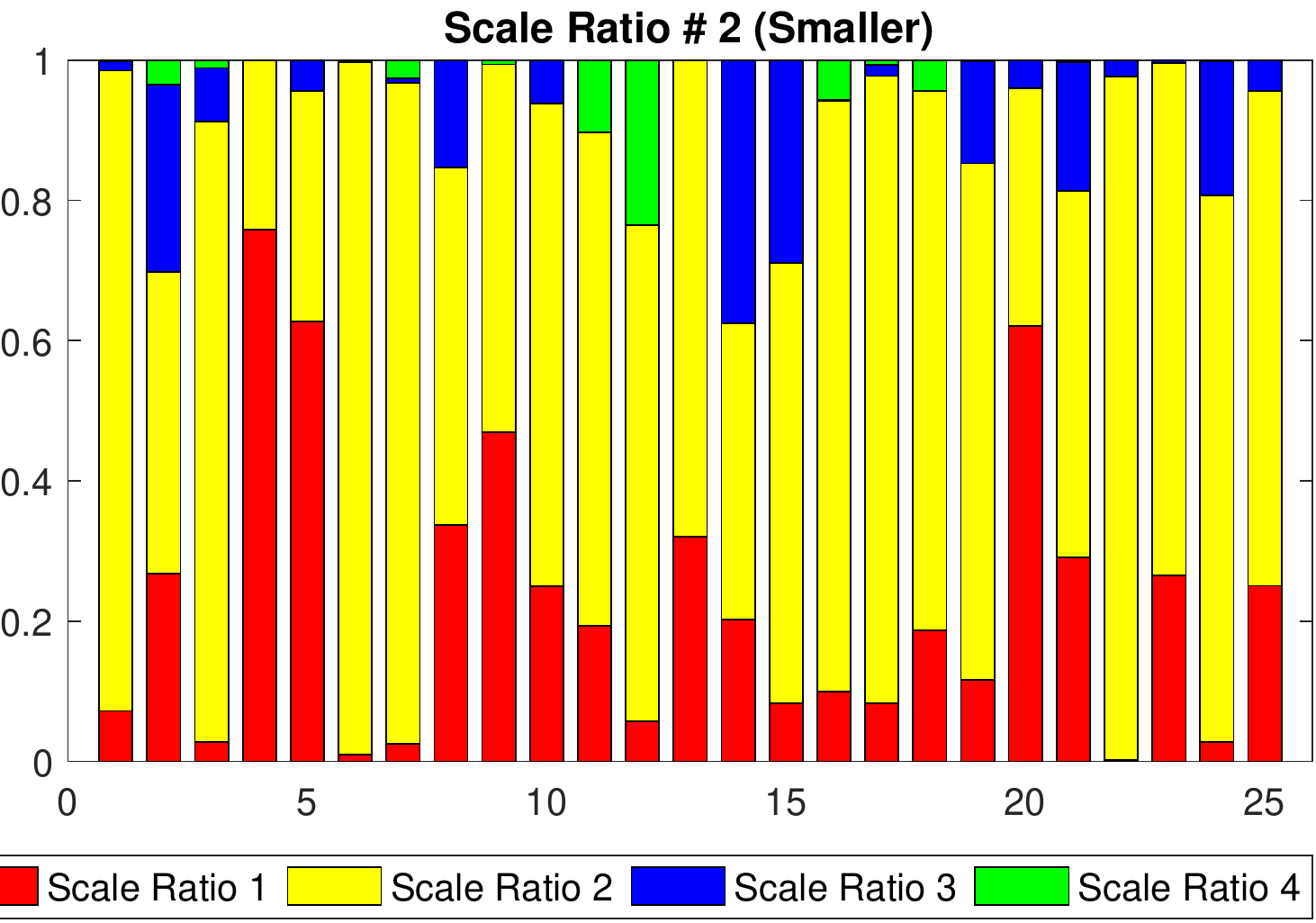}} \\
\vspace{0.5mm}
\fbox{\includegraphics[width=0.45\linewidth]{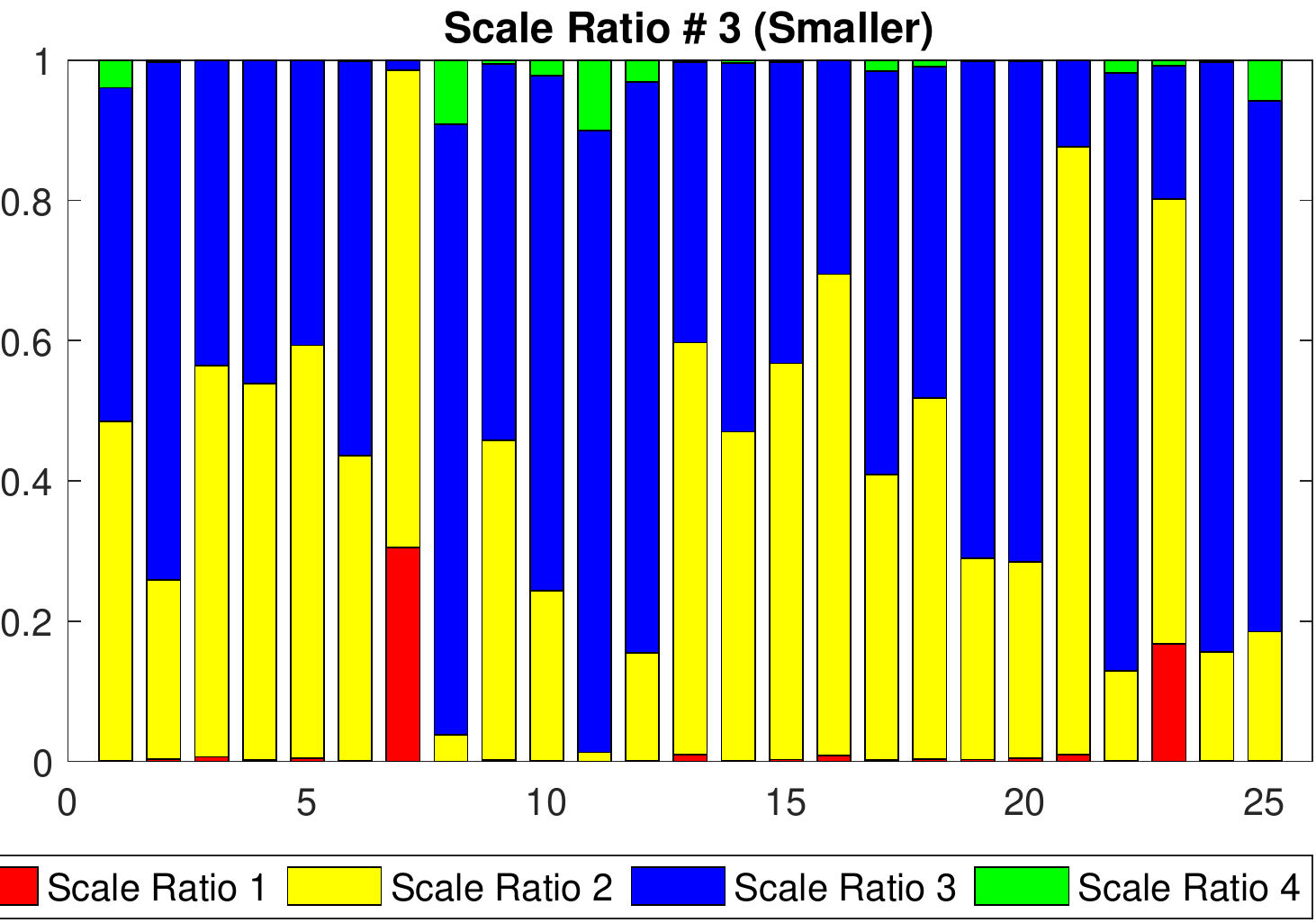} \hspace{2mm}
\includegraphics[width=0.45\linewidth]{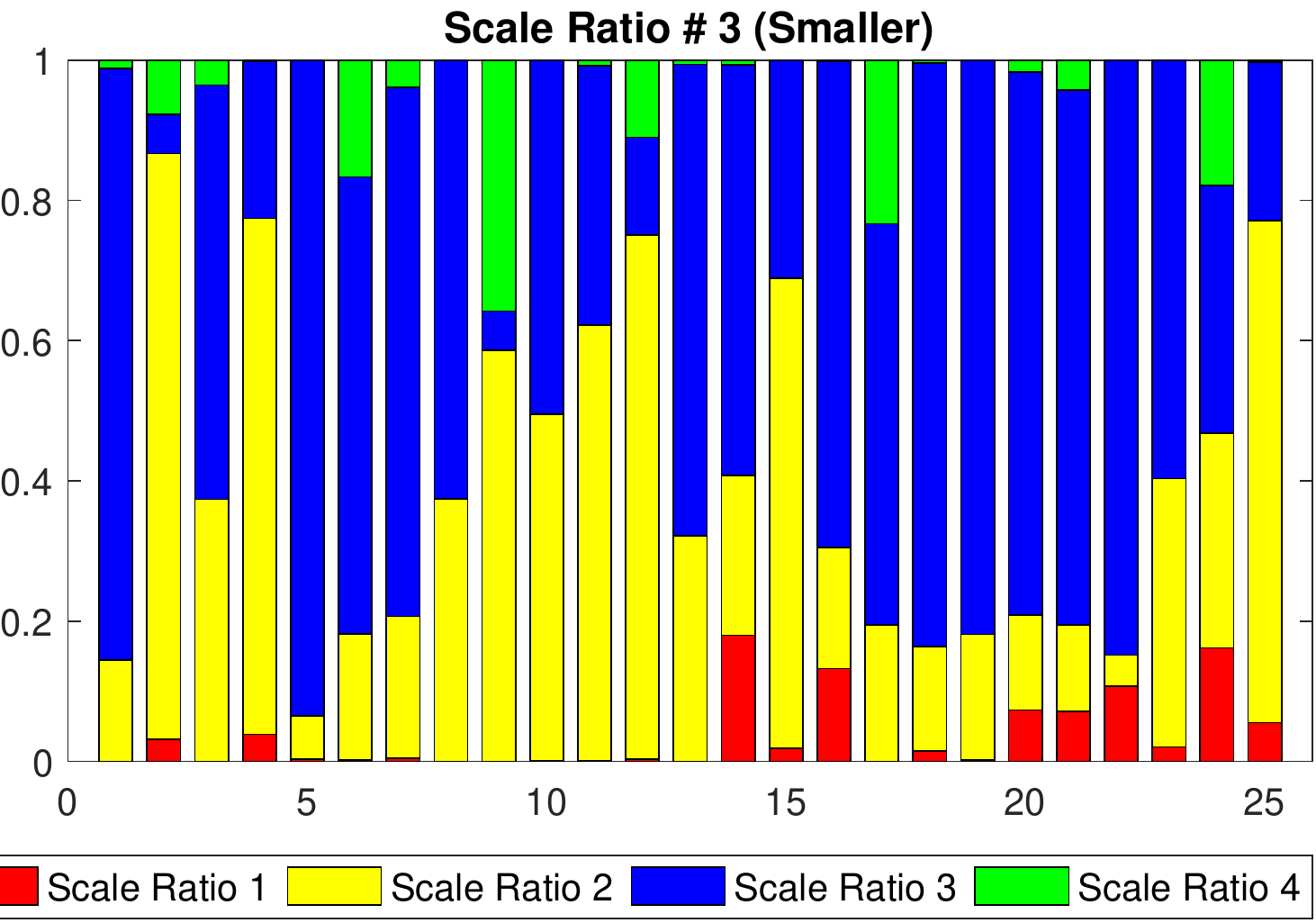}} \\
\vspace{0.5mm}
\fbox{\includegraphics[width=0.45\linewidth]{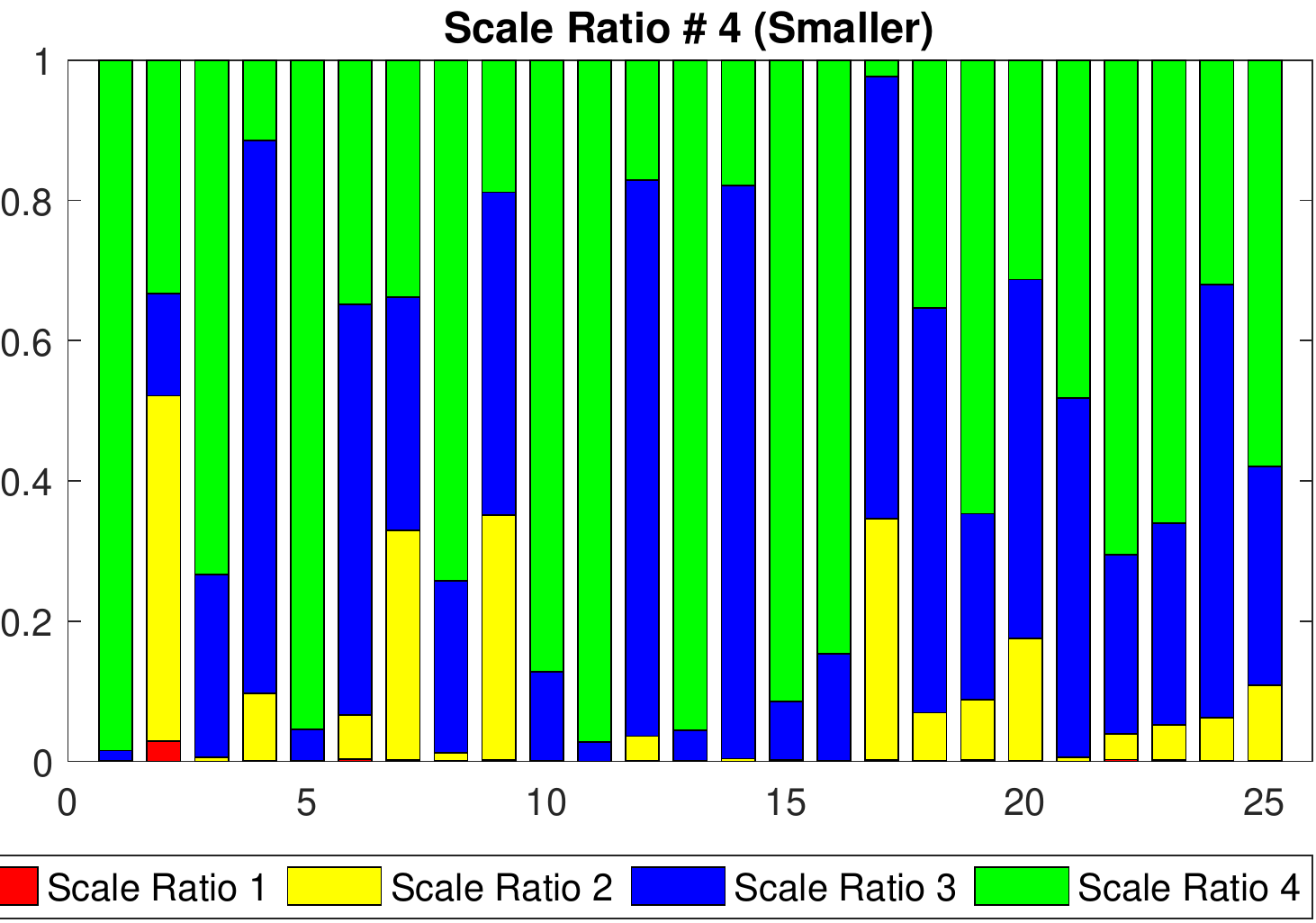} \hspace{2mm}
\includegraphics[width=0.45\linewidth]{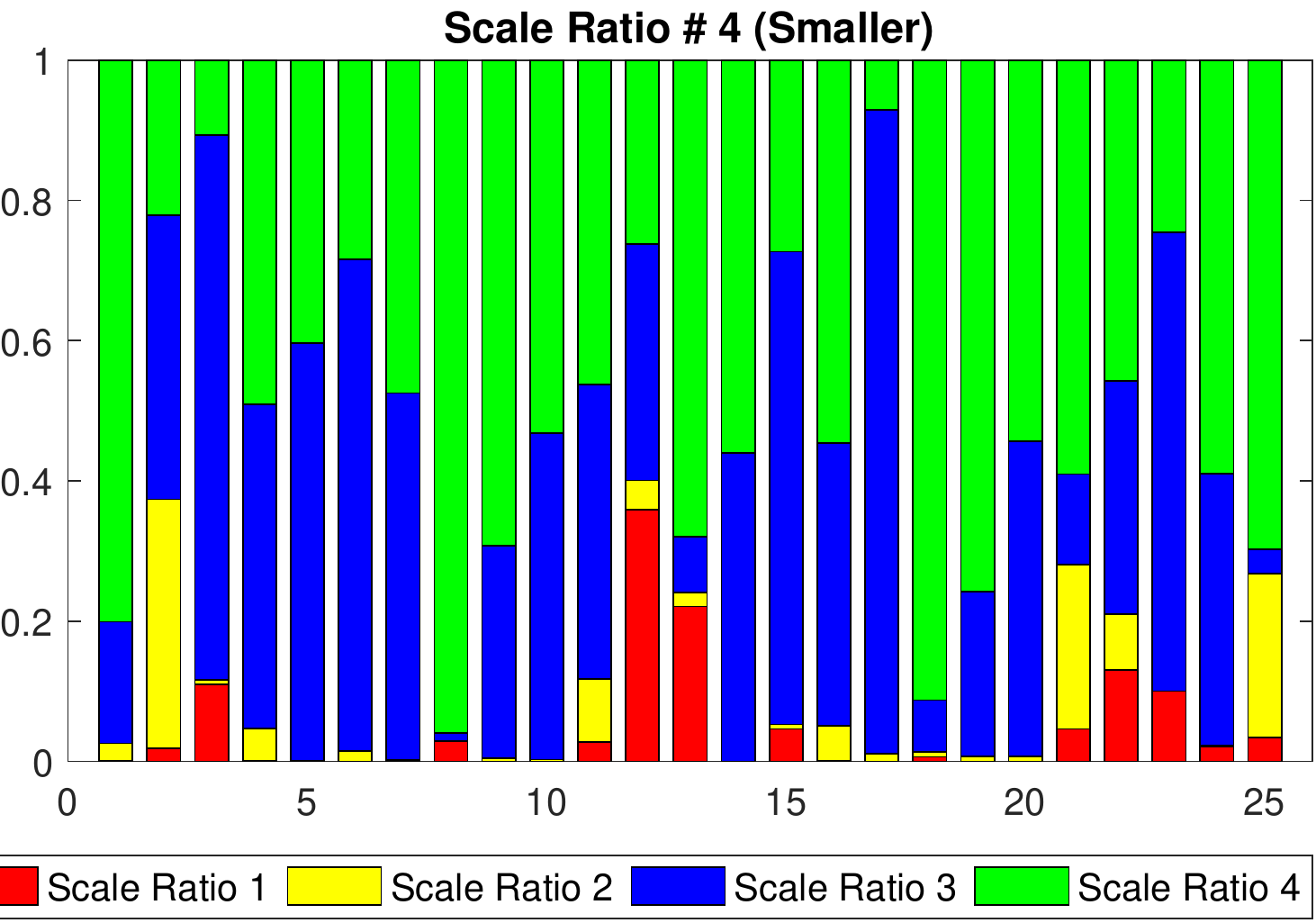}} \\
\caption[Stacked bar ScaleNet and Dirichlet samples.]{\small{Stacked bar visualization of samples from ScaleNet output (left) and Dirichlet model (right).}}
\label{fig:ScaleNet_TrainDirchSamples}
\end{figure}

Figure~\ref{fig:ScaleNet_TrainDirchSamples} (similar to Figure~\ref{fig:CatNet_TrainDirchSamples1}) provides some ScaleNet stacked bars visualizations for 4 (out of $2d-1 = 7$) scale configurations.

\subsection{SceneNet}
\label{TableNet}

SceneNet combines binary classifiers predicting whether or not an input patch belongs to the dominant plane. The basic architecture is the same as that of the CatNet and ScaleNet.  It returns a region $\hat M$ in the image plane. For a given scene geometry $s$ and camera properties $w$, let $M(s, w)$ be the representation of $s$ in the image plane. We discretize the image plane into non-overlapping patches, and let
$\mathbf X^g = (X^g_j, j=1, \ldots, m_t)$ be the corresponding SceneNet
outputs. Let $Y^g_j=1$ if the corresponding patch belongs to $M(s, w)$
and zero otherwise.

\section{JHU Table-Setting Dataset}
\label{sec:dataset}

\begin{figure}
\centering	
\includegraphics[width=\linewidth]{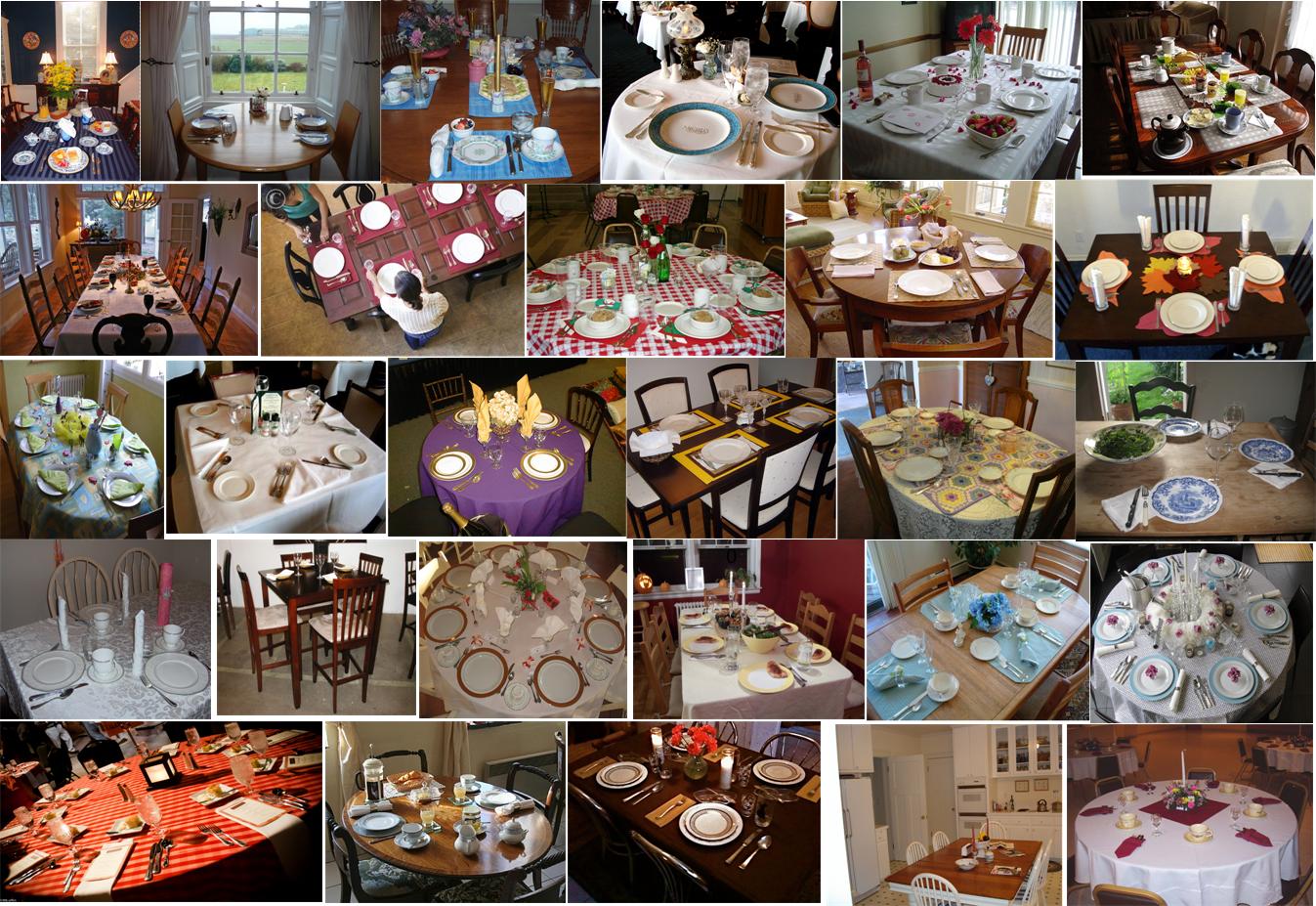}
\caption[Snapshot of the JHU table-setting dataset]{A snapshot of the JHU Table-Setting Dataset.}
\label{fig::JHUDatasetSnapshot}	
\smallskip
\centering
\includegraphics[width=\linewidth]{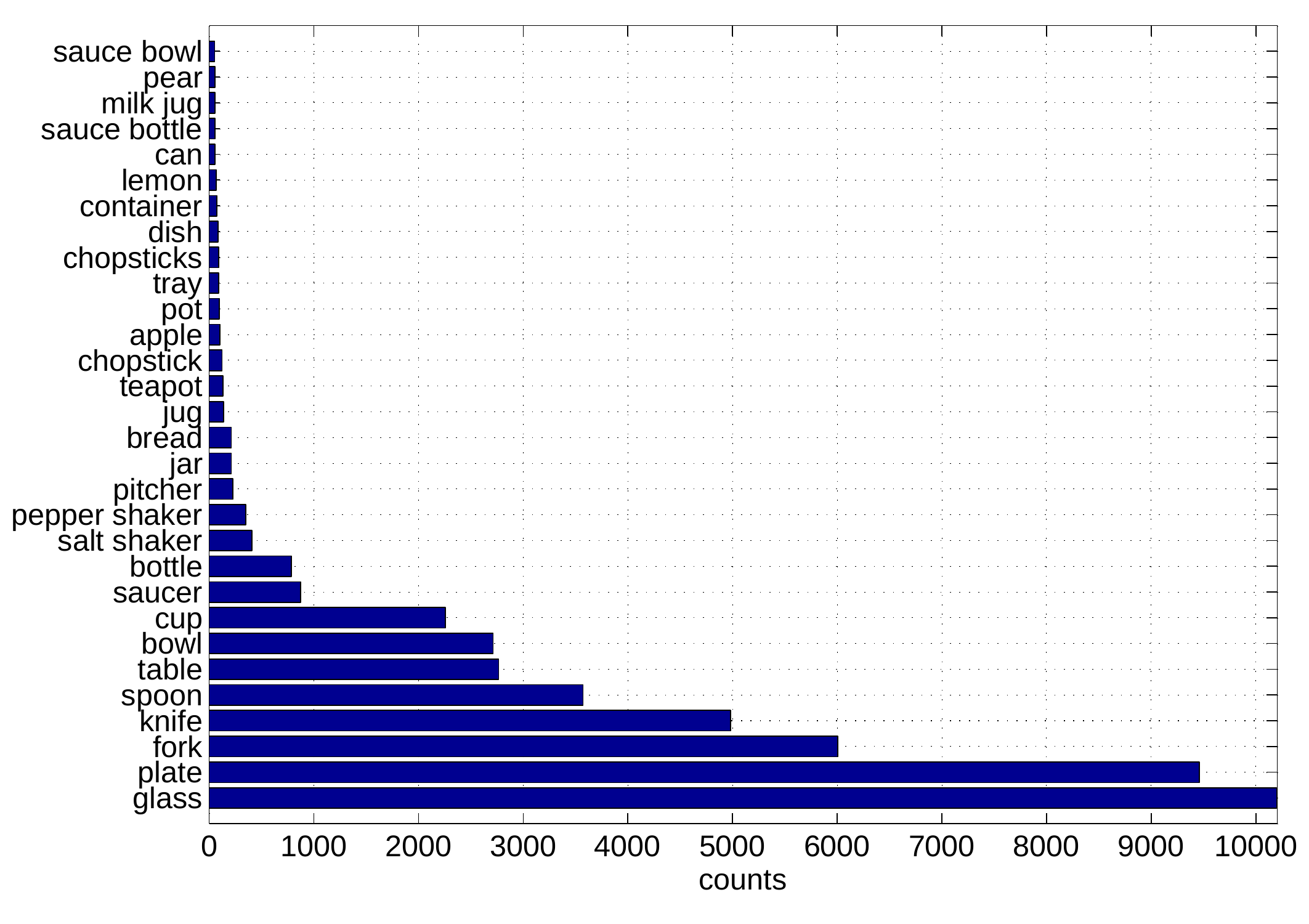} \\
\caption{The number of annotated instances of each object category in the whole dataset for the 30 top most annotated object categories.}
\label{fig:DataSet_stat}
\smallskip
\centering
\includegraphics[width=\linewidth]{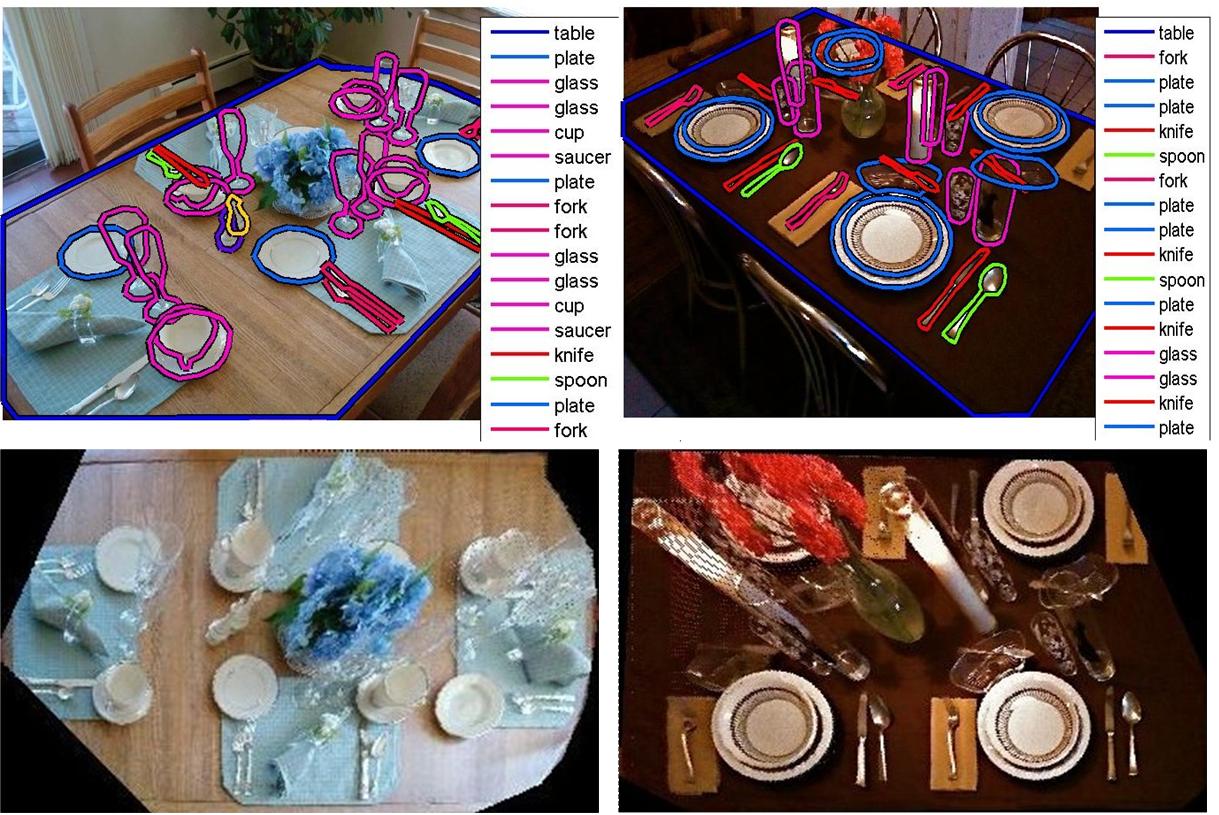}
\caption[Table surfaces rectification]{Rectification of table surfaces after back-projection.}
\label{fig::rectifiedImages}
\smallskip
\centering
\includegraphics[width=\linewidth]{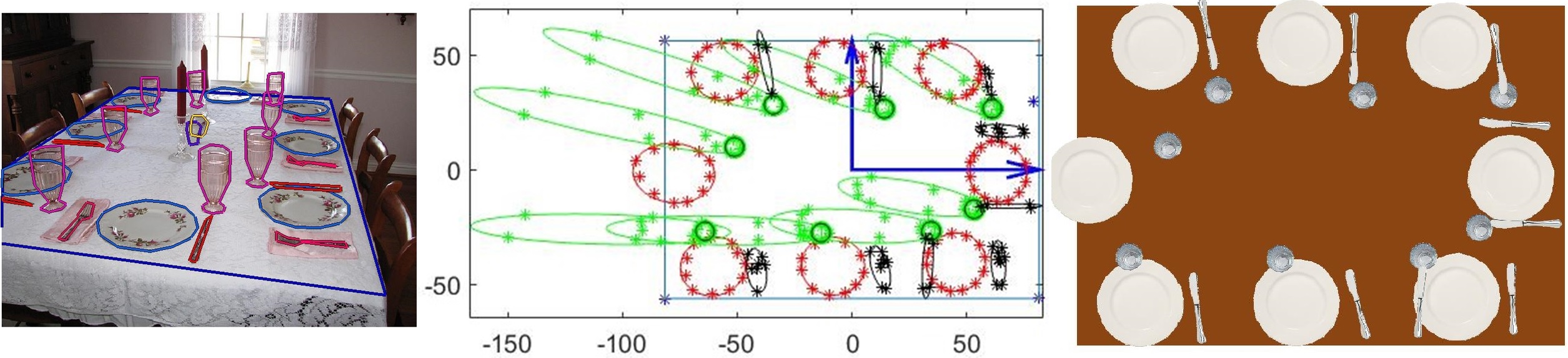}
\caption[An example annotated image and its tov-view visualization]{{\bf Left}: an annotated image from the table-setting dataset. {\bf Middle}: back-projection of table (in blue), plates (in red), glasses (in green), and utensils (in black). The unit of axes is centimeter. {\bf Right}: top-view visualization of the table-setting (all utensil instances including fork, knife, and spoon are shown by knife icon).}
\label{fig:examples3in1}
\end{figure}

We collected and annotated  the ``JHU Table-Setting Dataset,'' which consists of about 3000 images of dining room table settings
with more than 30 object categories. The images in this dataset were
collected from multiple sources such as Google, Flickr, Altavista,
\etc. Figure~\ref{fig::JHUDatasetSnapshot} shows a snapshot of the dataset, which is made publicly available \footnote{Available at: \url{http://www.cis.jhu.edu/~ehsanj/JHUTableSetting.html}}.

The images were annotated by three annotators over a
period of about ten months using the
\href{http://labelme.csail.mit.edu/}{``LabelMe''} online annotating
website ~\cite{Russell2008}.
The consistency of labels across annotators was then verified and synonymous labels were consolidated.
The annotation task was carried out with careful supervision resulting in high quality annotations, better than what we normally get from crowd-sourcing tools like Amazon
Mechanical Turk.
Figure~\ref{fig:DataSet_stat} shows the annotation histogram of the 30 most annotated categories. The average number of annotations per image is about 17.

To estimate the homography (up to scale) at least four pairs of corresponding points are needed according to the Direct Linear Transformation (DLT) algorithm~\cite[p. 88]{Hartley-Zisserman04}. These four pairs of corresponding points were located in the image coordinate system by annotators' best visual judgment about four corners of a square in real world whose center coincides with the origin of the table (world) coordinate system. We are able to undo the projective distortion due to the perspective effect by back-projecting the table surface in the image coordinate system onto the world coordinate system. The homography matrices are scaled appropriately (using object's typical sizes in real world) such that after back-projection the distance of object instances in the world coordinate system (measured in meters) can be computed. Figure~\ref{fig::rectifiedImages} shows two typical images from this dataset and their rectified versions. Clearly, the main distortions occur for objects which are out of the table plane.

Each object instance was annotated with an object category label plus an enclosing polygon. Then, an ellipse was fit to the vertices of the polygon to estimate the object's shape and pose in the image plane. Figure~\ref{fig:examples3in1} (left) shows an example annotated image; Figure~\ref{fig:examples3in1} (middle) shows the corresponding back-projection of vertices of annotation polygons for plates (in red), glasses (in green), and utensils (in black). Note that non-planar objects (\eg glass) often get distorted after back projection (\eg elongated green ellipses) since the homography transformation is a perspective projection from points on the table surface to the camera's image plane. Hence, we estimated the base of vertical objects (shown by black circles in the middle figure) to estimate their location in the table (world) coordinate system since the center of fitting ellipse to the back-projection of such objects' annotation points is not a good estimate of their 3D location in the real world. Figure~\ref{fig:examples3in1} (right) shows top-view visualization of the annotated scene in the left using top-view icons of the corresponding object instances for plates, glasses, and utensils (note that all utensil instances are shown by top-view knife icons).

\begin{figure}[t]
\centering
\includegraphics[width=\linewidth]{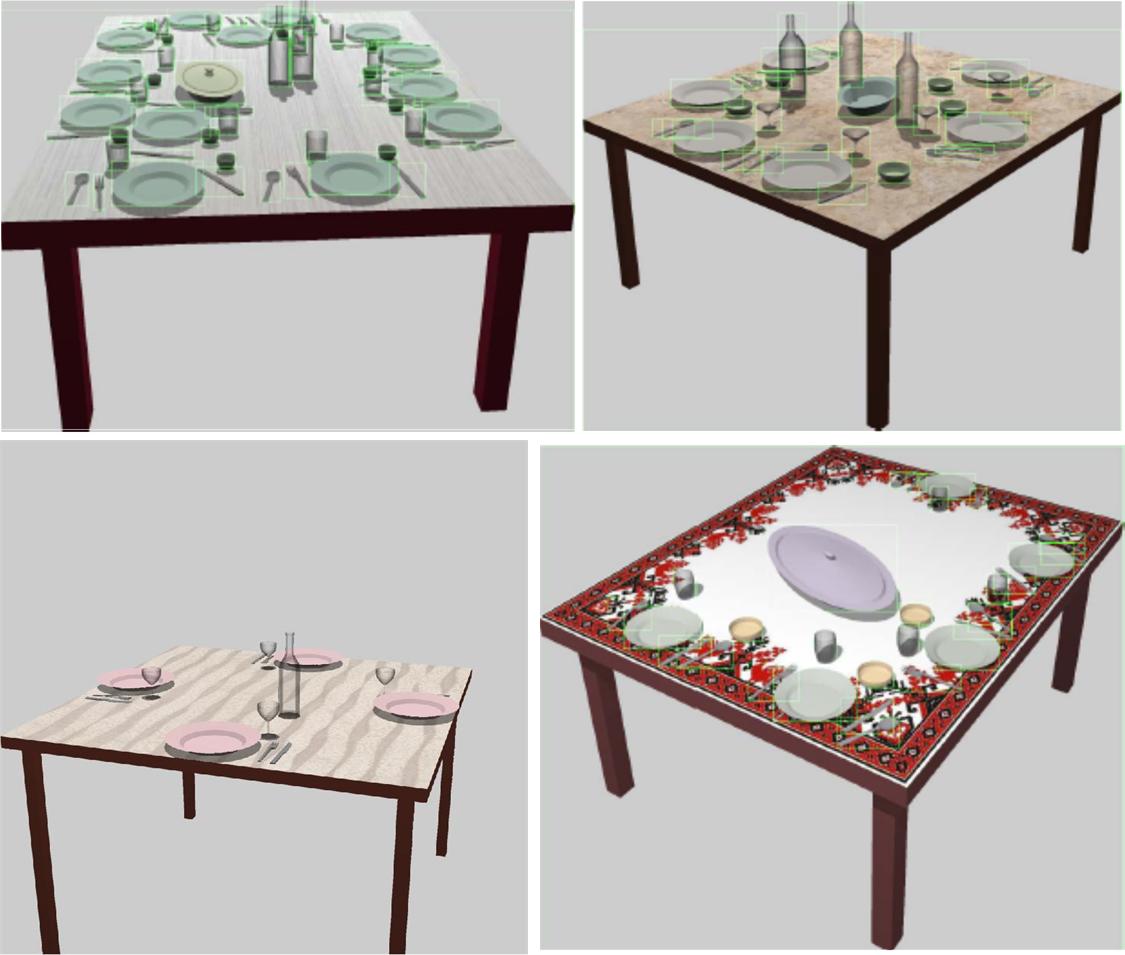}
\caption[Synthetic table-setting scene samples]{Synthetic table-setting scene samples.}
\label{fig::syntheticImages}
\end{figure}

We also utilized a synthetic table-setting scene renderer for verification purposes. This synthetic image renderer inputs the camera's calibration parameters, six rotation and translation camera's extrinsic parameters, table length and width, and 3D object poses in the table's coordinate system and outputs the corresponding table setting scene. Figure~\ref{fig::syntheticImages} shows some synthetic images generated by this renderer.

\section{Experiments and Results}
\label{sec:experiments}
\subsection{Classifier Training}

\subsubsection{CatNet and ScaleNet}
We fine-tuned CatNet using a set of 344,149 patches. The training set
contained 170,830 patches from the ``No Object'' category, 36,429
patches from the ``Plate'' category, 2,074 patches from the ``Bottle''
category, 49,401 patches from the ``Glass'' category, and 85,415
patches from the ``Utensil'' category. If a patch includes multiple
object instances, it is repeated in the training set, once for
each instance. The train and test patches were extracted from the
``JHU Table-Setting Dataset'' using the image partitioning scheme
explained in section~\ref{sec:annocell.hierarchy}. The ``No Object'' category
patches were selected from the set of annocell patches whose overlap
with the table area is less than $10\%$ of the patch. The number of
such background training patches was chosen to be twice the
number of patches from the most frequent category (utensil). We evaluated the performance of CatNet on a test set of 62,157 patches. Results from the raw output of CatNet are provided in
Table~\ref{tab:size_shape}, which shows the average scores in the
vector of SoftMax scores returned by CatNet's when it is applied to a patch
from the corresponding class at different levels of the
hierarchy. Unsurprisingly, for each category, the scores for that
category increase as the patch size decreases (usually resulting in tighter patches to objects) when the category is
present in the patch, which leads to higher classification accuracy being achieved for patches from finer levels of the annocell hierarchy.

\begin{table}[t]
\centering
\caption[Average CatNet score at different levels of resolution.]{\small{Average
score at different levels of resolution of the annocell hierarchy
when CatNet is applied to an input patch from the corresponding class.}}
\begin{tabular}{lllll}
  \topline
  \hline
  \headcol Category & Level-0 & Level-1 & Level-2 & Level-3 \\
  \midline
  ``No Object'' & 0.31  &  0.72  &  0.96  &  0.99 \\
  \rowcol \smash{{``Plate''}} & 0.32  &  0.34  &  0.39 &   0.44 \\
  ``Bottle''    & 0.08  &  0.19  &  0.31  &  0.36 \\
   \rowcol \smash{{``Glass''}} & 0.33  &  0.44  &  0.56  &  0.68 \\
  ``Utensil''    & 0.48  &  0.54  &  0.71  &  0.81 \\
  \bottomlinec
\end{tabular}
\label{tab:size_shape}
\vspace{10mm}
\end{table}

We fine-tuned ScaleNet on 171,395 patches. Each patch was labeled by one
label $l \in \{1,2,3,4\}$, respectively associated to the closest scale ratios in
$\{0.1, 0.35, 0.65, 1\}$, the number of patches in each category being
42,567, 82,509, 37,443 and 8,876.

\begin{figure}[t]
\centering
\includegraphics[width=0.47\linewidth]{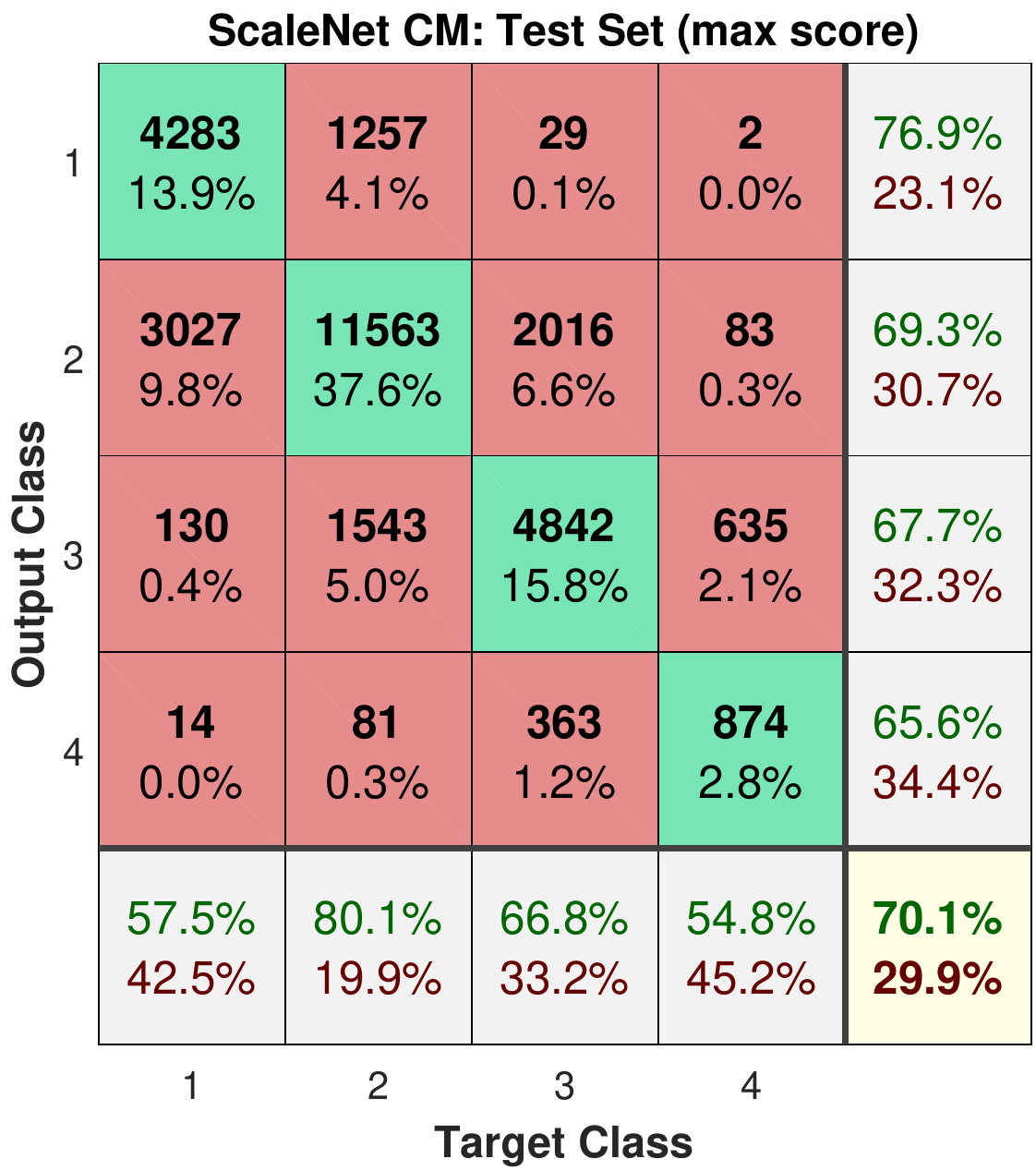} \hspace{2mm}
\includegraphics[width=0.47\linewidth]{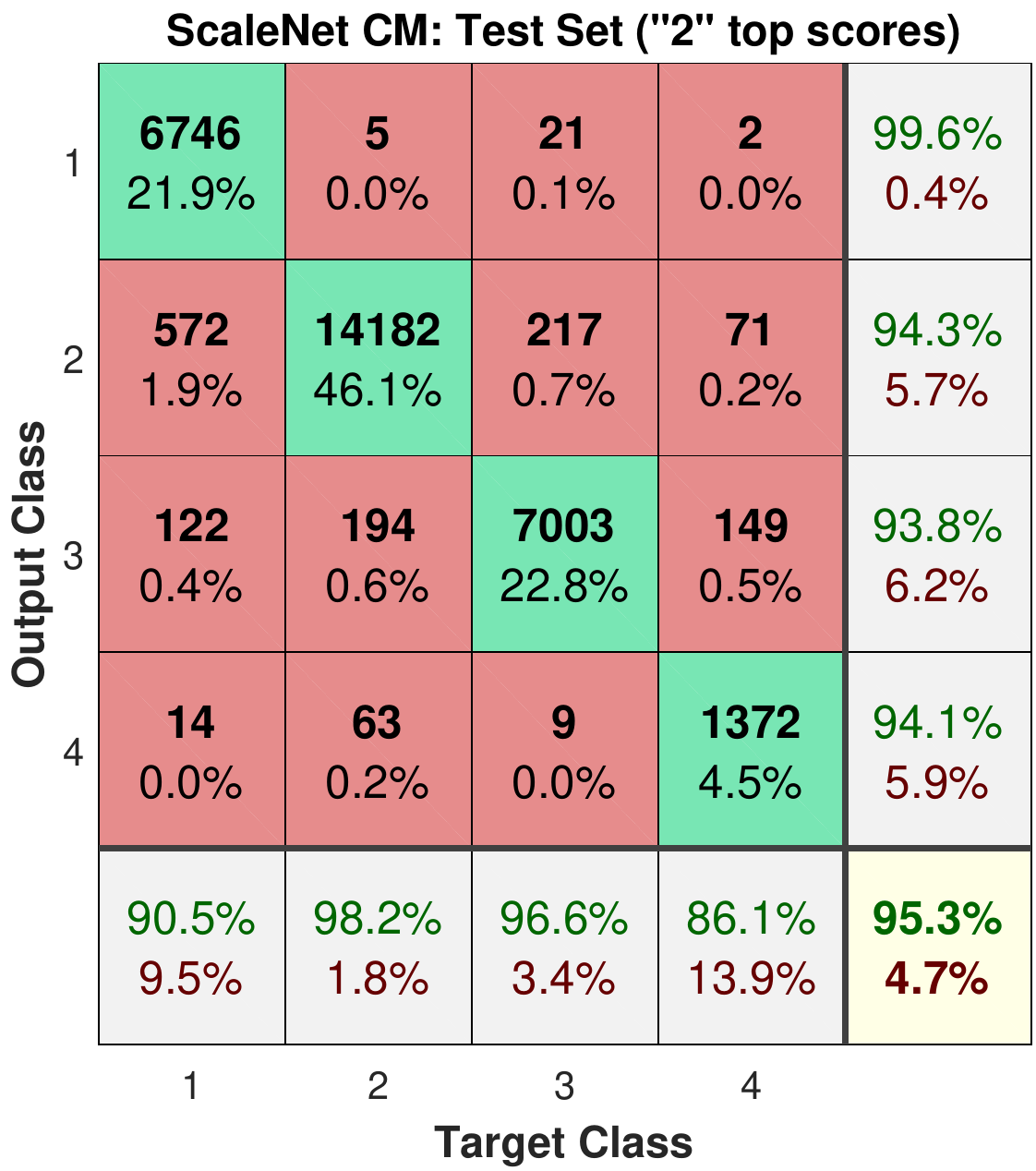} \\
\vspace{0mm}
\caption[ScaleNet confusion matrix.]{\small{ScaleNet confusion matrix on ``training'' and ``test'' set considering both max score classification and top-2 classification.}}
\label{fig:ScaleNet_train_test_CM}
\end{figure}

We evaluated the performance of ScaleNet on a test set of 30,742 patches. Figure~\ref{fig:ScaleNet_train_test_CM} shows confusion matrices for test set in two cases of classification based on the maximum score class and top-2 score classes. A match is declared in the case of top-2 score classification if the true class is among the top two scores. It can be seen that the most common mistakes are made between consecutive classes which makes sense since consecutive classes are associated with consecutive scale ratios which have closer output distributions.

\begin{figure*}[t]
\centering
\includegraphics[width=0.99\linewidth]{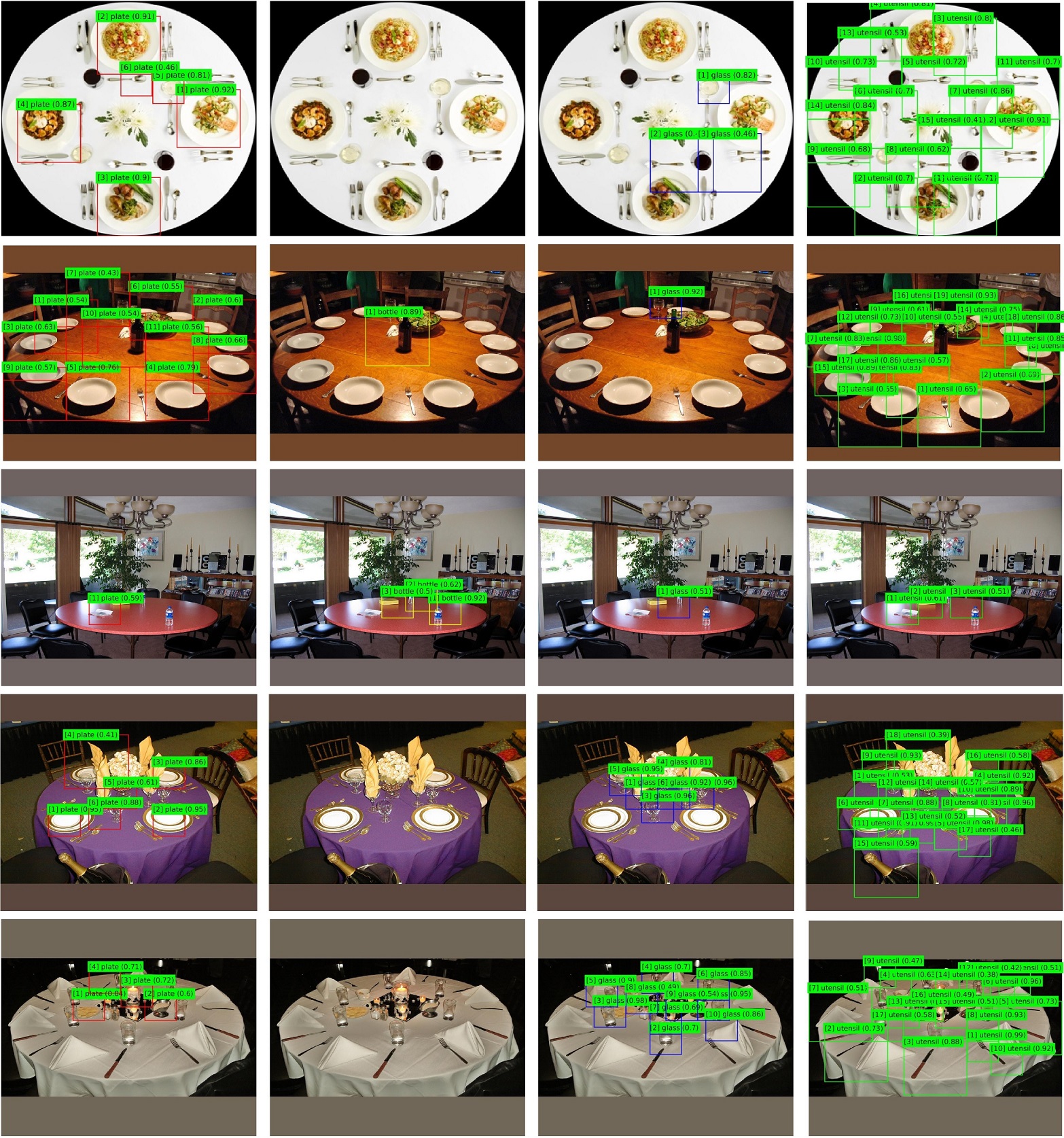}
\caption[CNN classifier detections.]{\small{CNN classifier detections for ``plate'', ``bottle'', and ``glass'', and ``utensil'' categories from left to right. The ordinal numbers in brackets represent the confidence rank of detections per category and the fractional values in parentheses indicate the scale ratio of detections.}}
\label{fig:CNN_SamplesSet}
\end{figure*}

\subsubsection{SceneNet}
\label{SceneNetSubSec}

\begin{figure*}[h]
\centering
\includegraphics[width=0.99\linewidth]{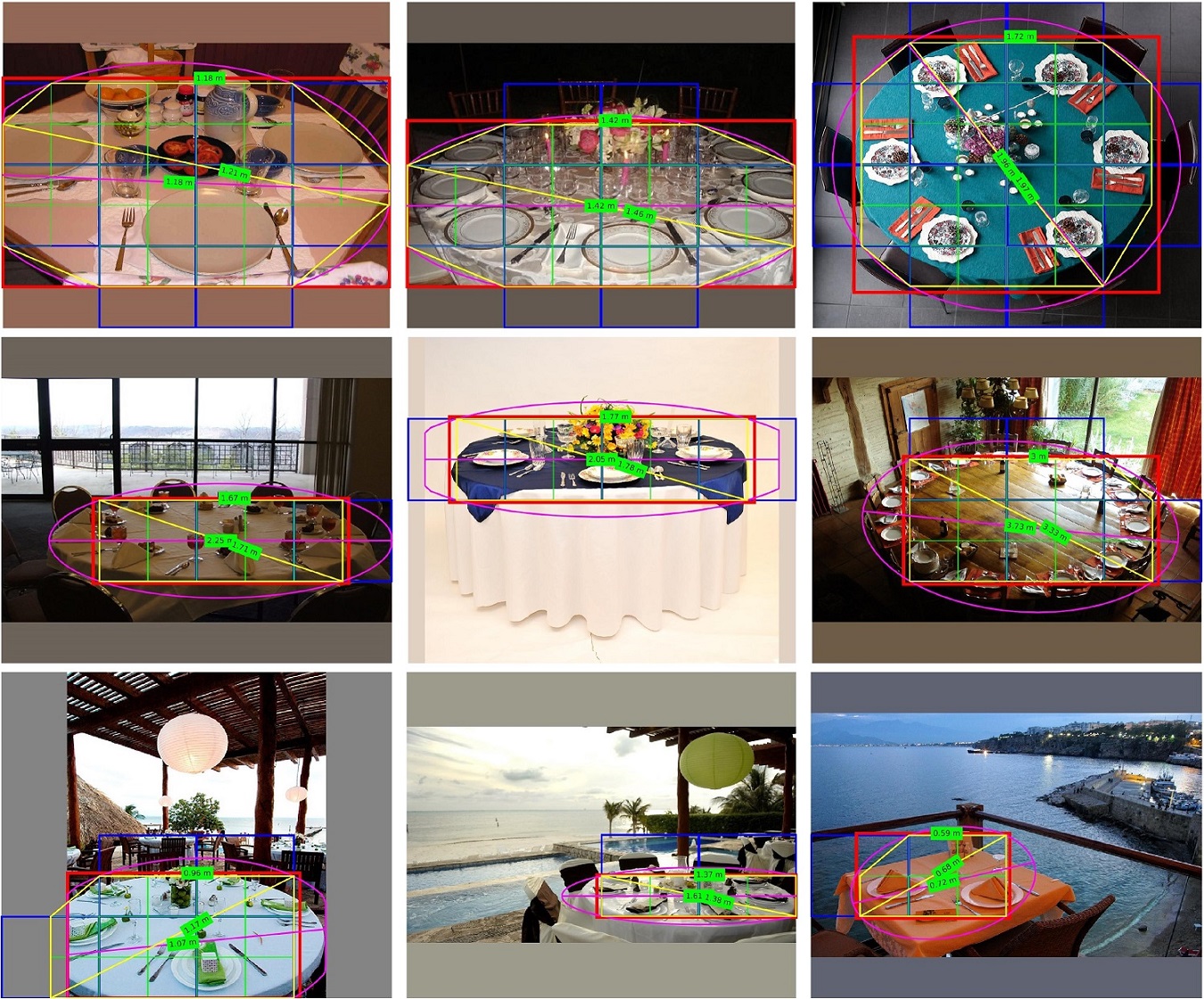}
\caption[Table estimation examples using GeomNet.]
{\small{Table detection examples using TableNet: The fitting polygon,
rectangle, and ellipse to the corner the points of patches at level--3
which were classified as table are shown in yellow, red, and magenta,
respectively. The blue and green boxes show patches from,
respectively, level-2 and level-3 classified as table. The estimated
table size (in meters) based on each shape is shown on the green text
boxes.}}
\label{fig:TableEst}
\end{figure*}
\begin{figure*}[h]
\centering
\includegraphics[width=0.99\linewidth]{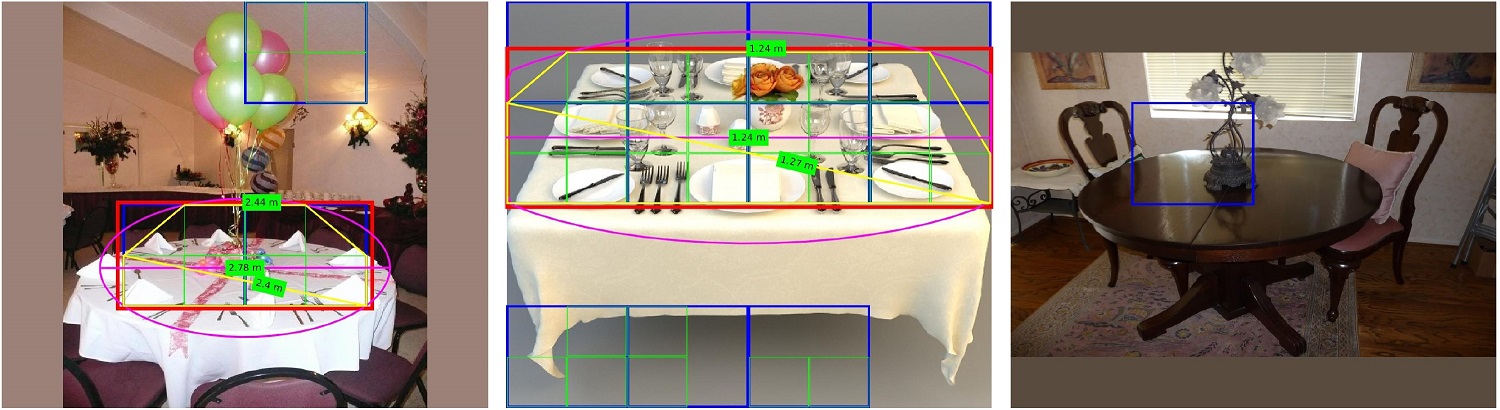}
\caption[Table estimation examples using TableNet (noisy).]
{\small{Noisy table detection examples using TableNet: The top row
shows two examples with off-table false positives which were
suppressed by considering the region with the maximum number of
connected positive detections. The bottom row shows two poor table
detection examples, perhaps due to the insufficient texture on the
tables.}}
\label{fig:TableEstNoisy}
\end{figure*}
\begin{figure}[h]
\centering
\includegraphics[width=0.48\linewidth]{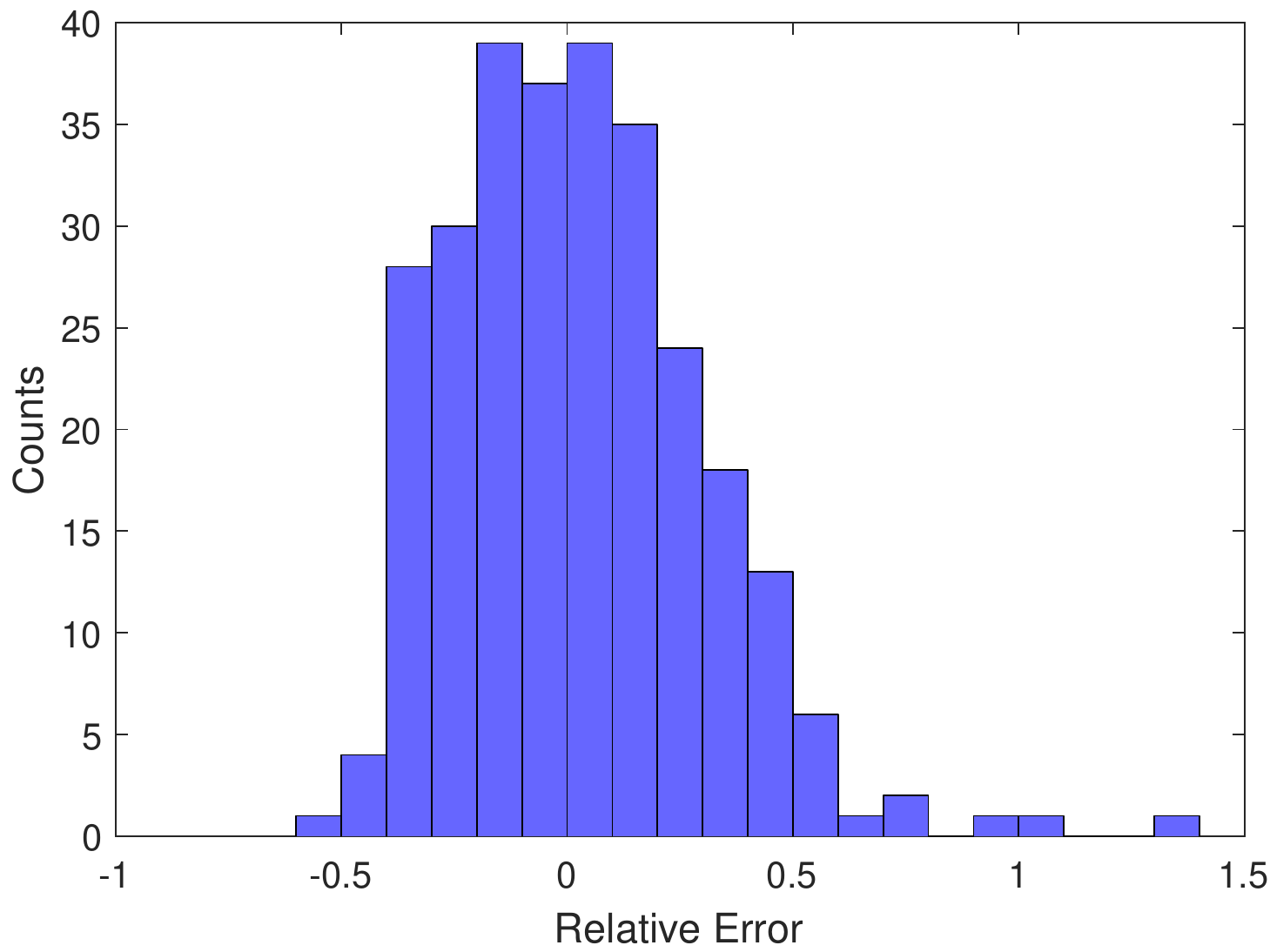}
\includegraphics[width=0.48\linewidth]{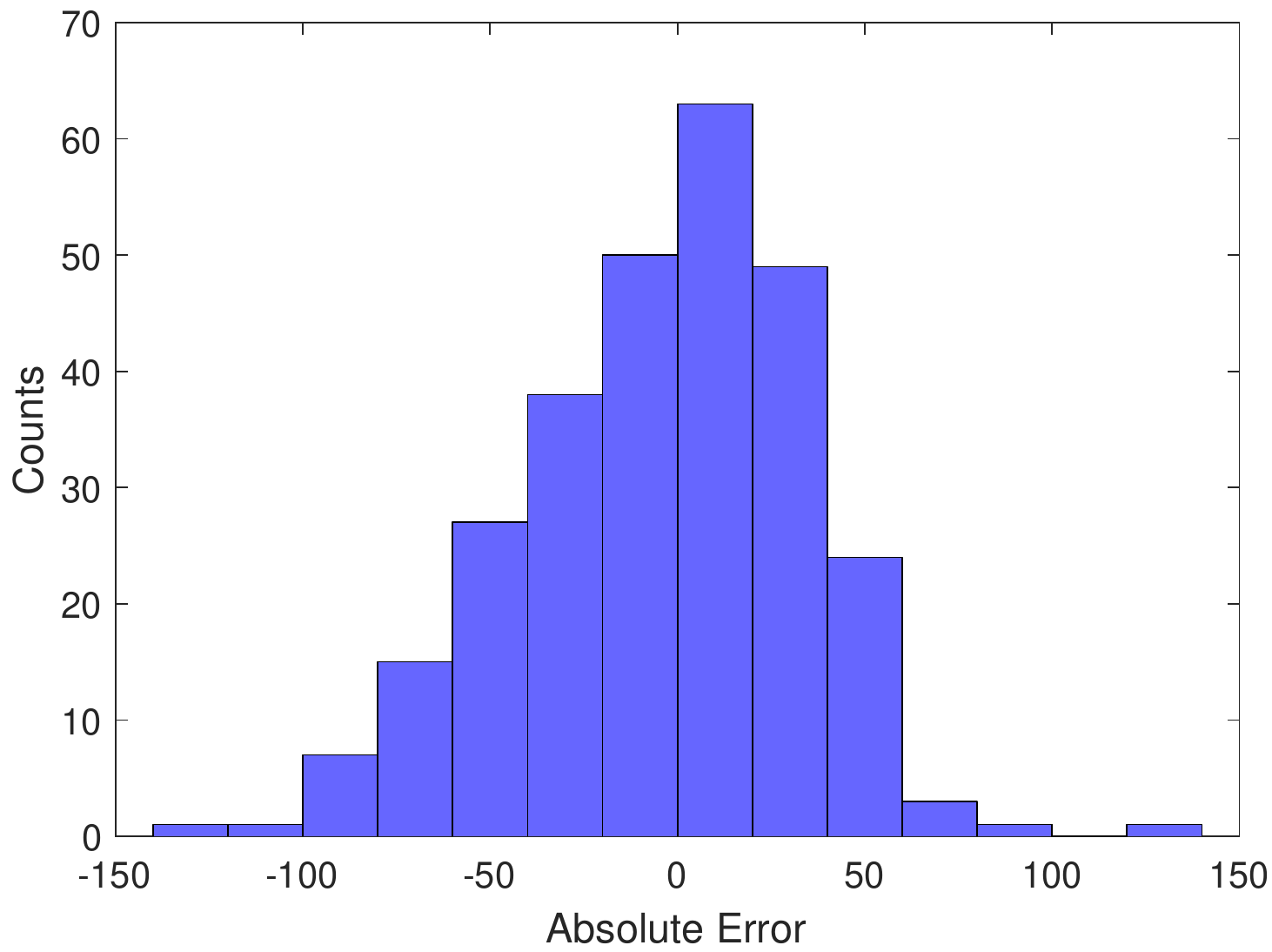} \\
\vspace{-1mm}
\caption[Table--size estimator error histogram.]{\small{Histogram of the relative (left) and absolute (right) error made by the table--size estimator.}}
\label{fig:TableEstHist}
\end{figure}

The main component of SceneNet is a CNN that detects whether a patch is
part of the table area.  We used 270,410 training patches (including 153,812
background and 116,598 table), and 38,651 test patches (including 18,888
background and 19,763 table). The background and table patches are
defined by having, respectively, at most $10\%$ and at least $50\%$
(of the patch) overlap with the table surface area. All of the
training and test patches were selected from level-2 and level-3 of
the annocell hierarchy.

We classify a level-3 patch as part of the table if both its
associated CNN and the one run on one of the level-2 patches that
contain it report a positive detection. The final table area
prediction, $\hat M$, is defined as the convex hull of the largest
connected component of the union of detected level-3 patches.
Figure~\ref{fig:TableEst} shows the estimated table area for some
example images. Figure~\ref{fig:TableEstNoisy} (left and middle) shows
two examples in which misdetected off-table patches are removed after
post-processing. Figure~\ref{fig:TableEstNoisy} (right) shows a poor
table detection example which seem to happen due to the lack of sufficient
texture on the tables. We tested our table detector on 284 images and
observed fewer than 5 poor table detections.

We estimate the table size (in 3D) by appropriately scaling the diameter length of its convex hall. The  scale was
calculated by running ScaleNet on patches from level 2 classified as table, and assuming that the table-setting objects have
an average size of $20$cm.
Figure~\ref{fig:TableEstHist} shows the histogram of the absolute and
relative errors made by our table size estimator. We calculated the true table size by
back-projecting the annotated table surface using the homography that
was estimated from the annotation of the images. The histogram is
centered roughly around 0 meaning that our table size estimator is
relatively unbiased.

\begin{figure}
\centering
\includegraphics[width=0.99\linewidth]{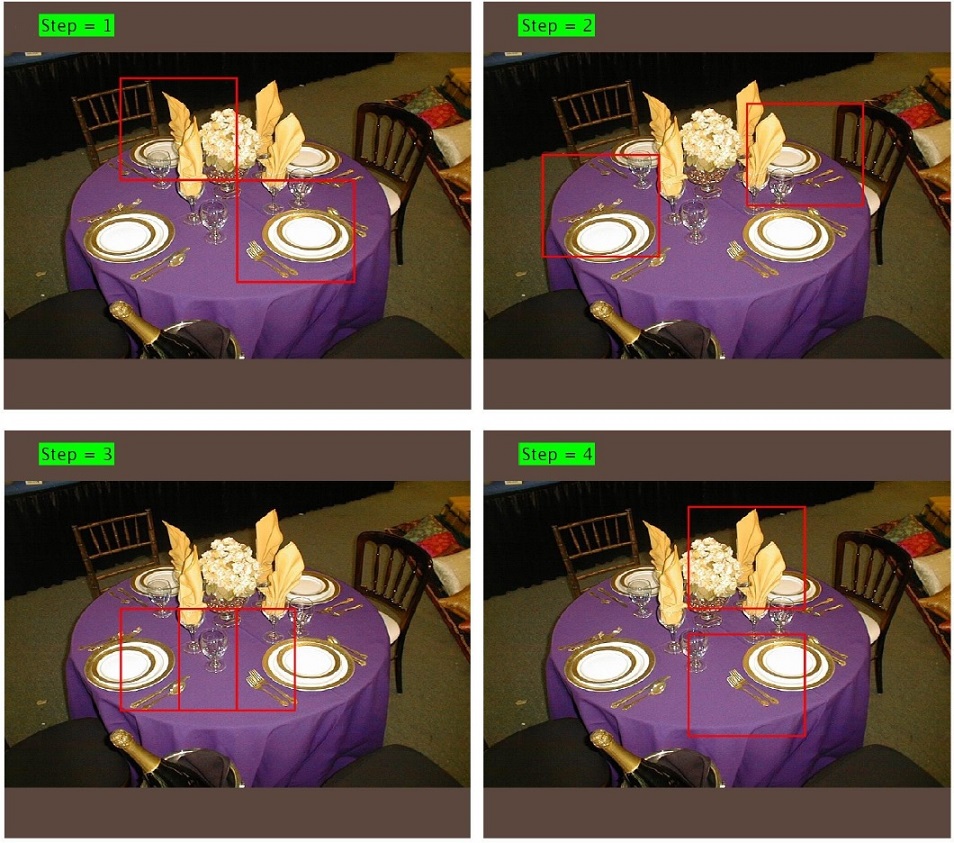}
\caption[Questions at ``early'' IP steps.]{\small{Questions at ``early'' IP steps.}}
\label{fig:IPQ_early}
\end{figure}
\begin{figure}
\centering
\includegraphics[width=0.99\linewidth]{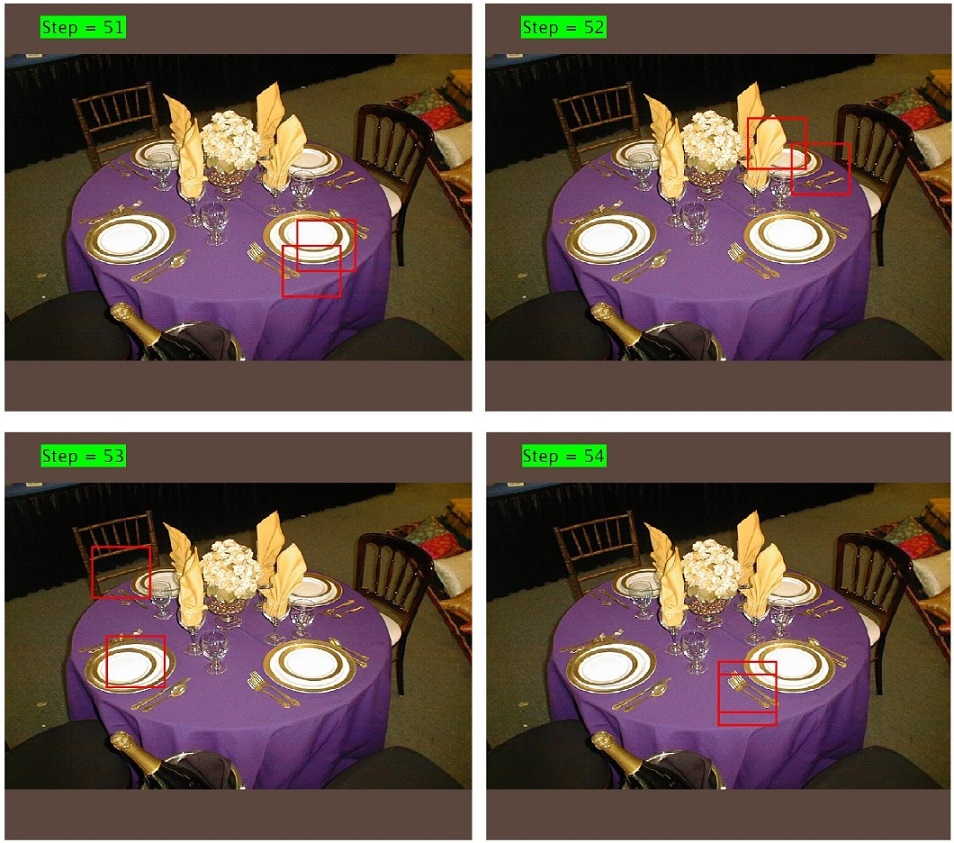}
\caption[Questions at ``middle'' IP steps.]{\small{Questions at ``middle'' IP steps.}}
\label{fig:IPQ_middle}
\end{figure}
\begin{figure}
\centering
\includegraphics[width=0.99\linewidth]{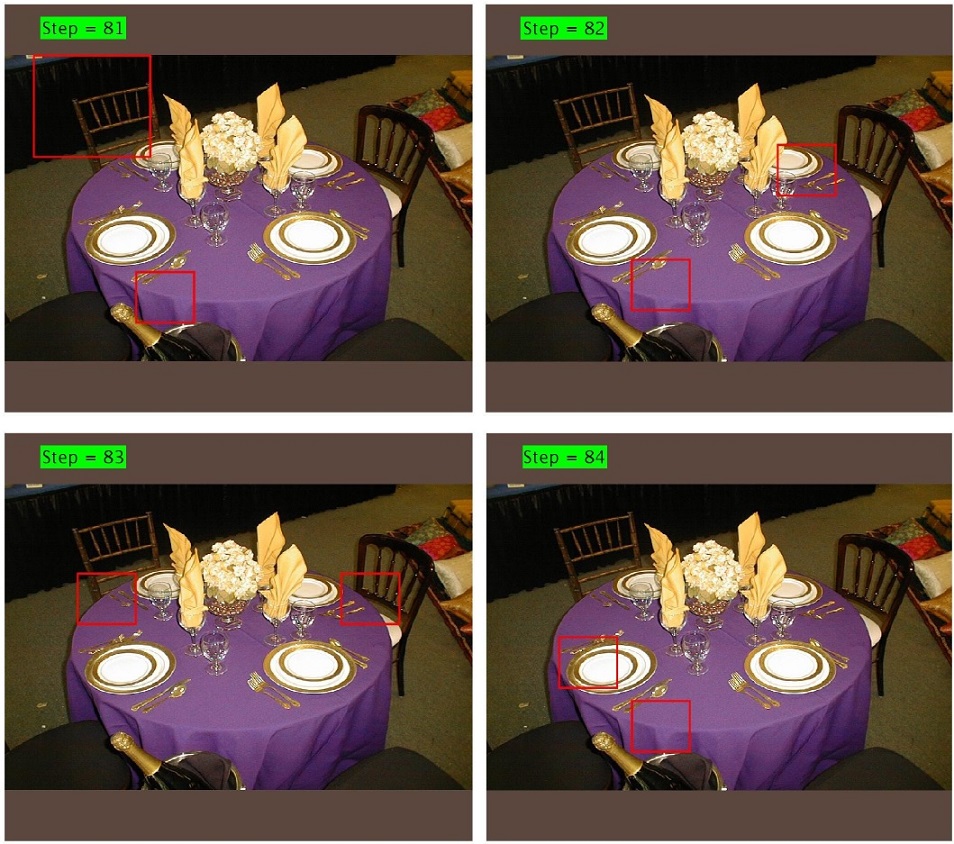}
\caption[Questions at ``later'' IP steps.]{\small{Questions at ``later'' IP steps.}}
\label{fig:IPQ_later}
\end{figure}

\begin{figure}[t]
\centering
\includegraphics[width=0.4\linewidth]{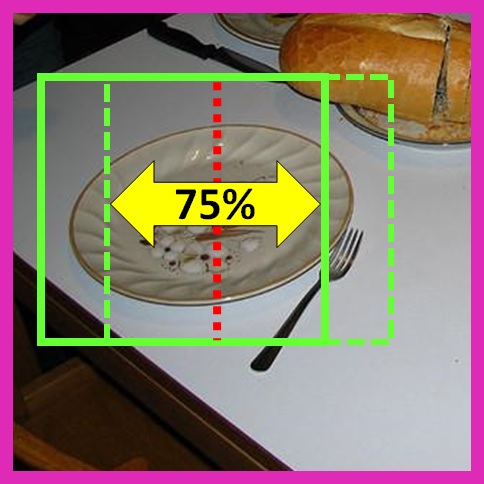} \\
\caption[An example showing a plate instance not captured by annocells from the possibly best-fitting level.]{\small{An example showing a plate instance not captured by annocells from the possibly best-fitting level (in green).}}
\label{fig:RelaxedCondition}
\end{figure}

\subsection{IP Experiments}

Conditional inference on the posterior distribution $p(\xi|{\bf E}_k)=
P(Z=z,S=s,W=w|{\bf E}_k)$ given the
accumulated evidence after $k$ steps of IP, was described in section
\ref{Conditional_Sampling} (including, in particular, approximations
made to the sampling of the scene geometry and camera properties). The
templates we used for the geometry $S$ are square tables whose sizes range
from 0.9 to 2.7 meters with 20cm intervals. We selected the template
closest to the estimated table size and its two nearest
neighbors (or one neighbor if the closest table size is 0.9 or 2.7). For each of them, we sampled 10 homographies which are
consistent with the detected table surface area (described in
section~\ref{SceneNetSubSec}).

\begin{figure*}[t]
\centering
\includegraphics[width=0.99\linewidth]{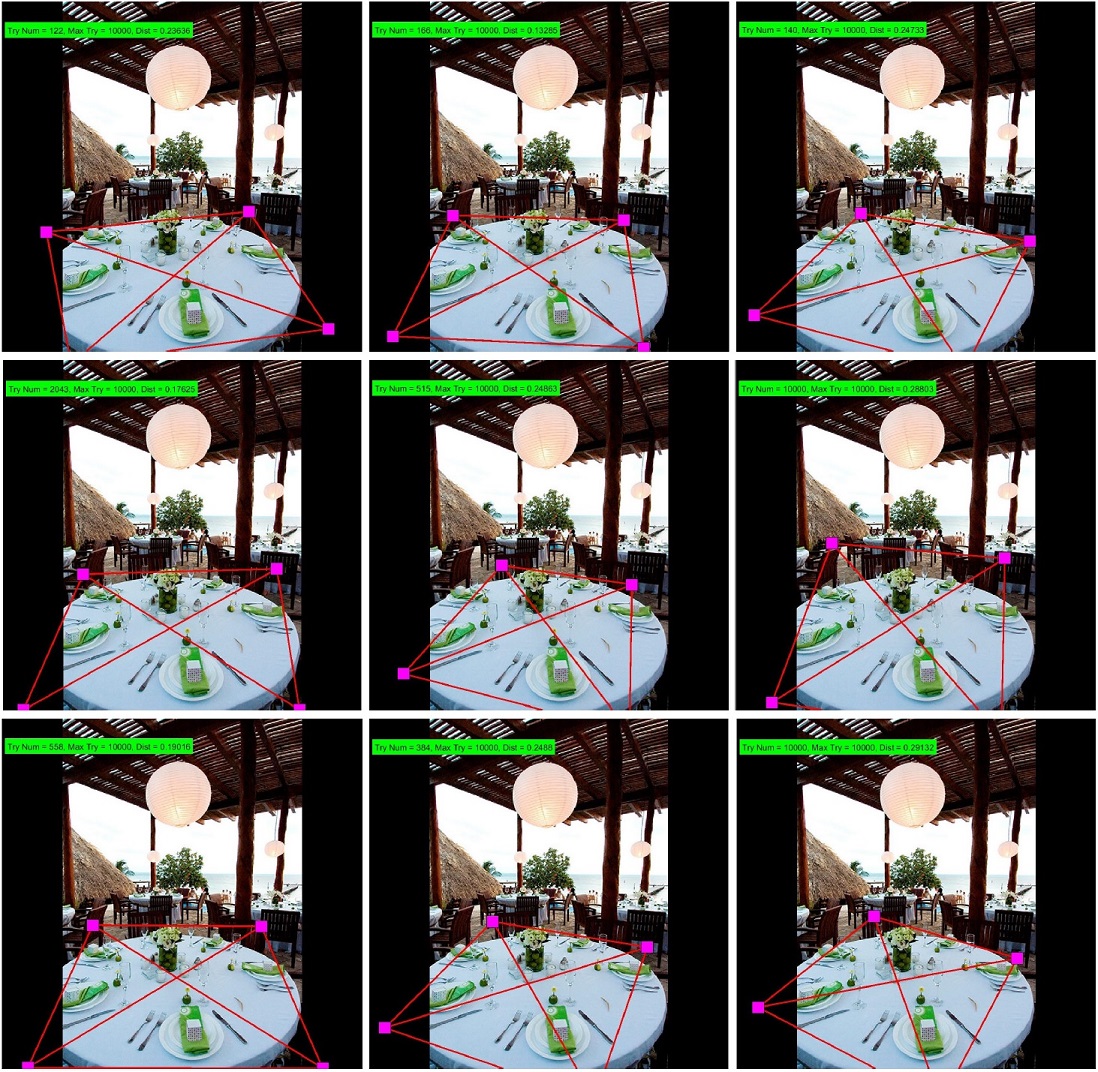}
\caption[Consistent homography samples (set 1).]{\small{Consistent homography samples satisfy condition~\eqref{fitting_condition}.}}
\label{fig:HomSamples1}
\end{figure*}

To generate homography samples that conform with the detected table area, assume a
rectangular table with length $L_s$ and width $W_s$ whose four corner
points are $(-L_s/2,-W_s/2)$, $(-L_s/2,W_s/2)$, $(L_s/2,-W_s/2)$, and
$(L_s/2,W_s/2)$. We draw samples from the distribution on camera
parameters $p(W)$ proposed in section~\ref{sec:camera_model} and
calculate the corresponding homography matrix. Then, we project the
four corners of the table to the image coordinate system using this
homography matrix and check if the resulting polygon (quadrilateral)
fits well to the detected table area using a similarity measure for
2D--shapes. We declare a ``good fit'' between two shapes $A_1$ and
$A_2$ if their distance defined as $d(A_1,A_2) = |(A_1 \cup A_2)-(A_1 \cap A_2)|$ satisfies
\begin{align}
\label{fitting_condition}
d(A_1,A_2) < 0.25 \min(|A_1|,|A_2|).
\end{align}
In an attempt to efficiently sample the homography (camera parameter)
distribution that is consistent with the detected table area, we first
try to find a set of camera parameters that result in a table
projection meeting a relaxation of~\eqref{fitting_condition}, namely
$d(A_1,A_2) < 0.4 \hspace{0.5mm} \min(|A_1|,|A_2|)$, and as soon as we
find such a sample we start to greedily fine--tune the camera
parameters to finally satisfy~\eqref{fitting_condition}. During
fine-tuning we randomly choose one camera parameter and change it
slightly by sampling a normal distribution with small variance
centered at the previous value; we accept this change if it resulted
in a smaller distance $d(A_1,A_2)$. We try a total of $10,000$
homographies obtained by sampling the camera model $p(W)$
(to satisfy the relaxed condition) or fine-tuning of parameters
$W$ (to satisfy~\eqref{fitting_condition}) and exit the loop
as soon as~\eqref{fitting_condition} is met; otherwise, if the
condition~\eqref{fitting_condition} was not met during $10,000$
trials, we output the camera parameters resulting in the minimum
$d(A_1,A_2)$. Figure~\ref{fig:HomSamples1} shows some example consistent homography samples.

Recall that at step $k$, IP maximizes the mutual information
$\mathcal{I}(X_q, Y_{\mathcal Q}| \mathbf E_{k-1})$ over queries $q
\in {\mathcal Q}$ and that this mutual information is the difference\\
$H(X_{q}| \mathbf E_{k-1}) - {H}(X_{q}|Y_{\mathcal Q}, \mathbf
E_{k-1})$ (see \eqref{EPdecision4}).  Moreover, under our conditional
independence assumptions, this reduces to the entropy of a mixture
minus a mixture of entropies where in both cases the mixture weights
are the conditional probabilities of the annobit $Y_q$ given the
evidence.  In the current case, the queries are indexed by the
annocells $A \in {\mathcal A}$, where $Y^{cat}_A$ assumes sixteen
possible values corresponding to the possible subsets of the four
object categories.  There are also scale annobits in correspondence
with the classifiers $X^{sc}_A$ but we do not consider these in the
selection of queries; of course each time we execute a CatNet
classifier for an annocell $A$ we also execute the corresponding
ScaleNet classifier for $A$ and {\it both} the CatNet and ScaleNet
results are part of the evidence.  Once the weights
$P(Y^{cat}_A=y|{\bf E}_{k-1})$ are computed by sampling (see below)
from the posterior, we can immediately evaluate the mixture of entropies
since the entropy of the Dirichlet distribution has a closed-form
solution.  For the entropy of the mixture, namely the entropy
of the mixture of $\sum_y Dir(X^{cat}_A|Y^{cat}_A=y)P(Y^{cat}_A=y|{\bf E}_{k-1})$
of Dirichlet densities, we estimate the integral by Monte Carlo integration. To generate a sample from the mixture distribution for the Monte Carlo integration, we first select one of the 16 Dirichlet densities with probabilities according to the posterior $P(Y^{cat}_A=y|{\bf E}_{k-1})$ and generate a sample from the selected Dirichlet distribution. Given generated samples from the mixture distribution we then evaluate negative logarithm of the mixture distribution at the generated samples and average to get an estimate of the entropy of the mixture. A similar approach can be taken to estimate the entropy of a Dirichlet distribution but since there is a closed-form solution for the entropy of Dirichlet distribution we used the closed-form solution in computing mixture of entropies. Nevertheless, by comparing the closed-form calculation of the Dirichlet distribution entropy and its Monte Carlo integration estimation we got insight about the appropriate number of mixture samples to reasonably estimate the entropy of mixture.

Turning back to the annobit posterior, we determine the states of the
annobits from posterior samples $(\xi,s,w)$ by projecting the
3D samples $z$ to the image coordinate system using the sampled
homography. More specifically, the projection of the sampled locations
on the table plane in 3D obviously allows us to answer any queries
about locations in the image plane appearing in the definition of an
annobit.  However, in order to determine what instances of objects are
contained in a given annocell, and to measure the average sizes of the
instances present, we need an estimate of the set of pixels which
constitute the image realization of each instance sampled.  For plates
and utensils, which are effectively 2D, we simply use the projected
circle for plates and projected ellipse for utensils, which of course
are again ellipses in the image plane.  For glasses and bottles, which
are three-dimensional, we know the image representation is larger than
the image ellipse obtained by projecting the base circle determined by
the sample. Also, the projection of these objects in 2D is oriented perpendicular to the orientation of their base circle projection. Hence, we estimate the projection we would obtain for instances from these categories with a fully 3D to image mapping by moving the center of projection from the center of projected base upward (in the image) and along a vector orthogonal to the main axis of the projected base ellipse; we place the updated object center at a distance from the projected base center equal to half of its size where the size is proportional to the main diameter of the projected base.

We ran IP on a dataset of 284 images. In each step of IP, two
most informative questions corresponding to annobits with maximum mutual informations were asked, \ie two patches were processed by CNNs.
Figure~\ref{fig:IPQ_early} shows the annocells selected in the
first four steps of IP for a given test image. Figure~\ref{fig:IPQ_middle},~\ref{fig:IPQ_later} show the
selected annocells at later IP steps. We can see that the patches
selected later are usually from the finer levels which follows a coarse-to-fine scene analysis paradigm. However, it is completely plausible, and actually happened during our experiments, to go back again to a coarser question after asking a sequence of finer questions. Analogously, we as humans may focus on a particular area while analyzing a scene and then depending on the collected evidence can zoom out and collect evidence at a coarser level.

It is worthy to mention the difference between the IP selection criterion in~\eqref{EPdecision2} and the approximate criterion in~\eqref{EPdecision3} in terms of the resolution level of selected patches. According to our experiments, the approximate selection criterion in~\eqref{EPdecision3} usually starts with selecting coarser patches compared to the IP selection criterion in~\eqref{EPdecision2}; more specifically the approximate criterion starts with level-1 whereas the exact criterion starts with level-2 (the reason of not starting with level-0, in the approximate criterion, is that in level-0, which is basically the whole image, most of categories exist. Therefore, analyzing the whole image will not result in much information gain if we are considering only one type of scene category). This is mainly due to the fact that the approximate criterion ignores the error rates of classifiers $X_q$ at the selection stage by replacing $X_q$ with $Y_q$. We know that our classifiers are more accurate at finer levels which leads to encouragement of their selection when using the IP criterion in~\eqref{EPdecision2}. Note that in both criterions the questions selected at the early steps are usually coarser and they progressively refine (coarse-to-fine analysis). This is an interesting contrast between the two criterions. In support of the IP criterion in~\eqref{EPdecision2}, assume Alice walks into a bookstore in Brooklyn, where Bob is the Bookstore clerk, in search for a novel that she does not remember its title. Bob wants to find the book that Alice is looking for by asking questions that are most informative to him and at the same time Alice can provide an answer to them. There is no point in asking a very informative question if Alice cannot provide an accurate answer to it \eg Alice may be able to tell Bob what is the color of cover but most probably will not be able to mention the name of a few non-first characters in the novel. The IP selection criterion in~\eqref{EPdecision2} is trying to strike a tradeoff between the information gain of questions and the accuracy of the classifier at providing answer to them.

\begin{figure}[t]
\centering
\includegraphics[width=0.95\linewidth]{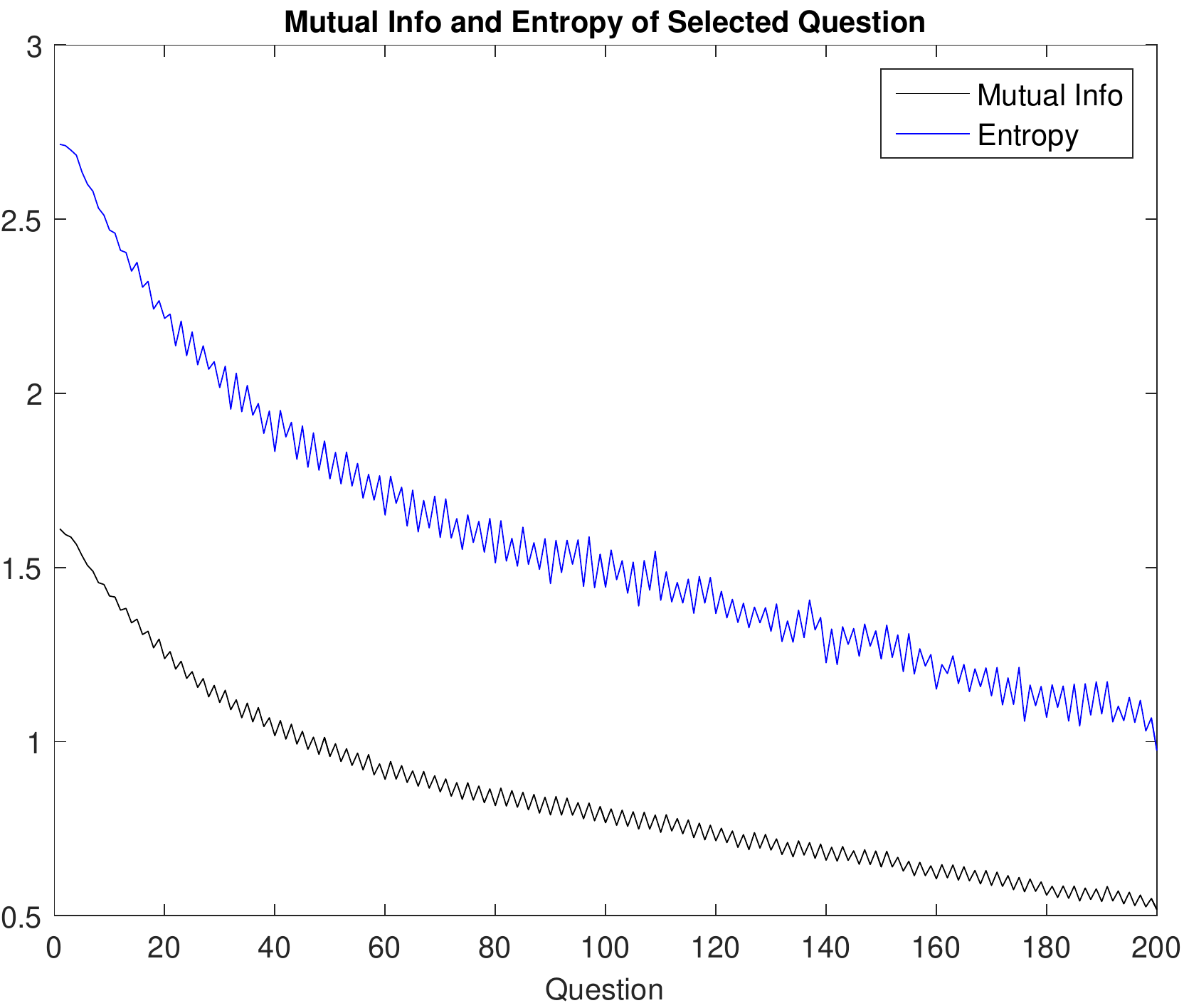} \\
\caption[Mutual Information and Entropy of Selected Questions.]{\small{Mutual Information and Entropy of Selected Questions.}}
\label{fig:IPQ_entropy}
\end{figure}

For the first 100 steps of IP, Figure~\ref{fig:IPQ_entropy} shows the
maximal mutual information $\mathcal{I}(X^{cat}_{A_k},Y_{\mathcal Q}|
\mathbf E_{k-1})$ for the selected annocell $A_k$ at step $k$, and the
corresponding conditional entropy $H(Y^{cat}_{A_k}| \mathbf E_{k-1})$,
both averaged across the 284 processed images.  Hence $k=1,...,100$
but 200 classifiers are involved which explains the ripples with
period two in this figure. This is because the second most informative question asked in each step usually has slightly lower conditional mutual information compared to the most informative question of the next step. Naturally the mutual information is
smaller than the conditional entropy.

\begin{figure*}
\centering
\includegraphics[width=0.8\linewidth]{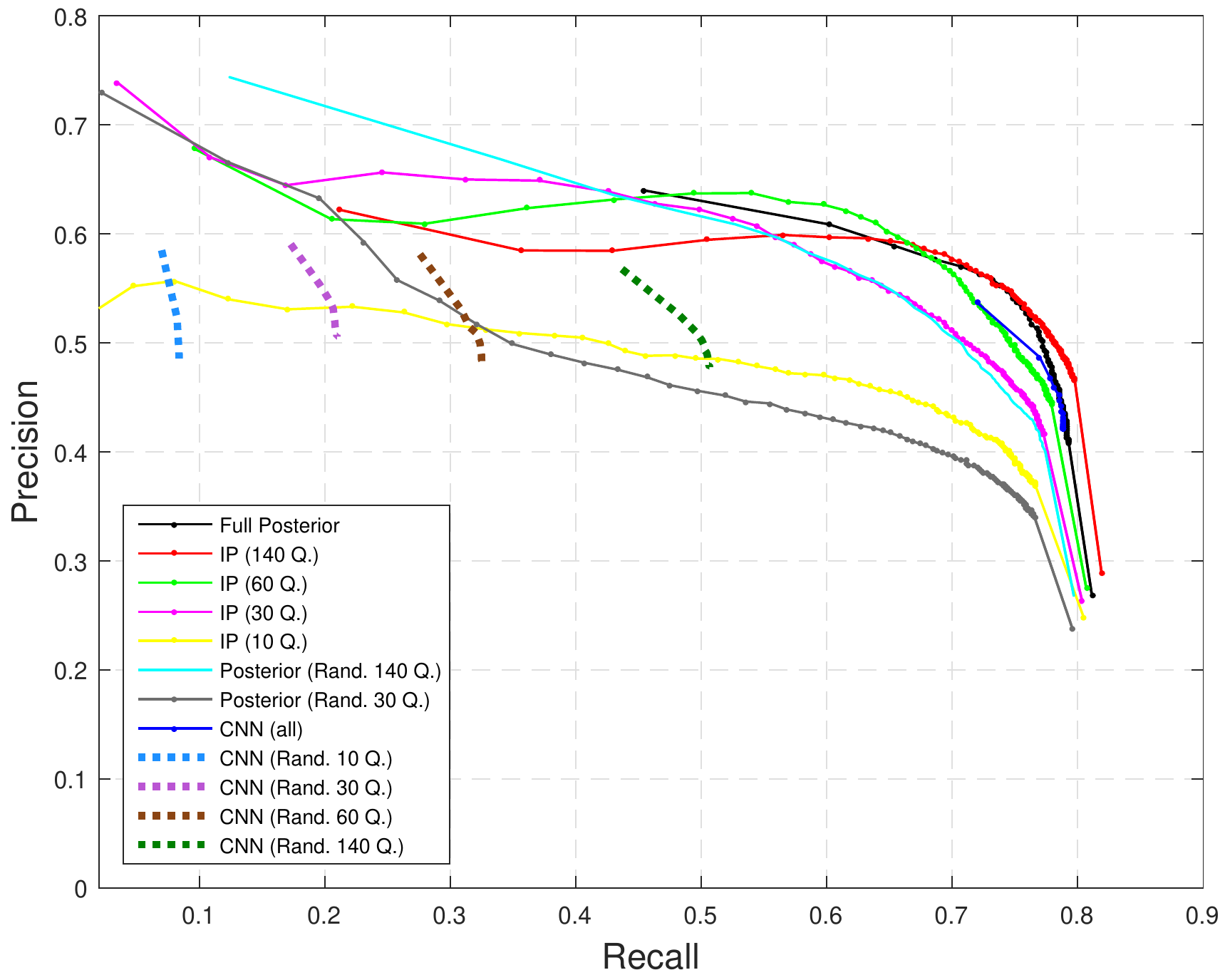} \\
\caption[Precision-Recall Curves.]{\small{Precision-Recall Curves.}}
\label{fig:PR_Curves}
\end{figure*}

In order to define and visualize the detections generated by sampling
from the 3D posterior distribution we superimpose a uniform grid of
size $25 \times 25$ on the image plane.  We earlier explained
how to associate a set of pixles with the projection of each sampled
object instance, which in turn generates a rectangular bounding box. The center of the
bounding box then falls into one of the above cells.  For each cell
and each category, we aggregate all samples from that category whose
center lies in the cell and compute the average of the top-left corner
and width/height of the corresponding bounding boxes; we take this
average bounding box as the detection for that cell. The score for
every detection is proportional to the number of 2D projections
contributing to that detection (used to compute the average). We then
run non-maximum suppression on the detections for each object category
separately; two bounding boxes are considered neighbors if their
intersection size over minimum size is greater than 0.3. This
yields a final set of scored detections, each of which
is labeled as a true positive if the intersection of the ground-truth bounding box and the estimated
bounding box is at least 0.7 of the minimum of the
two boxes and the ratio of their longest sides is between $0.5$ and $2$. Otherwise it is labeled a false positive.

\subsection{Experiments with Stand-Alone Classifiers}

In this section we consider parsing an image with the results of the
classifiers alone, i.e., without the Bayesian model.  For CatNet, from
the softmax layer output, $X^{cat}_A$, we estimate the set of categories present in
the annocell $A$ as follows. Let $S_c(A)$ denote the
weight for category $c$ with input patch $A$.  Order the weights, starting with the top one,
then add new categories until the difference between the weights of
the previous one and the new one is greater than a threshold $S_g=0.3$,
or until three categories have been selected (including the ``No Object'' category).

For ScaleNet, from the output of $X^{sc}_A$ (a sequence of weights indexed by the scale
categories), we compute an expected scale ratio $\widehat{SR}$ as a weighted
average of the top two categories, i.e., letting $s,s'$ be the top two
categories with scores $w,w'$, we take $\widehat{SR} =
(ws+w's')/(w+w')$.  We impose a selection criterion to declare an appropriate bounding box detection, ensuring in particular that objects
present in the patch occupy a significant portion of it, by requiring
that:
\begin{align}
\label{ScaleCond}
\widehat{SR} \geq 0.5 - c \hat\sigma_{SR}
\end{align}
where $\hat\sigma_{SR} = \sqrt{(w^2+w'^2)/(w+w')^3} |s-s'|$ and $c = \sqrt{2 \log
2}$. The choice made for $\hat \sigma_{SR}$ favors large differences
between the top two scales. Note that
ScaleNet returns the correct scale among its top two ratios more than 95\%
of the time when run on the test set. We also assign a score to the
output of ScaleNet, namely $S_{scale}(A) = \exp(-\max(0,0.5 - \widehat{SR})^2/2{\hat{\sigma}_{SR}}^2)$.

Finally, each patch $A$ from the annocell hierarchy is given a mixed
``Category--Scale'' score per category. The mixed score for a given
patch with scale score $S_{\text{scale}}(A)$ and the $c$-th category
score $S_c(A)$ is $S_c(A) \times S_{\text{scale}}(A)$.
We declare an
annocell patch $A$ to be the bounding box of a positive detection for the $c$-th
category if both $S_{\text{scale}}(A) \geq 0.5$ and $S_c(A)$ is among the
CatNet's top-3 scores with score gap $S_g = 0.3$. We perform
``non-maximum suppression'' on the mixed scores of the positive
detections per category to obtain a sparse set of boxes. Non-maximum
suppression is performed by picking the most confident (maximum score)
detection and removing its neighboring detections; then, picking the
second most confident detection left and removing its neighbors, and
continuing this process until there are no positive detections
left. We consider two patches to be neighbors if at least $30\%$ of
the smaller patch overlaps with the bigger patch (intersection over minimum greater than $0.3$).

It should be noted that an object instance may not
necessarily fall completely inside any cell from our annocell
hierarchy at a certain level even if there might exist a patch of the
same size outside the hierarchy that completely includes that object
instance. This is because the annocell hierarchy is constructed with $75\%$ overlap ($= 25\%$ shift) between neighboring cells at the same level of
resolution, and can therefore miss some object instances at a given
level even if the cell size is large enough to include the object (\eg
see Figure~\ref{fig:RelaxedCondition}). The only way to avoid this is
to make the hierarchy exhaustive at each resolution, \ie shifting
patches by only one pixel at the time.
Figure~\ref{fig:CNN_SamplesSet} illustrates some detection examples after running non-maximum suppression on the combined scores from the CatNet and ScaleNet.

\subsection{Results}

\begin{figure}[t]
\centering
\includegraphics[width=0.95\linewidth]{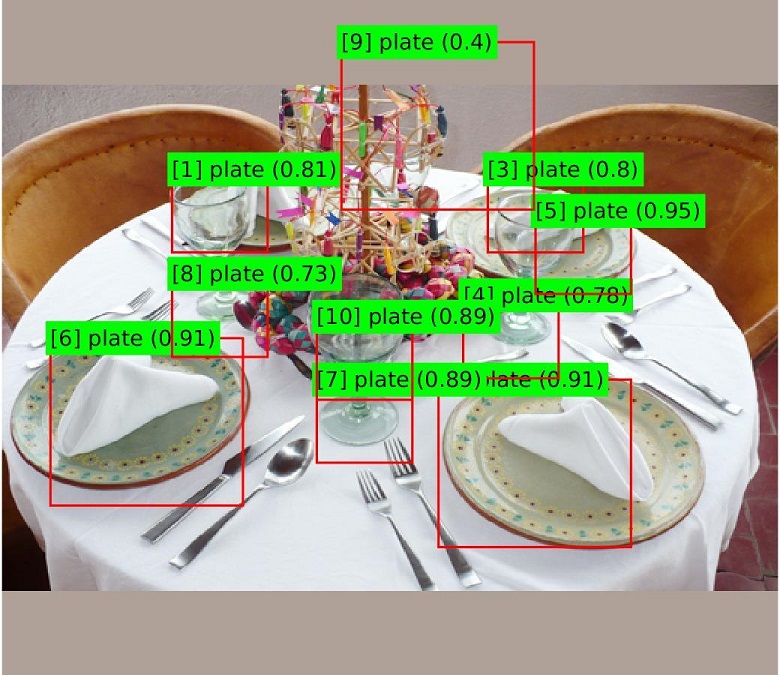} \\
\vspace{2mm}
\caption[Confusing CNN output example]{Confusing CNN detections example.}
\label{fig::ConfusingTheModel}
\end{figure}

We generate PR-curves by thresholding the scores of surviving detections after non-maximum suppression. Note that we would like to detect as many true instances as possible (high recall) for as few mistakes as possible (high precision or low false detection rate) which invariably necessitates a trade-off. Figure~\ref{fig:PR_Curves} shows Precision--Recall curves for twelve different methods that we ran on the data set of 284 images for all object categories. According to Figure~\ref{fig:PR_Curves} the EP model-based detection performance improves as more classifiers are run and incorporated into the model. However, the full posterior detector seems to perform worse than information pursuit after 140 questions (after 70 IP steps with the batch size of 2) which seems counterintuitive because we expect to achieve better performance by incorporating more evidence. Note that by incorporating more classifiers we do not necessarily get better results due to the classifiers' noise and increased likelihood of inconsistencies between the classifiers' output. For example, consider Figure~\ref{fig::ConfusingTheModel} where the 1st, 3rd, and
4th most confident plate detections (from CNN) are actually not a plate but the
top or bottom of glasses; all of the annocells corresponding to these
detections are from the finest level of the annocell hierarchy that
are expected to be chosen later during the IP selection criterion and
potentially degrade detection performance. Hence, integrating these classifiers would result in poorer inference due to the added classifiers inconsistency. Note that the model integrates the ScaleNet outputs in an attempt to suppress
configurations with scale inconsistency (\eg the incorrect plate
detections in Figure~\ref{fig::ConfusingTheModel}). However, since a
multiplication of CatNet and ScaleNet data model are used during posterior sampling (see~\eqref{eq:cond.prob} and consider the conditional independence assumption of CatNet and ScaleNet Dirichlet distributions), the model may not be able to completely suppress such configuration if the output of one of the CatNet or ScaleNet networks is large enough to compensate for the smaller one.

In Figure~\ref{fig:PR_Curves} we have included the P-R curves of
model-based detection for two variations ``Rand. 140 Q.'' and
``Rand. 30 Q.'' with the same number of questions as in the two IP
tests except that the questions are chosen at random; we have also
included the result of CNN classifiers (no model) when 140, 60, 30, and 10
patches are randomly chosen and processed. One can see that the result
with 140 randomly selected questions (the cyan curve in
Figure~\ref{fig:PR_Curves}) is almost the same as IP with only 30
questions asked (the magenta curve in Figure~\ref{fig:PR_Curves}) which
emphasizes the importance of efficient question selection in the
Bayesian approach. The Bayesian approach provides a natural framework
unifying the evidence collected from running tests and our prior
knowledge encoding the contextual relations between different scene
entities. Our experiments demonstrate that it makes significant
difference to choose patches appropriately using our IP strategy
versus randomly choosing them. In addition to saving the time it takes
to process patches that do not provide much information, we can
monitor the confidence of our detections (measured by conditional
entropy) and stop processing more patches once the uncertainty saturates or starts to increase in case of conflicting evidence. The model-based approach with enough questions asked outperforms the CNN classifiers (higher precision at high recall area in the right). The result of running the
CNN classifier on a small fraction of randomly selected annocells does not achieve high recall (see Figure~\ref{fig:PR_Curves}).

\begin{figure*}[t]
\centering
\includegraphics[width=0.99\linewidth]{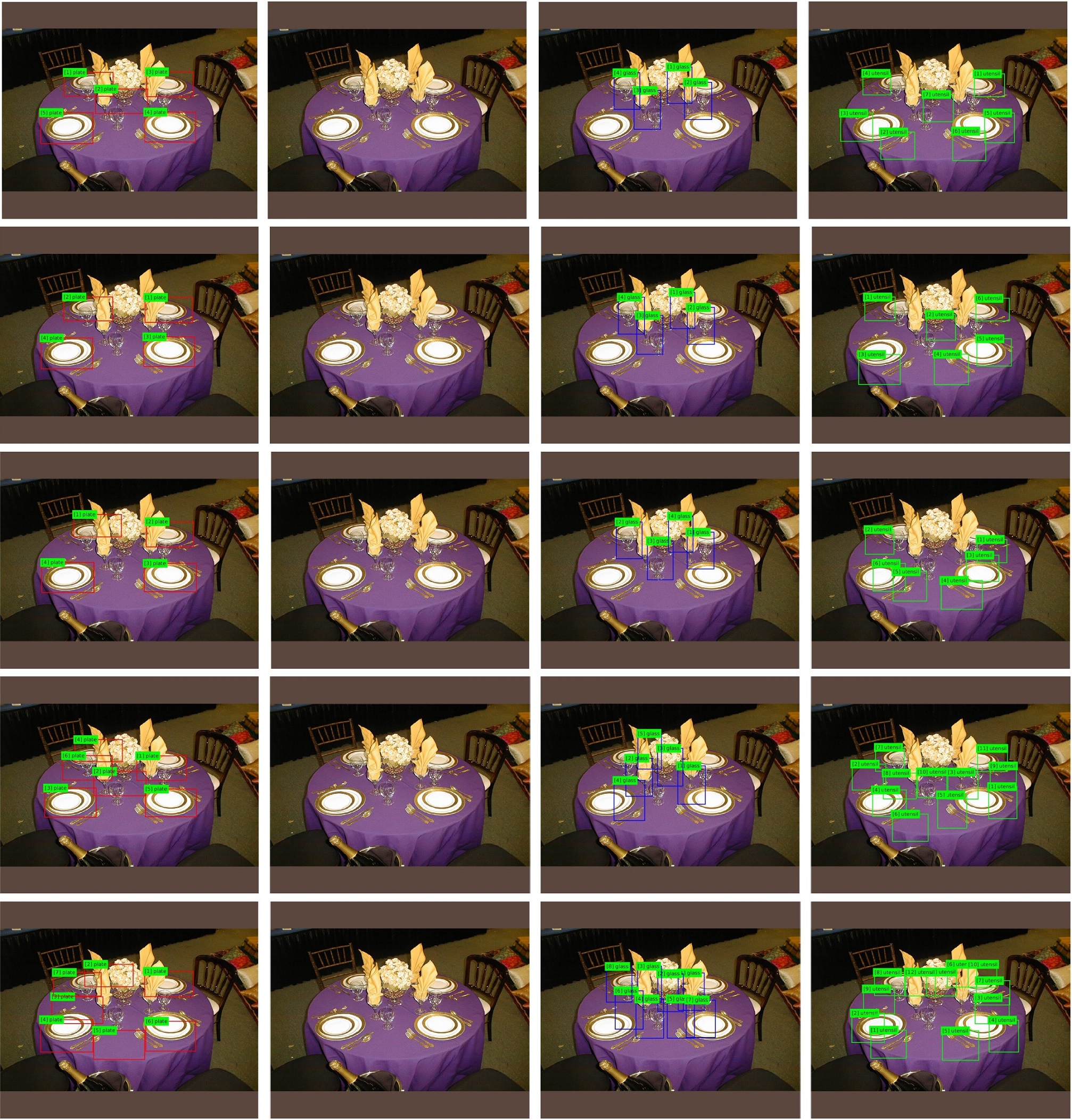}
\caption[Model Based detections.]{\small{Detections based on full posterior, IP-140, IP-30, Rand-140, and Rand-30 (from top to bottom) for ``plate'', ``bottle'', and ``glass'', and ``utensil'' categories (from left to right). The ordinal numbers in brackets represent the confidence rank of detections per category.}}
\label{fig:ModelBasedDetec1}
\end{figure*}

\begin{figure*}[t]
\centering
\includegraphics[width=0.99\linewidth]{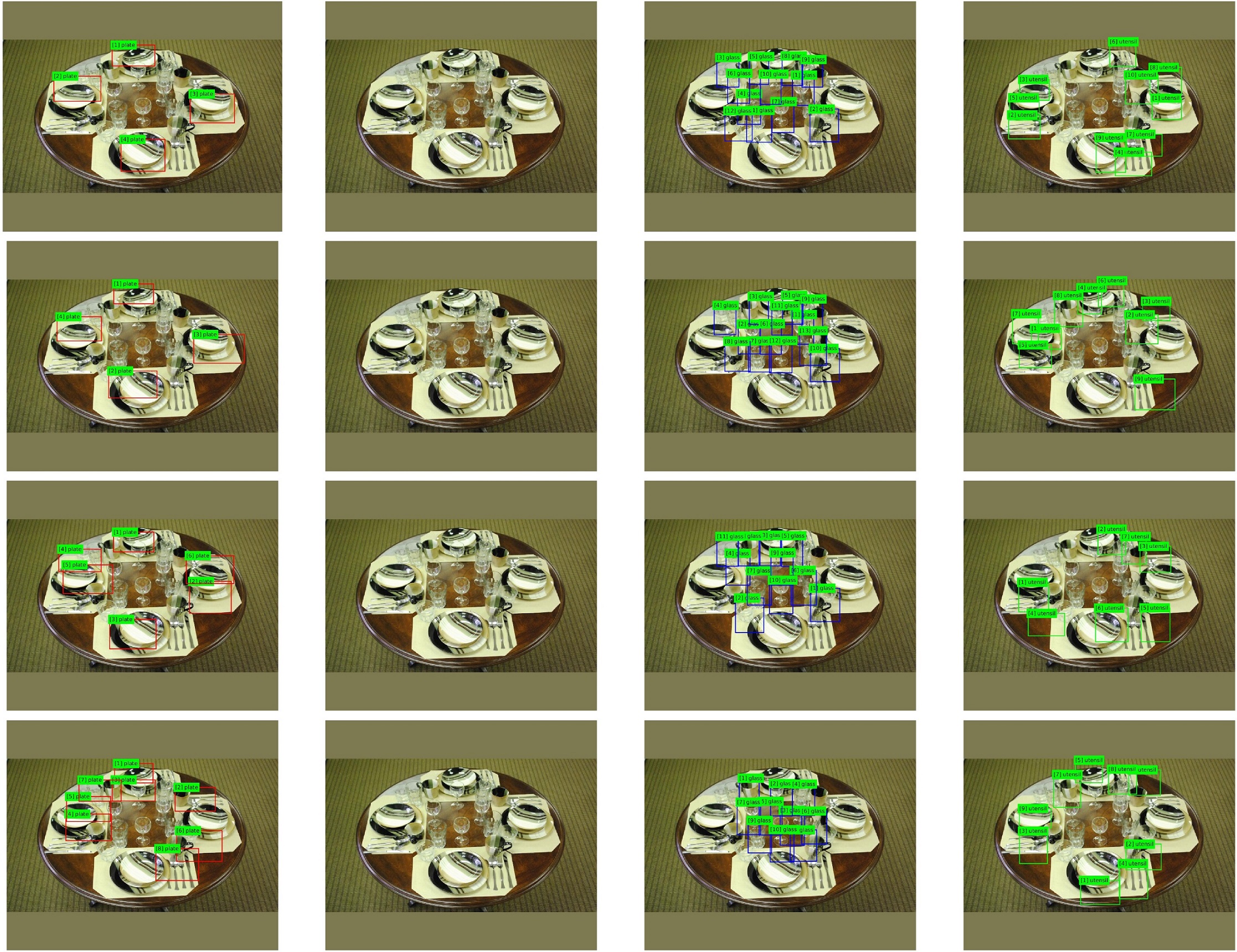}
\caption[Model Based detections.]{\small{Detections based on full posterior, IP-140, IP-30, and Rand-30 (from top to bottom) for ``plate'', ``bottle'', and ``glass'', and ``utensil'' categories (from left to right). The ordinal numbers in brackets represent the confidence rank of detections per category.}}
\label{fig:ModelBasedDetec2}
\end{figure*}

\begin{figure*}[t]
\centering
\includegraphics[width=0.99\linewidth]{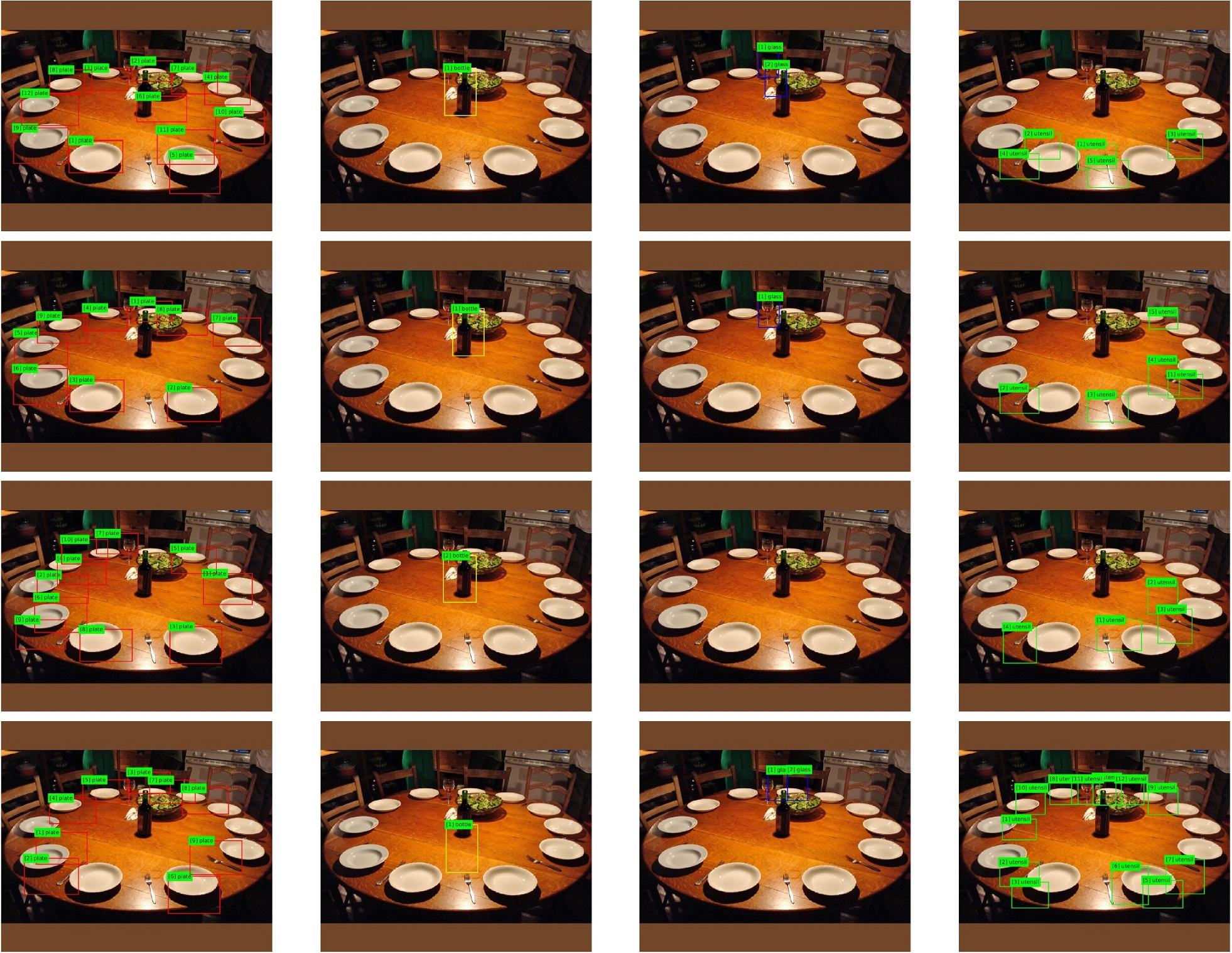}
\caption[Model Based detections.]{\small{Detections based on full posterior, IP-140, IP-30, and Rand-30 (from top to bottom) for ``plate'', ``bottle'', and ``glass'', and ``utensil'' categories (from left to right). The ordinal numbers in brackets represent the confidence rank of detections per category.}}
\label{fig:ModelBasedDetec3}
\end{figure*}

Figure~\ref{fig:ModelBasedDetec1},~\ref{fig:ModelBasedDetec2},~\ref{fig:ModelBasedDetec3} show some qualitative model based detections based on full posterior, IP, and random selection of patches. One can see that detections based on IP outperform random selection for the same number of patches.

\section{Conclusion}
\label{sec:conclusion}

We proposed a new approach for multi-category object recognition,
called ``Information Pursuit'' (IP), that sequentially investigates
patches from an input test image in order to come up with an accurate
description by processing as few patches as possible. Our approach
follows the Bayesian framework with a prior model that incorporates
the contextual relations between different scene entities such as the
spatial and semantic relations among object instances, consistency of
scales, constraints imposed by coplanarity of objects, \etc. As proof
of concept we applied the IP approach to table-setting scenes. We
designed a novel generative model on attributed graphs with flexible
structure where each node in the graph corresponds to an object
instance attributed by its category label and 3D pose. This scene generation model model was not directly used in our IP framework, but the statistics
calculated from its samples were used to learn a Markov Random Field
(MRF) model employed directly by IP. Whereas, the scene generation model could be
learned efficiently from the limited number of annotated images, the
MRF model offered faster conditional inference. The entropy pursuit
search strategy selects patches from the input image sequentially and
investigates them to collect evidence about the scene. To investigate
each patch we utilized state-of-the-art convolutional neural networks
(CNNs). We introduced a new dataset of about 3000 fully annotated
table-setting scenes to learn the scene generation model,
to train a battery of CNN classifiers, and to test the
performance of the IP algorithm.
In summary, we studied the possibility of generating a scene
interpretation by investigating only a fraction of all patches from an
input image using the entropy pursuit Bayesian approach. The Bayesian
framework is the natural approach for integrating contextual relations
and the evidence collected using tests. We were able to show that by
choosing the right patches in the right order we can identify an
accurate interpretation by processing only a fraction of all patches
from an input image.

\appendix
\section*{Appendix}

\section{Prior Distributions on Table Settings}
\begin{figure}
\centering
\includegraphics[width=0.95\linewidth]{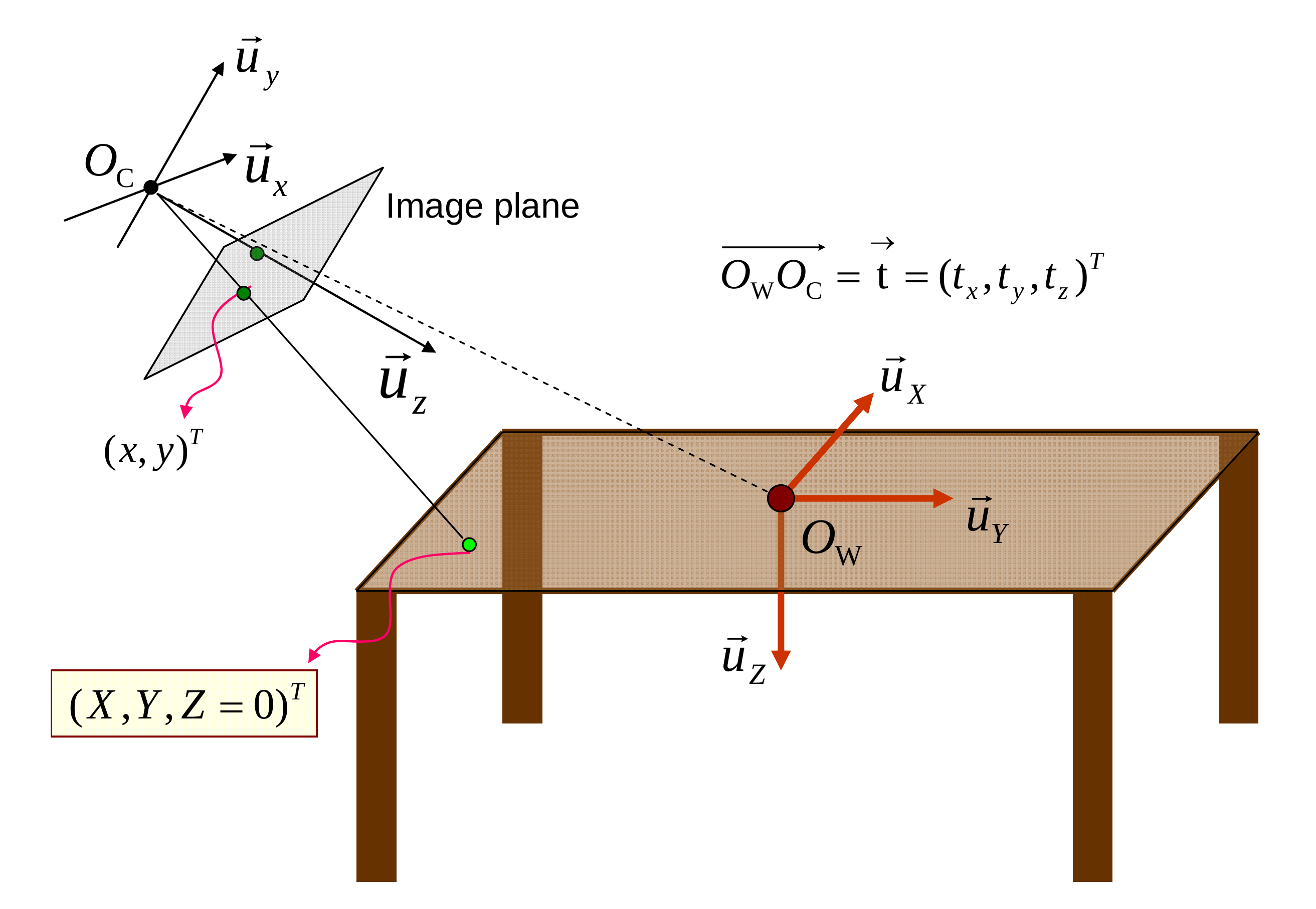}
\caption[Table and camera coordinate systems]{Table and camera coordinate systems.}
\label{fig:table}
\end{figure}

We work with categories $\mathcal{C} = \{\text{plate}, \text{bottle}, \text{glass}, \text{utensil}\}$ which are amongst the most annotated categories in our table-setting dataset. Instances from $\mathcal{C}$ are placed on a table whose geometric properties are denoted by $S$. In the simplified case that the table is rectangular we have $S = (L_s,W_s)$ where $L_s$ and $W_s$, respectively, represent the length and width of the table. We consider a world coordinate system whose origin is located at the center of the table's surface, whose $z$ axis is orthogonal to the table's surface, and assuming a rectangular table, the $x$ and $y$ axes are parallel to the edges of the table as illustrated in Figure~\ref{fig:table}. We also define a coordinate systems attached to the camera as shown in Figure~\ref{fig:table}.

\subsection{Attributed Graph Model}
The general form of the attributed graph model is given by \eqref{eq:graph.model}.
We assume that given $S$ (and of course the scene type) the number of root nodes from different
categories are independent:
\begin{align}
\label{RootIndpendence}
p^{(0)}(\mathbf n | S) =
\prod_{c\in \mathcal{C}} p^{(0)}_{c}(n_{c} | S),
\end{align}
where each of the univariate conditional distributions is modeled using a {\em Poisson} distribution with an average rate proportional to the
scene area $A_s = L_s W_s$, so that
\[
p^{(0)}_{c}(n_{c} | S) = e^{-\alpha_c A_s} \frac{(\alpha_c A_s)^{n_c}}{n_c!},
\]
 resulting in $|\mathcal{C}|=4$
parameters $\alpha_c, c\in \mathcal C$.  We also decouple the offspring
counts, letting $ch_M(c)$ denote the children of category $c$ in the master graph.
\begin{align}
\label{NonRootIndpendence}
p^{(c)}(\mathbf n) = \prod_{c' \in ch_M(c)}
p^{(c)}_{c'}(n_{c'}).
\end{align}
These distributions are modeled non-parametrically between $0$ and $l_{c_0,c}$ chosen as follows (following edges $c_0 \rightarrow c$ of the master graph): $l_{\text{plate},\text{utensil}}=l_{\text{plate},\text{glass}}=l_{\text{utensil},\text{utensil}}=3$ and $l_{\text{bottle},\text{glass}}=4$. This means that for example we allow at most three utensils to be adopted by a plate instance.

We now describe the pose distributions, starting with the root (spontaneous) objects.
For each category $c$, the table region is divided into two parts: a rectangular strip of width $d_c$ starting from the edges, and the remainder interior region. The object's center is placed in the central region with probability $\rho_c$ and in the outer strip with probability $1-\rho_c$.  Conditionally to this choice, the distribution is uniform within each area. Plates are represented by circles on the table (they are flat), glasses and bottles by ellipsoids with a vertical principal direction and rotation invariant around this axis. Utensils are represented as horizontal ellipses (flat also), with orientation following a Von Mises distribution whose mean is set to be $90$ degrees from the orientation of the nearest table edge. The dispersion parameter of the von Mises distribution is set to zero if this instance is located farther than 40 centimeters from all sides of the table and greater than zero otherwise. Note that a von Mises distribution with zero dispersion parameter is basically a uniform distribution in $(0,2\pi]$. For simplicity, the object sizes are fixed (e.g., 25 centimeters for plate diameters).

We specify the pairwise pose distributions, $p^{(c)}(\theta | c_{0}, \theta_{0},S)$, where $(c_0,\theta_0)$ is the category and pose of the parent object, by a radial distribution and a conditional angular distribution in a polar system centered at the location of the parent object. We model the relative pose of a parent-child object pair assuming that their relative location is independent from their relative orientation. We chose a scaled beta distribution for the radial distance between pairs of parent-child objects and either a von Mises (single or mixture) or uniform distribution for the angular location of the child in the periphery of the parent object. Normally, we expect a $c_1$-category parent and a $c_2$-category child to be within some distance from each other in order to justify their local contextual relationship. Let $d_{(c_0,c)}$ denote this user-defined distance. The scale of the beta distribution used for the radial distance of a $(c_0,c)$ object pair is set to $d_{(c_0,c)}$ and kept fixed throughout design and learning. The set of pose distribution parameters therefore includes the set of beta and von
Mises distributions' parameters for different categories.

\section{Prior Distribution on Camera Rotation}

The distribution of the rotation angles $\mathbf \psi$ is defined conditionally to translation $T$. Let $u_x,u_y,u_z$ denote the orthonormal axes for world coordinates, and $u_{x'}, u_{y'}, u_{z'}$ the same axes for camera coordinates, the following constraints will be used: $u_{z'} \sim -T/\|T\|$ (the camera points to the center of the table); $u_{x'} \perp u_z$ (the horizontal direction in the image plane is  nearly horizontal in 3D space); $u_{y'}^{\top} u_z < 0$ (the vertical direction in image plane points upward in 3D). Let
\begin{align*}
& \bar u_{z'} = -\frac{T}{\|T\|}\\
&\bar u_{x'} = \frac{\bar u_{z'} \times u_z }{ \|\bar u_{z'} \times u_z\|} \\
& \bar u_{y'} = \bar u_{z'} \times \bar u_{x'}.
\end{align*}
Letting $\boldsymbol \mu = (\mu_x, \mu_y, \mu_z)$ denote the angle defining the rotation angles mapping $(u_x, u_y, u_z)$ to $(\bar u_{x'}, \bar u_{y'}, \bar u_{z'})$, we take  $\psi_x, \psi_y, \psi_z$ conditionally independent given translation $T$, the marginals being von Mises distribution with means  $\mu_x, \mu_y$ and $\mu_z$. These angles are explicitly given by the formula
\begin{align}
\nonumber
&\mu_y = \sin^{-1}(-\bar u_{x'}\cdot u_z)\\
\label{eq:angles}
&\mu_x = \measuredangle\left(\frac{\bar u_{z'}\cdot u_z}{\cos\mu_y}, \frac{\bar u_{y'}\cdot u_z}{\cos\mu_y}\right) \\
\nonumber
& \mu_z = \measuredangle\left(\frac{\bar u_{x'}\cdot u_x}{\cos\mu_y}, \frac{\bar u_{x'}\cdot u_y}{\cos\mu_y}\right)
 \end{align}
 where $\measuredangle(a,b)$ is the angle $\theta$ defined by $\cos\theta=a$ and $\sin\theta=b$.

\newpage
\bibliographystyle{spbasic}
\bibliography{ref}

\end{document}